\pdfoutput=1
\documentclass{article}

    \PassOptionsToPackage{numbers, compress}{natbib}
    \usepackage[final]{neurips_data_2023}

\usepackage[utf8]{inputenc} % allow utf-8 input
\usepackage[T1]{fontenc}    % use 8-bit T1 fonts
\usepackage{hyperref}       % hyperlinks
\usepackage{url}            % simple URL typesetting
\usepackage{booktabs}       % professional-quality tables
\usepackage{amsfonts}       % blackboard math symbols
\usepackage{nicefrac}       % compact symbols for 1/2, etc.
\usepackage{xcolor}         % colors

\usepackage{amsmath}
\usepackage{caption}
\usepackage{amsfonts}
\usepackage[pdftex]{graphicx}
\usepackage{setspace}
\usepackage{url}
\usepackage{multirow}
\usepackage{makecell}
\usepackage{subfigure}
\usepackage{color}
\usepackage{xspace}
\usepackage{amsmath}
\usepackage{adjustbox}
\usepackage{booktabs}
\usepackage{soul}
\usepackage{float}
\usepackage{textcomp}
\usepackage{enumitem}
\usepackage{algorithm}
\usepackage{algpseudocode}
\usepackage{tabularray}
\usepackage{wrapfig}
\usepackage{makecell}

\definecolor{firstcolor}{HTML}{009E73}
\definecolor{secondcolor}{HTML}{0072B2}
\definecolor{thirdcolor}{HTML}{D55E00}

\newcommand\colorfirst[1]{\textcolor{firstcolor}{\textbf{#1}}}
\newcommand\colorsecond[1]{\textcolor{secondcolor}{\textbf{#1}}}\newcommand\colorthird[1]{\textcolor{thirdcolor}{\textbf{#1}}}

\title{Evaluating Graph Neural Networks for Link Prediction: Current Pitfalls and New Benchmarking}

\author{%
Juanhui Li$^1$\thanks{Equal contribution.}\,\,,  Harry Shomer$^{1\ast}$\,\,, Haitao Mao$^1$\,\,, Shenglai Zeng$^1$ \\ 
\textbf{Yao Ma}$^2$\,\,, \textbf{Neil Shah}$^3$\,\,, \textbf{Jiliang Tang}$^1$\,\,, \textbf{Dawei Yin}$^4$ \\
 $^1$Michigan State University, $^2$Rensselaer Polytechnic Institute\\
 $^3$Snap Inc., $^4$Baidu Inc.\\
 \texttt{\{lijuanh1,shomerha,haitaoma,zengshe1,tangjili\}@msu.edu}
 \\
 \texttt{may13@rpi.edu, nshah@snap.com, yindawei@acm.org}
}

\begin{document}
\maketitle

\begin{abstract}
  Link prediction attempts to predict whether an unseen edge exists based on only a portion of edges of a graph. A flurry of methods have been introduced in recent years that attempt to make use of graph neural networks (GNNs) for this task. Furthermore, new and diverse datasets have also been created to better evaluate the effectiveness of these new models. However, multiple pitfalls currently exist that hinder our ability to properly evaluate these new methods.  These pitfalls mainly include: (1) Lower than actual performance on multiple baselines, (2) A lack of a unified data split and evaluation metric on some datasets, and (3) An unrealistic evaluation setting that uses easy negative samples. 
  To overcome these challenges, we first conduct a fair comparison across prominent methods and datasets, utilizing the same dataset and hyperparameter search settings. We then create a more practical evaluation setting based on a {\bf He}uristic {\bf R}elated Sampling {\bf T}echnique (HeaRT), which samples hard negative samples via multiple heuristics.  The new evaluation setting helps promote new challenges and opportunities in link prediction by aligning the evaluation with real-world situations. Our implementation and data are available at \url{https://github.com/Juanhui28/HeaRT}. 
\end{abstract}

\section{Introduction} \label{sec:intro}

The task of link prediction is to determine the existence of an edge between two unconnected nodes in a graph. Existing link prediction algorithms attempt to estimate the proximity of different pairs of nodes in the graph, where node pairs with a higher proximity are more likely to interact~\cite{lu2011link}. Link prediction is applied in many different domains including social networks~\cite{daud2020applications}, biological networks~\cite{kovacs2019network}, and recommender systems~\cite{huang2005link}. 

Graph neural networks (GNNs)~\cite{wu2020comprehensive} have gained prominence in recent years with many new frameworks being proposed for a variety of different tasks. Corresponding to the rise in popularity of GNNs, there has been a number of studies that attempt to critically examine the effectiveness of different GNNs on various tasks. This can be seen for the task of node classification~\cite{shchur2018pitfalls}, graph classification~\cite{errica_fair_2020}, knowledge graph completion (KGC)~\cite{olddog, ali2021bringing, sun2020re}, and others~\cite{dwivedi2023benchmarking}.

\begin{figure*}[t]
    \begin{center}
     \centerline{
        {\subfigure[ogbl-collab]
        {\includegraphics[width=0.33\linewidth]{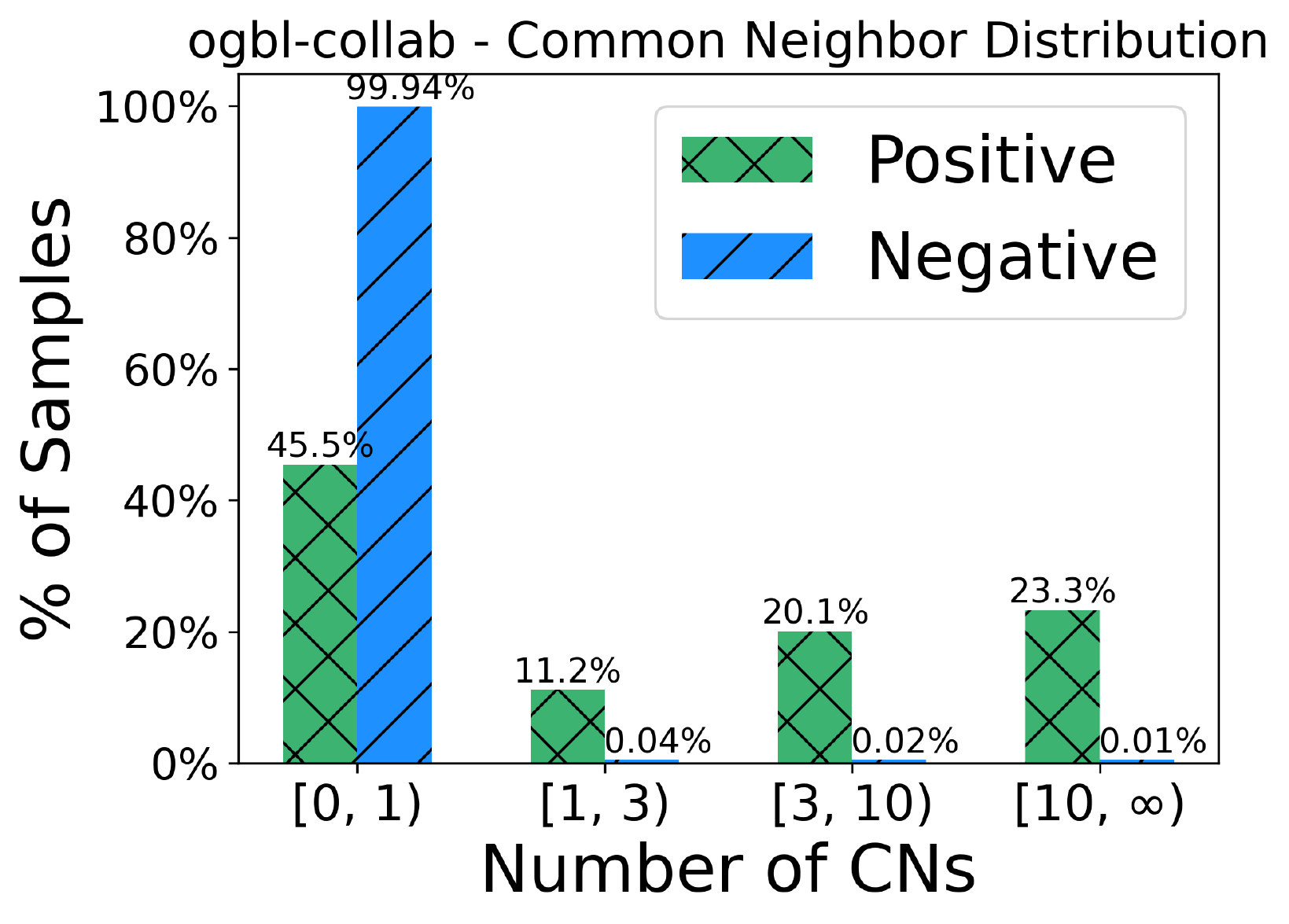} }}
        
        {\subfigure[ogbl-ppa]
        {\includegraphics[width=0.33\linewidth]{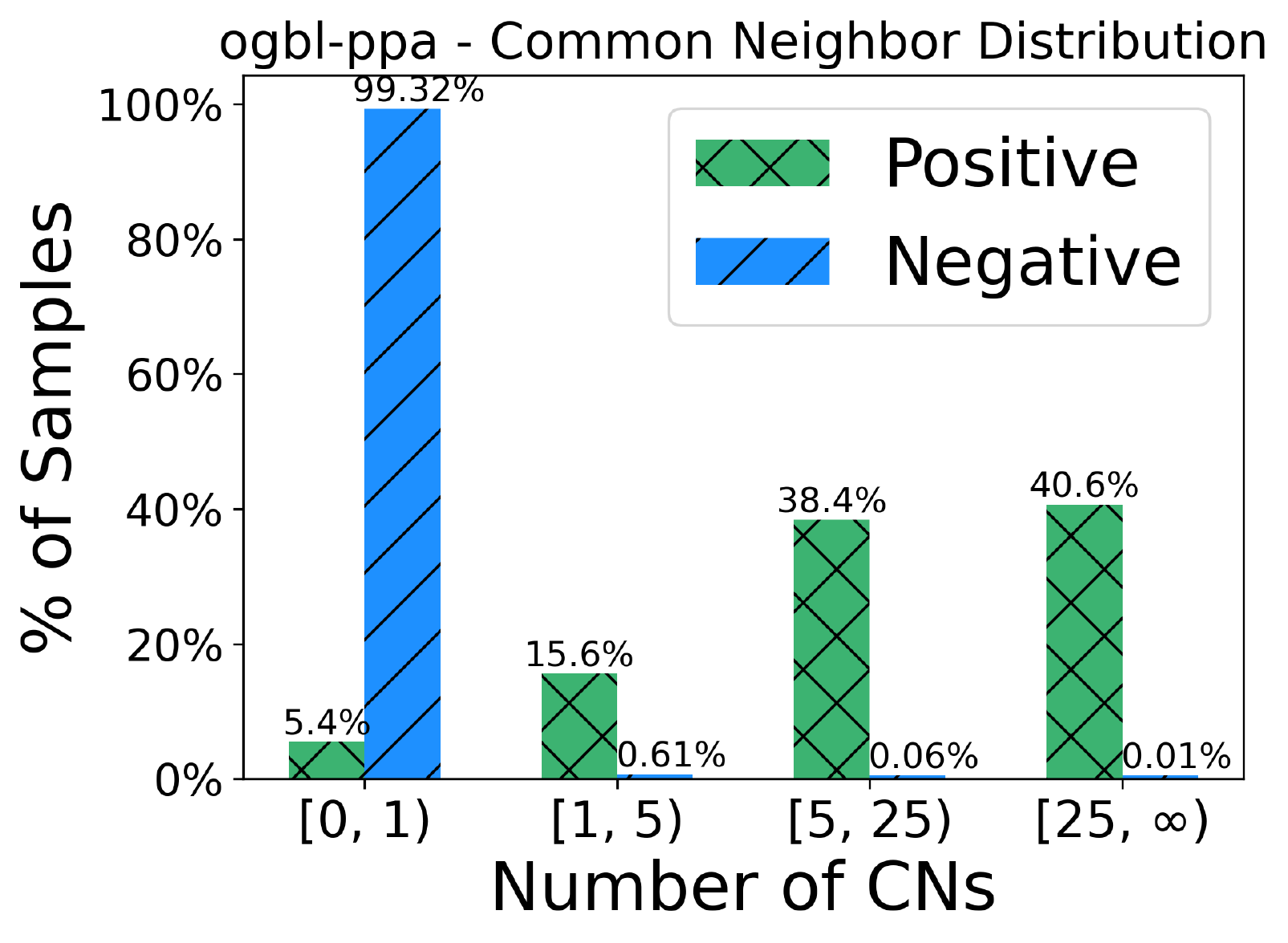} }}
        
        {\subfigure[ogbl-citation2]
        {\includegraphics[width=0.33\linewidth]{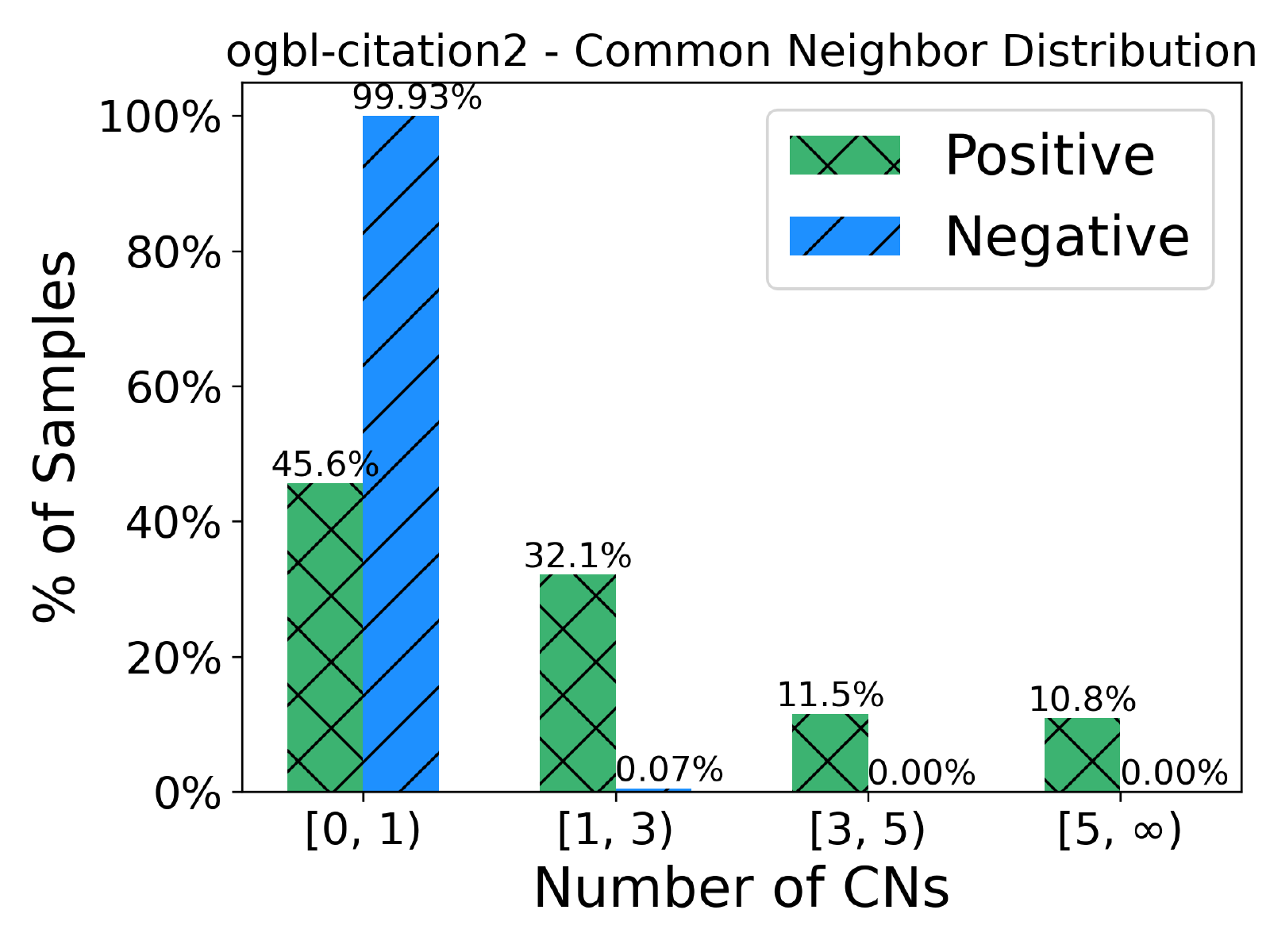} }}
    }
    
    \caption{Common neighbor distribution for the  positive and negative test samples for the ogbl-collab, ogbl-ppa, and ogbl-citation2 datasets under the existing evaluation setting.}

\label{fig:cn_ogb}
\end{center}
% \vspace{-.3in}
\end{figure*}

However, despite a number of new GNN-based methods being proposed~\cite{zhang2018link, chamberlain2022graph, yun2021neo, wang2023neural} for link prediction, there is currently no work that attempts to carefully examine recent advances in link prediction methods. Upon examination, we find that there are several pitfalls in regard to model evaluation that impede our ability to properly evaluate current methods. This includes: 
\begin{enumerate} [leftmargin=0.3in]
    \item {\bf Lower than Actual Performance}. We observe that the current performance of multiple models is underreported. For some  methods, such as standard GNNs, this is due to poor hyperparameter tuning. Once properly tuned, they can even achieve the best overall performance on some metrics (see SAGE~\cite{hamilton2017inductive}  in Table~\ref{table:small}). Furthermore, for other methods like Neo-GNN~\cite{yun2021neo} we can achieve around an 8.5 point increase in Hits@50 on ogbl-collab relative to the originally reported performance. This results in Neo-GNN achieving the best overall performance on ogbl-collab in our study (see Table~\ref{table:ogb}).   
    Such problems obscure the true performance of different models, making it difficult to draw reliable conclusions from the current results.

    \item {\bf Lack of Unified Settings}.  {For Cora, Citeseer, and Pubmed datasets~\cite{planetoid}, there exists no unified data split and evaluation metrics used for each individually.} For the data split, some works~\cite{velickovic2017graph, zhu2021neural} use a single fixed train/valid/test split with percentages 85/5/10\%. More recent works~\cite{chamberlain2022graph, wang2023neural} use 10 random splits of size 70/10/20\%. 
    In terms of the evaluation metrics, some studies~\cite{ chamberlain2022graph, wang2023neural} use ranking-based metrics such as MRR or Hits@K while others~\cite{kipf2016variational, zhu2021neural} report the area under the curve (AUC). 
    This is despite multiple studies that argue that AUC is a poor metric for evaluating link prediction~\cite{yang2015evaluating, huang2023link}. 
     Additionally, {for both the planetoid (i.e., Cora, Citeseer and Pubmed) and  ogbl-collab datasets}, some methods incorporate the validation edges during testing~\cite{chamberlain2022graph, hu2020open}, while others~\cite{yun2021neo, wang2023neural} don't.
    This lack of a unified setting {makes it difficult to draw a comparison and
    hampers our ability to determine which methods perform best on these datasets.}
   
    \item {\bf Unrealistic Evaluation Setting}. 
    During the evaluation, we are given a set of true samples (i.e., positive samples) and a set of false samples (i.e., negative samples).  We are tasked with learning a classifier $f$ that assigns a higher probability to the positive samples than the negatives. The current evaluation setting uses the same set of randomly selected negative samples for each positive sample. We identify two potential problems with the current evaluation procedure. {\bf (1)} It is not aligned with real-world settings.  In a real-world scenario, we typically care about predicting links for a specific node. For example, in friend recommendations, we aim to recommend friends for a specific user $u$. To evaluate such models for $u$, we strive to rank node pairs including $u$. However, this does not hold in the current setting as $u$ is not included in most of the negative samples. 
    {\bf (2)} The current evaluation setting makes the task too easy. As such, it may not reflect the model performance in real-world applications. This is because the nodes in a randomly selected negative ``node pair'' are likely to be unrelated to each other. As shown in Figure~\ref{fig:cn_ogb}, almost all negative samples in the test data have no common neighbors, a typically strong heuristic, making them trivial to classify them.

%     For example, let's say we are given a social network where an edge indicates friendship and we want to predict friends of a user {\it John}. Intuitively, we want to assign a higher probability to links where {\it John} is connected to a friend than ones where he isn't. If {\it John} is a friend of {\it Peter} but not {\it Sarah}, we hope that $f(\textit{John}, \textit{Peter}) > f(\textit{John}, \textit{Sarah})$ where $f$ is a classifier. However, the current evaluation doesn't necessarily test for this. Instead it compares positive samples to randomly sampled negative pairs, i.e., $f(\textit{John}, \textit{Peter}) > f(\textit{Rachel}, \textit{Joseph})$. A more comprehensive discussion is agiven in Section~\ref{sec:new_eval_problem}. Furthermore, we find that current negative samples utilized in evaluation are too easy. As shown in Figure~\ref{fig:cn_ogb}, almost all negative samples in the test data have no common neighbors, a typically strong heuristic, making them trivial to classify. 

\end{enumerate}

To account for these issues, we propose to first conduct a fair and reproducible evaluation among current link prediction methods under the existing evaluation setting. We then design a new evaluation strategy that is more aligned with a real-world setting and detail our results. Our key contributions are summarized below:

\begin{itemize} [leftmargin=0.3in]
    \item {\bf Reproducible and Fair Comparison}. We conduct a fair comparison of different models across multiple common datasets. To ensure a fair comparison, we tune all models on the same set of hyperparameters. We further evaluate different models using multiple types of evaluation metrics. For the Planetoid datasets~\cite{planetoid}, we further use a unified data split to facilitate a point of comparison between models. To the best of our knowledge,  there are no recent efforts to comprehensively benchmark link prediction methods (several exist for KGC~\cite{sun2020re, ali2021bringing, olddog}). Furthermore, we open-source the implementation in our analysis to enable others in their analyses. 
    \item {\bf New Evaluation Setting}. We recognize that the current negative sampling strategy used in evaluation is unrealistic and easy. To counter these issues, we first use a more realistic setting of  tailoring the negatives to each positive sample. This is achieved by restricting them to be corruptions of the positive sample (i.e., containing one of its two nodes {as defined in Eq.~(\ref{eq:candidate_set})}). Given the prohibitive cost of utilizing all possible corruptions, we opt instead to only rank against $K$ negatives for each positive sample. In order to choose the most relevant and difficult corruptions, we propose a {\bf He}uristic {\bf R}elated Sampling {\bf T}echnique (HeaRT),
    which selects them based on a combination of multiple heuristics. This creates a more challenging task than the previous evaluation strategy and allows us to better assess the capabilities of current methods.
  
\end{itemize}
 
The rest of the paper is structured as follows. In Section~\ref{sec:prelims} we introduce the models, datasets, and settings used for conducting a fair comparison between methods. In Section~\ref{sec:benchmark} we show the results of the fair comparison under the existing evaluation setting and discuss our main observations. Lastly, in Section~\ref{sec:new_eval} we introduce our new evaluation setting. We then detail and discuss the performance of different methods using our new setting.

\section{Preliminaries} \label{sec:prelims}

\subsection{Task Formulation} \label{sec:link_pred_task}
{
In this section we formally define the task of link prediction. Let $\mathcal{G} = \{ \mathcal{V}, \mathcal{E} \}$ be a graph where $\mathcal{V}$ and $\mathcal{E}$ are the set of nodes and edges in the graph, respectively. Furthermore, let $X \in \mathcal{R}^{\lvert V \rvert \times d}$ be a set of $d$-dimensional features for each node. Link prediction aims to predict the likelihood of a link existing between two nodes given the structural and feature information. For a pair of nodes $u$ and $v$, the probability of a link existing, $p(u, v)$, is therefore given by:
\begin{equation} \label{eq:link_prob}
    p(u, v) = p(u, v \: | \: \mathcal{G}, X).
\end{equation}
Traditionally, $p(u, v)$ was estimated via non-learnable heuristic methods~\cite{menon2011link,liben2003link}. More recently, methods that use learnable parameters have gained popularity~\cite{zhang2018link, chamberlain2022graph}. These methods attempt to estimate $p(u, v)$ via a learnable function $f$ such that:
\begin{equation}
    p(u, v) = f(u, v \: | \: \mathcal{G}, X, \Theta),
\end{equation}
where $\Theta$ represents a set of learnable parameters. A common choice of $f$ are graph neural networks~\cite{kipf2017semi}. In the next subsection we detail the various link prediction methods used in this study.
}
\subsection{Link Prediction Methods} 
\label{sec:method}

In this section we given an overview of the different methods used in this study. 
Conventional methods~\cite{menon2011link,liben2003link} often exploit hand-craft graph structural properties (i.e., heuristics) between node pairs. GNNs attempt to learn the structural information to facilitate link prediction~\cite{zhang2021labeling,wang2023neural,chamberlain2022graph}. Given the strong performance of pairwise-based  heuristics~\cite{yun2021neo,wang2023neural}, some recent works use both GNNs and pairwise information, demonstrating strong performance. 

For our study, we consider both traditional and  state-of-the-art GNN-based models. They can be roughly organized into four categories. \textbf{1) Heuristic methods}:  Common Neighbor (CN)~\cite{newman2001clustering}, Adamic Adar  
 (AA)~\cite{adamic2003friends}, Resource Allocation  (RA)~\cite{zhou2009predicting}, Shortest Path~\cite{liben2003link}, and Katz~\cite{katz1953new}. These methods define a score to indicate the link existence based on the graph structure. Among them, CN, AA, and RA are based on the common neighbors, while Shortest Path and Katz are based on the path information. \textbf{2) Embedding methods}:  Matrix factorization (MF)~\cite{menon2011link}, Multilayer Perceptron (MLP)  and  Node2Vec~\cite{grover2016node2vec}. These methods are trained to learn low-dimensional node embeddings that are used to predict the likelihood of node pairs existing. \textbf{3) GNN methods}: GCN~\cite{yao2019graph}, GAT~\cite{velickovic2017graph}, SAGE~\cite{hamilton2017inductive}, and GAE~\cite{kipf2016variational}. These methods attempt to integrate the multi-hop graph structure based on the message passing paradigm. \textbf{4) GNN +  Pairwise Information methods}: Standard GNN methods, while powerful, are not able to capture link-specific information~\cite{zhang2021labeling}. As such, works have been proposed that augment GNN methods by including additional information to better capture the relation between the nodes in the link we are predicting. 
 SEAL~\cite{zhang2021labeling}, BUDDY~\cite{chamberlain2022graph}, and NBFNet~\cite{zhu2021neural} use the subgraph features. Neo-GNN~\cite{yun2021neo}, NCN~\cite{wang2023neural}, and NCNC~\cite{wang2023neural} are based on common neighbor information. Lastly, PEG~\cite{wang2022equivariant} uses the positional encoding derived from the graph structure.

\subsection{Datasets and Experimental Settings} \label{sec:prelim_exp}
In this section we summarize the datasets and evaluation and training settings. We note that the settings depend on the specific dataset. 
More details are given in Appendix~\ref{sec:app_data_parameter}.

\label{sec:setting}
\textbf{Datasets}. We limit our experiments to homogeneous graphs, which are the most commonly used datasets for link prediction. This includes the small-scale datasets, i.e., Cora, Citeseer, Pubmed~\cite{planetoid}, and  large-scale datasets in the OGB benchmark~\cite{hu2020open}, i.e., ogbl-collab, ogbl-ddi, ogbl-ppa, and ogbl-citation2. We summarize the statistics and split ratio of each dataset in Appendix~\ref{sec:app_data_parameter}.

{\bf Metrics}. For evaluation, we use both the area under the curve (AUC) and ranking-based metrics, i.e.,  mean reciprocal rank (MRR) and Hits@K. For Cora, Citeseer, and Pubmed we adopt $K \in \{1, 3,10, 100\}$. We note that $K=100$ is reported in some recent works~\cite{chamberlain2022graph,wang2023neural}). However due to the small number of negatives used during evaluation (e.g., $\approx 500$ for Cora and Citeseer) $K=100$ is likely not informative. For the OGB datasets, we adopt $K \in \{20, 50, 100\}$ to keep consistent with the original study~\cite{hu2020open}. Please see Appendix~\ref{sec:eval_metrics_defs} for the formal definitions of the various evaluation metrics.

{\bf Hyperparameter Ranges}. We conduct a grid hyperparameter search  across a comprehensive range of values. For Cora, Citeseer, and Pubmed this includes: learning rate (0.01, 0.001), dropout (0.1, 0.3, 0.5), weight decay (1e-4, 1e-7, 0), number of model layers (1, 2, 3), number of prediction layers (1, 2, 3), and the embedding size (128, 256). Due to the large size of the OGB datasets, it's infeasible to tune over such a large range. Therefore, following the most commonly used settings among published hyperparameters, we fix the weight decay to 0, the number of model and prediction layers to be 3, and the embedding size to be 256. The best hyperparameters are chosen based on the validation performance. 
 We note that several exceptions exist to these ranges when they result in significant performance degradations (see Appendix~\ref{sec:app_data_parameter} for more details). We further follow the existing setting and only sample one negative sample per positive sample during training.

{\bf Existing Evaluation Settings}. In the  evaluation stage, the same set of randomly sampled negatives are used for all positive samples. We note that one exception is ogbl-citation2, where they randomly sample 1000 negative samples per positive sample. For Cora, Citeseer, and Pubmed the number of negative samples is equal to the number of positive samples. For the OGB datasets, we use the existing fixed set of randomly chosen negatives found in~\cite{hu2020open}. 
Furthermore, for ogbl-collab we follow the existing protocol~\cite{hu2020open} and include the validation edges in the training graph during testing. This setting is adopted on ogbl-collab under both the existing and new evaluation setting.

\begin{table}[]
\centering
% \footnotesize
 \caption{Results on Cora, Citeseer, and Pubmed(\%) under the existing evaluation setting. Highlighted are the results ranked \colorfirst{first}, \colorsecond{second}, and  \colorthird{third}.}
 \begin{adjustbox}{width =1 \textwidth}
\begin{tabular}{cc|cccccc}

\toprule
 &\multirow{2}{*}{Models} & \multicolumn{2}{c}{Cora} &\multicolumn{2}{c}{Citeseer}  &\multicolumn{2}{c}{Pubmed}   \\ 
  & &MRR& AUC &MRR &AUC &MRR &AUC \\
  \midrule
   \multirow{5}{*}{Heuristic}  &CN & {20.99} & {70.85} &{28.34} & {67.49} & {14.02} & {63.9} \\
&AA &{31.87} & {70.96} & {29.37} &{67.49} & {16.66} &{63.9} \\
&RA &{30.79} &{70.96} & {27.61} &{67.48} & {15.63} & {63.9} \\
&Shortest Path &{12.45} &{81.08} &{31.82} & {75.5} & {7.15} &{74.64} \\
&Katz &{27.4} &{81.17} &{38.16} &{75.37} & {21.44} &{74.86} \\
\midrule
  \multirow{3}{*}{Embedding}  & Node2Vec & \colorthird{37.29 ± 8.82} & 90.97 ± 0.64 & 44.33 ± 8.99 & 94.46 ± 0.59 & 34.61 ± 2.48 & 93.14 ± 0.18 \\
 & MF & 14.29 ± 5.79 & 80.29 ± 2.26 & 24.80 ± 4.71 & 75.92 ± 3.25 & 19.29 ± 6.29 & 93.06 ± 0.43 \\
 & MLP & 31.21 ± 7.90 & 95.32 ± 0.37 & 43.53 ± 7.26 & 94.45 ± 0.32 & 16.52 ± 4.14 & 98.34 ± 0.10 \\
 \midrule
  \multirow{4}{*}{GNN} & GCN & 32.50 ± 6.87 & 95.01 ± 0.32 & 50.01 ± 6.04 & 95.89 ± 0.26 & 19.94 ± 4.24 & 98.69 ± 0.06 \\
 & GAT & 31.86 ± 6.08 & 93.90 ± 0.32 & 48.69 ± 7.53 & 96.25 ± 0.20 & 18.63 ± 7.75 & 98.20 ± 0.07 \\
 & SAGE & \colorfirst{37.83 ± 7.75} & \colorthird{95.63 ± 0.27} & 47.84 ± 6.39 & \colorsecond{97.39 ± 0.15} & 22.74 ± 5.47 & \colorthird{98.87 ± 0.04} \\
 & GAE & 29.98 ± 3.21 & 95.08 ± 0.33 & \colorsecond{63.33 ± 3.14} & \colorthird{97.06 ± 0.22} & 16.67 ± 0.19 & 97.47 ± 0.08 \\
 \midrule
  \multirow{7}{*}{\makecell{GNN+Pairwise Info}} & SEAL & 26.69 ± 5.89 & 90.59 ± 0.75 & 39.36 ± 4.99 & 88.52 ± 1.40 & \colorsecond{38.06 ± 5.18} & 97.77 ± 0.40 \\
  & BUDDY & 26.40 ± 4.40 & 95.06 ± 0.36 & \colorthird{59.48 ± 8.96} & 96.72 ± 0.26 & 23.98 ± 5.11 & 98.2 ± 0.05 \\
 & Neo-GNN & 22.65 ± 2.60 & 93.73 ± 0.36 & 53.97 ± 5.88 & 94.89 ± 0.60 & 31.45 ± 3.17 & 98.71 ± 0.05 \\
  & NCN & 32.93 ± 3.80 & \colorsecond{96.76 ± 0.18} & 54.97 ± 6.03 & {97.04 ± 0.26} & \colorthird{35.65 ± 4.60} & \colorsecond{98.98 ± 0.04} \\
 & NCNC & 29.01 ± 3.83 & \colorfirst{96.90 ± 0.28} & \colorfirst{64.03 ± 3.67} & \colorfirst{ 97.65 ± 0.30} & 25.70 ± 4.48 & \colorfirst{99.14 ± 0.03} \\
 & NBFNet & \colorsecond{37.69 ± 3.97} & 92.85 ± 0.17 & 38.17 ± 3.06 & 91.06 ± 0.15 & \colorfirst{44.73 ± 2.12} & 98.34 ± 0.02 \\

 & PEG & 22.76 ± 1.84 & 94.46 ± 0.34 & 56.12 ± 6.62 & 96.15 ± 0.41 & 21.05 ± 2.85 & 96.97 ± 0.39 \\
 \bottomrule
\end{tabular}
 \label{table:small}
 \end{adjustbox}
 % \vspace{-0.15in}
\end{table}

\section{Fair Comparison Under the Existing Setting} \label{sec:benchmark}

In this section, we conduct a fair comparison among link prediction methods. This comparison is spurred by the multiple pitfalls noted in Section~\ref{sec:intro}, which include lower-than-actual model performance, multiple data splits, and inconsistent evaluation metrics. 
These pitfalls hinder our ability to fairly compare different methods.
To rectify this, we conduct a fair comparison adhering to the settings listed in section~\ref{sec:setting}.

The results are split into two tables. The results for Cora, Citeseer, and Pubmed are shown in Table~\ref{table:small} and OGB in Table~\ref{table:ogb}. For simplicity, we only present the AUC and MRR for Cora, Citeseer, and Pubmed. For OGB datasets, we include AUC and the original ranking metric reported in~\cite{hu2020open} to allow a convenient comparison ($\text{Hits}@20$ for ogbl-ddi, $\text{Hits}@50$ for ogbl-collab, $\text{Hits}@100$ for ogbl-ppa, and MRR for ogbl-citation2). We use ``>24h" to denote methods that require more than 24 hours for either training one epoch or evaluation. OOM indicates that the algorithm requires over 50Gb of GPU memory. {Since ogbl-ddi has no node features, we mark the MLP results with a ``N/A".}
Additional results in terms of other metrics are presented in Appendix~\ref{sec:app_ben_more}. We have several noteworthy observations concerning the methods, the datasets, the evaluation settings, and the overall results. We highlight the main observations below.

\begin{table}
\centering
 \caption{Results on OGB datasets (\%) under the existing evaluation setting. Highlighted are the results ranked \colorfirst{first}, \colorsecond{second}, and  \colorthird{third}. }
 
 \begin{adjustbox}{width =1 \textwidth}
\begin{tabular}{cc|ccccccc}
\toprule
 &\multirow{2}{*}{Models} & \multicolumn{2}{c}{ogbl-collab} &\multicolumn{2}{c}{ogbl-ddi}  &\multicolumn{2}{c}{ogbl-ppa} &    ogbl-citation2  \\ 
 & & Hits@50 &AUC &Hits@20 &AUC &Hits@100 &AUC &MRR  \\
  \midrule
  \multirow{5}{*}{Heuristic}  &CN & 61.37	&82.78 &17.73	&95.2 &27.65	&97.22 & 74.3 \\
& AA	&64.17	&82.78 &18.61	&95.43 & 32.45	&97.23 & 75.96\\
& RA &63.81	&82.78 &6.23	&96.51 & \colorthird{49.33}	& 97.24 & 76.04\\
 &Shortest Path& {46.49}	&96.51 &0	&59.07 & 0	&99.13 & >24h\\
 &Katz& {64.33}	&90.54 & {17.73} & {95.2} & {27.65}&{97.22}&{74.3} \\
   \midrule
  \multirow{3}{*}{Embedding}  &  Node2Vec	& 49.06 ± 1.04	&96.24 ± 0.15& {34.69 ± 2.90}	&99.78 ± 0.04&   26.24 ± 0.96&	99.77 ± 0.00 & 45.04 ± 0.10\\\
&MF	&41.81 ± 1.67	&83.75 ± 1.77 & 23.50 ± 5.35	&99.46 ± 0.10 & 28.4 ± 4.62	&99.46 ± 0.10&50.57 ± 12.14 \\
&MLP	&35.81 ± 1.08	&95.91 ± 0.08 &N/A &N/A &  0.45 ± 0.04	&90.23 ± 0.00 & 38.07 ± 0.09\\
   \midrule
 \multirow{4}{*}{GNN}  &GCN	&54.96 ± 3.18	&97.89 ± 0.06 & \colorthird{49.90 ± 7.23}& {99.86 ± 0.03} &29.57 ± 2.90&	\colorthird{99.84 ± 0.03} &84.85 ± 0.07\\
&GAT	&55.00 ± 3.28&	97.11 ± 0.09 & 31.88 ± 8.83	& 99.63 ± 0.21 & OOM	&OOM&OOM\\
&SAGE	&59.44 ± 1.37&	\colorthird{98.08 ± 0.03} & {49.84 ± 15.56} &	\colorthird{99.96 ± 0.00} &41.02 ± 1.94	&99.82 ± 0.00 & 83.06 ± 0.09\\
% GIN	&58.42 ± 1.00	&98.23 ± 0.10 &38.44 ± 7.17	&99.87 ± 0.04 & 36.14 ± 2.37	&99.82 ± 0.02& 84.37 ± 0.14\\
&GAE	&OOM	&OOM &7.09 ± 6.02	&75.34 ±15.96 & OOM	&OOM &OOM\\
\midrule
 \multirow{7}{*}{GNN+Pairwise Info}  &SEAL & 63.37 ± 0.69&	95.65 ± 0.29 & 25.25 ± 3.90	&97.97 ± 0.19 &48.80 ± 5.61&	99.79 ± 0.02 &86.93 ± 0.43\\
&BUDDY	& \colorthird{64.59 ± 0.46}	&96.52 ± 0.40& 29.60 ± 4.75&	99.81 ± 0.02 & 47.33 ± 1.96	&99.56 ± 0.02& \colorthird{87.86 ± 0.18} \\
% \midrule
&Neo-GNN	&\colorfirst{66.13 ± 0.61}&	\colorfirst{98.23 ± 0.05} & 20.95 ± 6.03	& {98.06 ± 2.00} & 48.45 ± 1.01	&97.30 ± 0.14 &83.54 ± 0.32\\
&NCN	&63.86 ± 0.51	&97.83 ± 0.04& \colorfirst{76.52 ± 10.47}&	\colorfirst{99.97 ± 0.00} & \colorfirst{62.63 ± 1.15}&	\colorsecond{99.95 ± 0.01} & \colorsecond{89.27 ± 0.05} \\
&NCNC & \colorsecond{65.97 ± 1.03}	& \colorsecond{98.20 ± 0.05} & \colorsecond{70.23 ± 12.11} & \colorsecond{99.97 ± 0.01}& \colorsecond{62.61 ± 0.76} & \colorfirst{99.97 ± 0.01}& \colorfirst{89.82 ± 0.43}\\
% \midrule
&NBFNet	&OOM	&OOM & >24h & >24h & OOM	&OOM & OOM	 \\
&PEG & 49.02 ± 2.99	&94.45 ± 0.89  &30.28 ± 4.92	&99.45 ± 0.04&  OOM	&OOM & OOM	 \\
 \bottomrule
\end{tabular}
 \label{table:ogb}
 \end{adjustbox}
 % \vspace{-0.2in}
\end{table}

\noindent {\bf \underline{Observation 1}: Better than Reported Performance.} We find that for some models we are able to achieve superior performance compared to what is reported by recent studies. For instance, in our study Neo-GNN~\cite{yun2021neo} achieves the best overall test performance on ogbl-collab with a Hits@50 of $66.13$. In contrast, the reported performance in~\cite{yun2021neo} is only $57.52$, which would rank seventh under our current setting. This is because the original study~\cite{yun2021neo} does not follow the standard setting of including validation edges in the graph during testing. This setting, as noted in Section~\ref{sec:prelim_exp}, is used by all other methods on ogbl-collab. However it was omitted  by~\cite{yun2021neo}, resulting in lower reported performance.
Furthermore, on ogbl-citation2~\cite{hu2020open}, our results for the heuristic methods are typically around $75\%$ MRR. This significantly outperforms previously  reported results, which report an MRR of around $50\%$~\cite{zhang2021labeling, chamberlain2022graph}. The disparity arises as previous studies treat the ogbl-citation2 as a directed graph when applying heuristic methods. However, for GNN-based methods, ogbl-citation2 is typically converted to a undirected graph. We remedy this by also converting ogbl-citation2 to an undirected graph when computing the heuristics, leading to a large increase in performance.

Furthermore, with proper tuning, conventional baselines like GCN~\cite{kipf2017semi} and GAE~\cite{kipf2016variational} generally exhibit enhanced performance relative to what was originally reported across all datasets.  
For example, we find that GAE can achieve the second best MRR on Citeseer and GCN the third best Hits@20 on ogbl-ddi. 
A comparison of the reported results and ours are shown in Table~\ref{table:compare_reported}. We note that we report AUC for Cora, Citeseer, Pubmed as it was used in the original study.
These observations suggest that the performance of various methods are better than what was reported in their initial publications. However, many studies~\cite{chamberlain2022graph, wang2023neural,zhang2021labeling} only report the original performance for comparison, which has the potential to lead to inaccurate conclusions.
% \vspace{-0.05in}

\begin{table}
\centering
\footnotesize
 \caption{Comparison of ours and the reported results for GCN and GAE.}
 
 \begin{adjustbox}{width =0.95 \textwidth}
\begin{tabular}{l|cccc|l|ccc}
\toprule
&ogbl-collab & ogbl-ppa & ogbl-ddi & ogbl-citation2& & Cora & Citeseer & Pubmed  \\ 
 GCN &Hits@50 & Hits@100  & Hits@20 & MRR& GAE & AUC &AUC &AUC  \\
 \midrule
  Reported &47.14 ± 1.45 &  18.67 ± 1.32& 37.07 ± 5.07  & 84.74 ± 0.21 &  Reported &91.00 ±  0.01 & 89.5 ± 0.05 & 96.4 ± 0.00  \\
  Ours & \textbf{54.96 ± 3.18}& \textbf{29.57 ± 2.90} & {\bf 49.90 ± 7.23}  & \textbf{84.85 ± 0.07}&  Ours & \textbf{95.08 ± 0.33} &  \textbf{97.06 ± 0.22} & \textbf{97.47 ± 0.08}\\
 \bottomrule
\end{tabular}
 \label{table:compare_reported}
 \end{adjustbox}
 % \vspace{-0.2in}
\end{table}

\noindent {\bf \underline{Observation 2}: Divergence from Reported Results on ogbl-ddi.} We observe that our results in Table~\ref{table:ogb} for ogbl-ddi differ from the reported results. Outside of GCN, which reports better performance, most other GNN-based methods report a lower-than-reported performance. For example, for BUDDY we only achieve a Hits@20 of 29.60 vs. the reported 78.51 (see Appendix~\ref{sec:app_reportVSour_ddi} for a comprehensive comparison among methods). We find that the reason for this difference depends on the method. BUDDY~\cite{chamberlain2022graph} reported \footnote{https://github.com/melifluos/subgraph-sketching} using 6 negatives per positive sample during training, leading to an increase in performance.  Neo-GNN~\cite{yun2021neo} first pretrains the GNN  under the link prediction task, and then uses the pretrained model as the initialization for Neo-GNN.\footnote{https://github.com/seongjunyun/Neo-GNNs} For a fair comparison among methods, we only use 1 negative per positive sample in training and we don't apply the pretraining. 
% \harry{What about PEG?} \jh{we use }
For other methods, we find that a weak relationship between the validation and test performance complicates the tuning process, making it difficult to find the optimal hyperparameters. Please see Appendix~\ref{sec:app_ddi_investigate} for a more in-depth study and discussion.  

\noindent {\bf \underline{Observation 3}: High Model Standard Deviation.} The results in Tables~\ref{table:small} and \ref{table:ogb} present the mean performance and standard deviation when training over 10 seeds. Generally, we find that for multiple datasets the standard deviation of the ranking metrics is often high for most models. For example, the standard deviation for MRR can be as high as $8.82$, $8.96$, or $7.75$ for Cora, Citeseer, and Pubmed, respectively. Furthermore, on ogbl-ddi the standard deviation of Hits@20 reaches as high as 10.47 and 15.56. 
A high variance indicates unstable model performance. This makes it difficult to compare results between methods as the true performance lies in a larger range. This further complicates replicating model performance, as even large differences with the reported results may still fall within variance (see observation 2). 
Later in Section~\ref{sec:new_eval_results} we find that our new evaluation can reduce the model variance for all datasets (see Table~\ref{table:variance}). This suggests that the high variance is related to the current evaluation procedure.

\noindent {\bf \underline{Observation 4}: Inconsistency of AUC vs. Ranking-Based Metrics.}  
The AUC score is widely adopted to evaluate recent advanced link prediction methods~\cite{kipf2016variational, zhu2021neural}. However, from our results in Tables~\ref{table:small} and \ref{table:ogb} we observe  that there exists a disparity between AUC and ranking-based metrics. 
In some cases, the AUC score can be high  when the ranking metric is very low or even 0. For example, the Shortest Path heuristic records a Hits@K of 0 on ogbl-ppa. However, the AUC score  is $>99\%$. Furthermore, even though RA records the third and fifth best performance on ogbl-ppa and ogbl-collab, respectively, it has a lower AUC score than Shortest Path on both. Previous works~\cite{huang2023link, yang2015evaluating} argued that AUC is not a proper metric for link prediction. This is due to the inapplicability of AUC for highly imbalanced problems~\cite{davis2006relationship, saito2015precision}.

\section{New Evaluation Setting} \label{sec:new_eval}

In this section, we introduce a new setting for evaluating link prediction methods. 
We first discuss the unrealistic nature of the current evaluation setting in Section~\ref{sec:new_eval_problem}. Based on this, we present our new evaluation setting in Section~\ref{sec:new_eval_procedure}, which aims to align better with real-world scenarios. Lastly, in Section~\ref{sec:new_eval_results}, we present and discuss the results based on our new evaluation setting.

\subsection{Issues with the Existing Evaluation Setting} \label{sec:new_eval_problem}

The existing evaluation procedure for link prediction is to rank a positive sample against a set of $K$ randomly selected negative samples. The same set of $K$ negatives are used for all positive samples (with the exception of ogbl-citation2 which uses 1000 per positive sample). We demonstrate that there are multiple issues with this setting, making it difficult to properly evaluate the effectiveness of current models.

\noindent {\bf \underline{Issue 1}: Non-Personalized Negative Samples.} The existing evaluation setting uses the same set of negative samples for all positive samples (outside of ogbl-citation2). This strategy, referred to as global negative sampling~\cite{wang2021pairwise}, is not a commonly sought objective.  Rather, we are often more interested in predicting links that will occur for a specific node. Take, for example, a social network that connects users who are friends.  In this scenario, we may be interested in recommending new friends to a user $u$. This requires learning a classifier $f$ that assigns a probability to a link existing. When evaluating this task, we want to rank links where $u$ connects to an existing friend above those where they don't. For example, if $u$ is friends with $a$ but not $b$, we hope that $f(u, a) > f(u, b)$. However, the existing evaluation setting doesn't explicitly test for this. Rather it compares a true sample $(u, a)$ with a potentially unrelated negative sample, e.g., $(c, d)$. This is not aligned with the real-world usage of link prediction on such graphs.

\noindent {\bf \underline{Issue 2}: Easy Negative Samples.} The existing evaluation setting randomly selects negative samples to use. However given the large size of most graphs (see Table~\ref{table:app_data} in Appendix~\ref{sec:app_data_parameter}), randomly sampled negatives are likely to choose two nodes that bear no relationship to each other. Such node pairs are trivial to classify. We demonstrate this by plotting the distribution of common neighbors (CN), a strong heuristic, for all positive and negative test samples in Figure~\ref{fig:cn_ogb}. Almost all the negative samples contain no CNs, making them easy to classify. We further show that the same problem afflicts even the smaller datasets in Figure~\ref{fig:app_cn_small} in Appendix~\ref{sec:app_cn_small}.

% To illustrate the issue, we hark back to the example in Section~\ref{sec:intro}. We are given a social network where a link indicates two users are friends. Our goal is to learn a classifier $f$ to predict links between any two users. This can then be utilized to recommend new friends to users. Let's say that we know a user {\it John} is friends with {\it Peter} but not {\it Sarah}. In such a case, we hope to give a higher probability to the true link such that $f(\textit{John}, \textit{Peter}) > f(\textit{John}, \textit{Sarah})$. However the current evaluation method doesn't explicitly test for this case. Instead, it compares $(\textit{John}, \textit{Peter})$ against a set of randomly sampled negatives that may not even include {\it John}. For example, $f(\textit{John}, \textit{Peter}) > f(\textit{Rachel}, \textit{Joseph})$. As such, our model is not guaranteed to assign a higher probability to $(\textit{John}, \textit{Peter})$ than $(\textit{John}, \textit{Sarah})$. This is problematic in a real world setting when recommending potential friends to {\it John} as we may end up recommending {\it Sarah} over {\it Peter} \ym{why this will be the case?}. 

These observations suggest that a more realistic evaluation strategy is desired. At the core of this challenge is which negative samples to use during evaluation. We discuss our design for solving this in the next subsection.

\subsection{Heuristic  Related Sampling  Technique (HeaRT)} \label{sec:new_eval_procedure}

\begin{figure*}[t]
\begin{center}
    
     \centerline{
        {\subfigure[Negative sample generation for one positive sample.]
        {\includegraphics[width=0.25\linewidth]{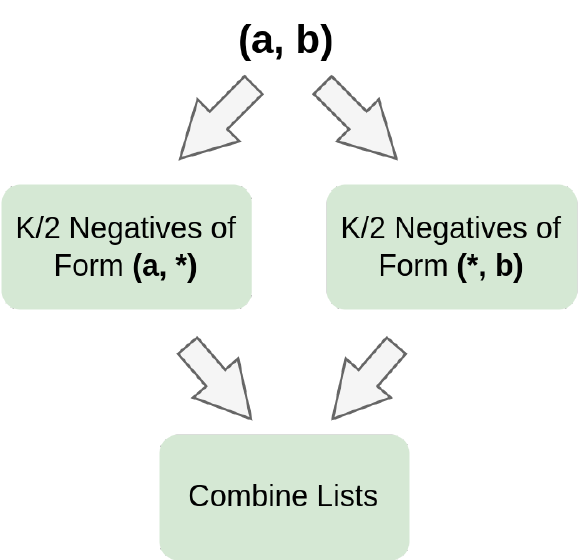} }
        \label{fig:gen_hard_negatives_1}
        }
        
        {\subfigure[Process of determining negative samples that contain a node $a$.]
        {\includegraphics[width=0.64\linewidth]{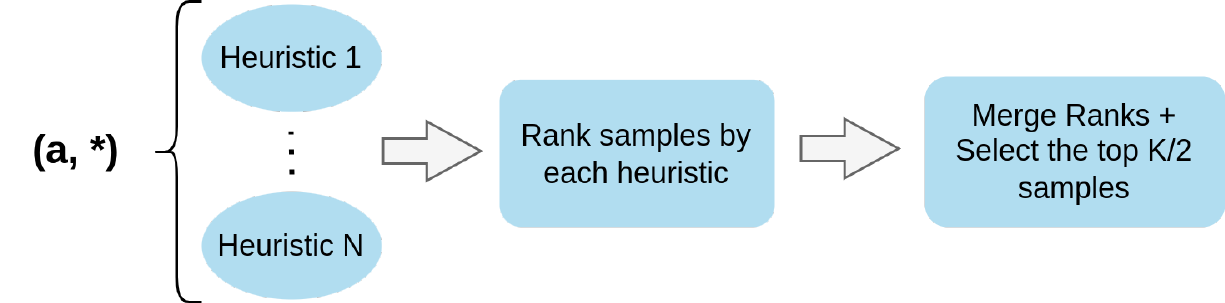} }
        \label{fig:gen_hard_negatives_2}
        }
    }   
    % \vspace{-0.05in}
    \caption{Pipeline for generating the hard negative samples for a positive sample (a, b).}
\label{fig:gen_hard_negatives}
\end{center}
% \vspace{-0.4in}
\end{figure*}

In this subsection, we introduce new strategy for evaluating link prediction methods. To address the concerns outlined in Section~\ref{sec:new_eval_problem}, we design a new method for sampling negatives during evaluation. Our strategy, HeaRT, solves these challenges by: (a) personalizing the negatives to each sample and (b) using heuristics to select hard negative samples. This allows for the negative samples to be directly related to each positive sample while also being non-trivial. We further discuss how to ensure that the negative samples are  both {\it personalized} and {\it non-trivial} for a specific positive sample.

From our discussion in Section~\ref{sec:new_eval_problem}, we are motivated in personalizing the negatives to each positive sample. Since the positive samples in the current datasets are node pairs, we seek to personalize the negatives to both nodes in the positive sample. Extending our example in Section~\ref{sec:new_eval_problem}, this is analogous to restricting the negatives to contain one of the two users from the original friendship pair. As such, for a positive sample $(u, a)$, the negative samples  will belong to the set: 
\begin{equation}   \label{eq:candidate_set}
    S(u, a) = \{(u', a) \: | \: u' \in \mathcal{V}\} \cup \{(u, a') \: | \: a' \in \mathcal{V}\},
\end{equation}
where $\mathcal{V}$ is the set of nodes. This is similar to the setting used for knowledge graph completion (KGC)~\cite{bordes2013translating} which uses all such samples for evaluation. However, one drawback of evaluating each positive sample against the entire set of possible corruptions is the high computational cost. To mitigate this issue we consider only utilizing a small subset of $S(u, a)$ during evaluation. 

{\it The key challenge is how to generate a subset of $S(u, a)$}. If we randomly sample from $S(u, a)$, we risk only utilizing easy negative samples. This is one of the issues of the existing evaluation setting (see Issue 2 in Section~\ref{sec:new_eval_problem}), whereby randomly selecting negatives, they unknowingly produce negative samples that are too easy. We address this by selecting the negative samples via a combination of multiple heuristics. Since heuristics typically correlate well with performance, we ensure that the negative samples will be non-trivial to classify. This is similar to the concept of candidate generation~\cite{gupta2013wtf, eksombatchai2018pixie}, which only ranks a subset of candidates that are most likely to be true.

An overview of the generation process is given in Figure~\ref{fig:gen_hard_negatives}. For each positive sample, we generate $K$ negative samples. To allow personalization to both nodes in the positive sample equally, we sample $K/2$ negatives with each node.  For the heuristics, we consider RA~\cite{zhou2009predicting}, PPR~\cite{pagerank}, and feature similarity. A more detailed discussion on the negative sample generation is given in Appendix~\ref{sec:appendix_neg_sample}.
{It's important to note that our work centers specifically on negative sampling during the evaluation stage (validation and test). This is distinct from prior work that concerns the negatives sampled used during the training phase~\cite{yang2020understanding,rendle2009bpr}. As such, the training process remains unaffected under both the existing and HeaRT setting.
% Specifically, in these papers~\cite{yang2020understanding,rendle2009bpr}, they study the influence of negative sampling on the training optimization objective,  and focus on how to use negative sampling to better train methods. 
% While  we focus on how to use negative sampling to better evaluate methods.
% The training process remains unaffected under both the existing and HeaRT setting.}
}

\begin{table}[]
\centering
% \footnotesize
 \caption{Results on Cora, Citeseer, and Pubmed (\%) under HeaRT. Highlighted are the results ranked \colorfirst{first}, \colorsecond{second}, and  \colorthird{third}.}
 \begin{adjustbox}{width =1 \textwidth}
\begin{tabular}{cc|cccccc}

\toprule
 &\multirow{2}{*}{Models} & \multicolumn{2}{c}{Cora} &\multicolumn{2}{c}{Citeseer}  &\multicolumn{2}{c}{Pubmed}   \\ 

 & &MRR &Hits@10 &MRR &Hits@10 &MRR &Hits@10 \\
  \midrule
\multirow{5}{*}{Heuristic}&CN            & {9.78}  & {20.11} & {8.42}  &{18.68} & {2.28} & {4.78} \\
&AA            &{11.91} & {24.10}  & {10.82} & {22.20}  &{2.63} & {5.51} \\
&RA            &{11.81} & {24.48} & {10.84} &{22.86} & {2.47} & {4.9}  \\
&Shortest Path & {5.04} & {15.37} & {5.83}  & {16.26} &{0.86} &{0.38} \\
&Katz          & {11.41} &{22.77} &{11.19} & {24.84} & {3.01} &{5.98} \\

\midrule
\multirow{3}{*}{Embedding}&Node2Vec      & 14.47 ± 0.60               & 32.77 ± 1.29              & 21.17 ± 1.01              & 45.82 ± 2.01              & 3.94 ± 0.24              & 8.51 ± 0.77              \\
&MF            & 6.20 ± 1.42                & 15.26 ± 3.39              & 7.80 ± 0.79                & 16.72 ± 1.99              & 4.46 ± 0.32              & 9.42 ± 0.87              \\
&MLP           & 13.52 ± 0.65              & 31.01 ± 1.71              & 22.62 ± 0.55              & 48.02 ± 1.79              & 6.41 ± 0.25              &  15.04 ± 0.67             \\
\midrule
\multirow{4}{*}{GNN}&GCN           & \colorsecond{16.61 ± 0.30}               & \colorthird{36.26 ± 1.14}              & 21.09 ± 0.88              & 47.23 ± 1.88              & 7.13 ± 0.27              & 15.22 ± 0.57             \\
&GAT           & 13.84 ± 0.68              & 32.89 ± 1.27              & 19.58 ± 0.84              & 45.30 ± 1.3                & 4.95 ± 0.14              & 9.99 ± 0.64              \\
&SAGE          & 14.74 ± 0.69              & 34.65 ± 1.47              & 21.09 ± 1.15              & 48.75 ± 1.85              & \colorfirst{9.40 ± 0.70}       & \colorfirst{20.54 ± 1.40}     \\
&GAE           & \colorfirst{18.32 ± 0.41}     & \colorfirst{37.95 ± 1.24}     & \colorsecond{25.25 ± 0.82}              & \colorthird{49.65 ± 1.48}              & 5.27 ± 0.25              & 10.50 ± 0.46              \\
\midrule
\multirow{7}{*}{GNN+Pairwise Info}&SEAL          & 10.67 ± 3.46              & 24.27 ± 6.74              & 13.16 ± 1.66              & 27.37 ± 3.20               & 5.88 ± 0.53              & 12.47 ± 1.23             \\
&BUDDY         & 13.71 ± 0.59              & 30.40 ± 1.18              & 22.84 ± 0.36              & 48.35 ± 1.18              & 7.56 ± 0.18              & 16.78 ± 0.53             \\
&Neo-GNN       & 13.95 ± 0.39              & 31.27 ± 0.72              & 17.34 ± 0.84              & 41.74 ± 1.18              & \colorthird{7.74 ± 0.30}              & \colorthird{17.88 ± 0.71}             \\
&NCN           & 14.66 ± 0.95              & 35.14 ± 1.04              & \colorfirst{28.65 ± 1.21}     & \colorsecond{53.41 ± 1.46}              & 5.84 ± 0.22              & 13.22 ± 0.56             \\
&NCNC          & 14.98 ± 1.00               & \colorsecond{36.70 ± 1.57}               & \colorthird{24.10 ± 0.65}               & \colorfirst{53.72 ± 0.97}     & \colorsecond{8.58 ± 0.59}              & \colorsecond{18.81 ± 1.16}      \\  
&NBFNet        & 13.56 ± 0.58              & 31.12 ± 0.75              & 14.29 ± 0.80               & 31.39 ± 1.34              & >24h                        & >24h                        \\
&PEG           & \colorthird{15.73 ± 0.39}              & 36.03 ± 0.75              & 21.01 ± 0.77              & 45.56 ± 1.38              & 4.4 ± 0.41               & 8.70 ± 1.26               \\
    
 \bottomrule
\end{tabular}
 \label{table:small_newsetting}
 \end{adjustbox}
 % \vspace{-0.3in}
\end{table}

\subsection{Results and Discussion} \label{sec:new_eval_results}

In this subsection we present our results  when utilizing HeaRT. We follow the parameter ranges introduced in Section~\ref{sec:prelim_exp}. For all datasets we use $K=500$ negative samples per positive sample during evaluation. Furthermore for ogbl-ppa we only use a small subset of the validation and test positive samples (100K each) for evaluation. This is because the large size of the validation and test sets (see Table~\ref{table:app_data} in Appendix~\ref{sec:app_data_parameter}) makes HeaRT prohibitively expensive.

The results are shown in Table~\ref{table:small_newsetting} (Cora, Citeseer, Pubmed) and  Table~~\ref{table:ogb_newsetting} (OGB). For simplicity, we only include the MRR and Hits@10 for Cora, Citeseer, Pubmed, and the MRR and Hits@20 for OGB. Additional results for other metrics can be found in Appendix~\ref{sec:app_new_more}.
We note that most datasets, outside of ogbl-ppa, exhibit much lower performance than under the existing setting. This is though we typically use much fewer negative samples in the new setting, implying that the negative samples produced by HeaRT are much harder.
We highlight the  main observations below.

\noindent {\bf \underline{Observation 1}: Better Performance of Simple  Models}. We find that under HeaRT, ``simple" baseline models (i.e.,  heuristic, embedding, and GNN methods) show a greater propensity to outperform their counterparts via ranking metrics than under the existing setting. Specifically, we focus on MRR in  Table~\ref{table:small},~\ref{table:small_newsetting}, and ~\ref{table:ogb_newsetting}, and the corresponding  ranking-based metrics in Table~\ref{table:ogb}.
Under the existing setting, such methods only rank in the top three for any dataset a total of {\bf 5} times. However, under HeaRT this occurs {\bf 10} times. Furthermore, under the existing setting only {\bf 1} ``simple" method ranks best overall while under HeaRT there are {\bf 3}. This suggests that recent advanced methods may have benefited from the easier negative samples in the existing setting.  

\begin{table}[]
\centering
% \footnotesize
 \caption{Results on OGB datasets (\%) under HeaRT. Highlighted are the results ranked \colorfirst{first}, \colorsecond{second}, and  \colorthird{third}.}
 \begin{adjustbox}{width =1 \textwidth}
\begin{tabular}{c|cccccccc}

\toprule
 \multirow{2}{*}{Models} & \multicolumn{2}{c}{ogbl-collab} &\multicolumn{2}{c}{ogbl-ddi}  &\multicolumn{2}{c}{ogbl-ppa} &    \multicolumn{2}{c}{ogbl-citation2}  \\ 
 
  &MRR &Hits@20 &MRR &Hits@20 &MRR &Hits@20 &MRR & Hits@20\\
 \midrule

CN & {4.20} & {16.46} & 6.71 &38.69 & 25.70  &68.25 & 17.11 &41.73 \\
AA & {5.07} & {19.59} &{6.97} & {39.75} &{26.85} &{70.22} &{17.83}  & {43.12} \\
RA & \colorthird{6.29} & \colorsecond{24.29} &{8.70} & {44.01}       &{28.34} &{71.50}  & 17.79	&43.34 \\
Shortest Path & {2.66} & {15.98} &{0} & {0}  &{0.54}  & {1.31}  & >24h & >24h \\
Katz &\colorsecond{6.31} & {\colorfirst{24.34}} & 6.71&38.69 & 25.70&68.25 & 14.10 & 35.55 \\ \midrule
Node2Vec & 4.68 ± 0.08 & 16.84 ± 0.17 & 11.14 ± 0.95 & 63.63 ± 2.05 & 18.33 ± 0.10  & 53.42 ± 0.11 & 14.67 ± 0.18 & 42.68 ± 0.20 \\
MF & 4.89 ± 0.25 & 18.86 ± 0.40 & \colorfirst{13.99 ± 0.47} & 59.50 ± 1.68 & 22.47 ± 1.53 & 70.71 ± 4.82  & 8.72 ± 2.60 & 29.64 ± 7.30 \\
MLP & 5.37 ± 0.14 & 16.15 ± 0.27 & N/A & N/A  & 0.98 ± 0.00 & 1.47 ± 0.00& 16.32 ± 0.07 & 43.15 ± 0.10  \\ \midrule
GCN & {6.09 ± 0.38} &{22.48 ± 0.81} & \colorsecond{13.46 ± 0.34} & 64.76 ± 1.45 & 26.94 ± 0.48 & 68.38 ± 0.73 & 19.98 ± 0.35 & \colorthird{51.72 ± 0.46} \\
GAT & 4.18 ± 0.33 & 18.30 ± 1.42 & \colorthird{12.92 ± 0.39} & \colorsecond{66.83 ± 2.23}  & OOM & OOM  & OOM  & OOM     \\
SAGE & 5.53 ± 0.5 & 21.26 ± 1.32 & 12.60 ± 0.72 & \colorfirst{67.19 ± 1.18} & 27.27 ± 0.30 & 69.49 ± 0.43  & \colorsecond{22.05 ± 0.12} & \colorsecond{53.13 ± 0.15}\\
GAE & {OOM} &{OOM}&{3.49 ± 1.73} & {17.81 ± 9.80} & OOM & OOM  & OOM & OOM            \\ \midrule
SEAL & \colorfirst{6.43 ± 0.32} & 21.57 ± 0.38 & 9.99 ± 0.90 & 49.74 ± 2.39 & \colorthird{29.71 ± 0.71}& \colorthird{76.77 ± 0.94} & \colorthird{20.60 ± 1.28} & 48.62 ± 1.93 \\
BUDDY & 5.67 ± 0.36 & \colorthird{23.35 ± 0.73}  & 12.43 ± 0.50 & 58.71 ± 1.63  & 27.70 ± 0.33 & 71.50 ± 0.68 & 19.17 ± 0.20  & 47.81 ± 0.37   \\
Neo-GNN & 5.23 ± 0.9 & 21.03 ± 3.39 & 10.86 ± 2.16 & 51.94 ± 10.33 & 21.68 ± 1.14 & 64.81 ± 2.26  & 16.12 ± 0.25   & 43.17 ± 0.53        \\
NCN & 5.09 ± 0.38 & 20.84 ± 1.31 & 12.86 ± 0.78 & \colorthird{65.82 ± 2.66} & \colorfirst{35.06 ± 0.26} & \colorsecond{81.89 ± 0.31}  & \colorfirst{23.35 ± 0.28}& \colorfirst{53.76 ± 0.20}      \\
NCNC & 4.73 ± 0.86 & 20.49 ± 3.97  & >24h    & >24h & \colorsecond{33.52 ± 0.26}   & \colorfirst{82.24 ± 0.40}  & 19.61 ± 0.54  & 51.69 ± 1.48       \\ 
NBFNet & OOM  & OOM     & >24h   & >24h      & OOM        & OOM     & OOM    & OOM                       \\
PEG & 4.83 ± 0.21 & 18.29 ± 1.06 & 12.05 ± 1.14  & 50.12 ± 6.55  & OOM   & OOM    & OOM     & OOM                       \\
     
 \bottomrule
\end{tabular}
 \label{table:ogb_newsetting}
 \end{adjustbox}

% \vspace{-0.25in}
\end{table}

Another interesting observation is that on ogbl-collab, heuristic methods achieve strong performance and are able to outperform  more complicated models. Specifically, we find that Katz is ranked the second and RA the third. Note that this underscores the significance of the common neighbors information (i.e., paths of length 2), as this information is incorporated in both RA and Katz. 
% Since Katz considers all paths between two nodes, it might benefit from the effective performance of methods based on common neighbors. 
Of note is that ogbl-collab is a dynamic collaboration graph, which is different from other datasets. Because of this, the negative sampling strategy also differs slightly from the other datasets. Please see Appendix~\ref{sec:collab_observation} for more discussion. 

\noindent {\bf \underline{Observation 2}: Lower Model Standard Deviation}. We observed earlier that, under the existing evaluation setting, the model variance across seeds was high (see observation 3 in Section~\ref{sec:benchmark}). This complicates model comparison as the model performance is unreliable. Interestingly, we find that HeaRT is able to dramatically reduce the variance for all datasets. We demonstrate this by first calculating the mean standard deviation across all models on each individual dataset. This was done for both evaluation settings with the results compared. As demonstrated in Table~\ref{table:variance}, the mean standard deviation decreases for all datasets. This is especially true for Cora, Citeseer, and Pubmed, which each decrease by over 85\%. 
Such a large decrease in standard deviation is noteworthy as it allows for a more trustworthy and reliable comparison between methods.

\begin{wraptable}{r}{0.5\textwidth}
% \adjustbox{width =1 \textwidth}
\centering
\footnotesize

 \caption{Mean model standard deviation for the existing  setting and HeaRT. We use Hits@20 for ogbl-ddi, Hits@50 for ogbl-collab, Hits@100 for ogbl-ppa, and MRR otherwise.}
 
\begin{tabular}{l | ccc}
\toprule
 Dataset & Existing  & HeaRT & \% Change \\ 
 \midrule
  Cora & 5.19 & 0.79 & -85\%\\
  Citeseer & 5.94 & 0.88 & -85\% \\
  Pubmed & 4.14  & 0.35 & -92\% \\
  % \midrule
  ogbl-collab & 1.49 & 0.96 & -36\% \\
  ogbl-ppa & 2.13 & 0.36 & -83\% \\
  ogbl-ddi & 6.77 & 3.49 & -48\% \\
  ogbl-citation2 & 1.39 & 0.59 & -58\% \\
 \bottomrule
\end{tabular}
 \label{table:variance}
% \end{table}
 % \end{adjustbox}
\vspace{-0.1in}
\end{wraptable}

% \begin{wraptable}{r}{0.5\textwidth}
% % \adjustbox{width =1 \textwidth}
% \centering
% \footnotesize

%  \caption{Mean model standard deviation for the existing  setting and HeaRT. We use Hits@20 for ogbl-ddi, Hits@50 for ogbl-collab, Hits@100 for ogbl-ppa, and MRR otherwise.}
 
% \begin{tabular}{l | cccc}
% \toprule
%  Model & GCN  & BUDDY & Neo-GNN &NCN \\ 
%  \midrule
%   Reported &37.07	&78.51&	63.57	&82.32\\
%   Ours &49.9	&29.6	&20.95	&76.52 \\

%  \bottomrule
% \end{tabular}
%  \label{table:variance}
% % \end{table}
%  % \end{adjustbox}
% % \vspace{-0.1in}
% \end{wraptable}

We posit that this observation is caused by a stronger alignment between the positive and negative samples under our new evaluation setting. Under the existing evaluation setting, the same set of negative samples is used for all positive samples. One consequence of this is that a single positive sample may bear little to no relationship to the negative samples (see Section~\ref{sec:new_eval_problem} for more discussion). However, under our new evaluation setting, the negatives for a positive sample are a subset of the corruptions of that sample. This allows for a more natural comparison via ranking-based metrics as the samples are more related and can be more easily compared.

\noindent {\bf \underline{Observation 3}: Lower Model Performance}.  We observe that the majority of datasets exhibit a significantly reduced performance in comparison to the existing setting. For example, under the existing setting, models typically achieve a MRR of around 30, 50, and 30 on Cora, Citeseer, and Pubmed (Table~\ref{table:small}), respectively. However, under HeaRT the MRR for those datasets is typically around 20, 25, and 10 (Table~\ref{table:small_newsetting}). 
Furthermore for ogbl-citation2, the MRR of the best performing model falls from a shade under 90 on the existing setting to slightly over 20 on HeaRT. 
Lastly, we note that the performance on ogbl-ppa actually increases. This is because we only utilize a small subset of the total test set when evaluating on HeaRT, nullifying any comparison between the two settings.

These outcomes are observed despite HeaRT using much fewer negative samples than the original setting. This suggests that the negative samples generated by HeaRT are substantially more challenging than those used in the existing setting. This underscores the need to develop more advanced methodologies that can tackle harder negatives samples like in HeaRT.

\section{Conclusion}
In this work we have revealed several pitfalls found in recent works on link prediction. 
To overcome these pitfalls, we first establish a benchmarking that facilitates a fair and consistent evaluation across a diverse set of models and datasets.
By doing so, we are able to make several illuminating observations about the performance and characteristics of various models. Furthermore, based on several limitations we observed in the existing evaluation procedure,
we introduce a more practical setting called HeaRT (Heuristic Related Sampling Technique). HeaRT incorporates a more real-world evaluation setting, resulting in a better comparison among methods. By introducing a more rigorous and realistic assessment, HeaRT could guide the field towards more effective models, thereby advancing the state of the art in link prediction.

\section{Acknowledgements}
This research is supported by the National Science Foundation (NSF) under grant numbers CNS 2246050, IIS1845081, IIS2212032, IIS2212144, IOS2107215, DUE 2234015, III-2212145, III-2153326, DRL2025244 and IOS2035472, the Army Research Office (ARO) under grant number W911NF-21-1-0198, the Home Depot, Cisco Systems Inc, Amazon Faculty Award, Johnson\&Johnson, JP Morgan Faculty Award and SNAP.

\bibliographystyle{unsrt}
\bibliography{bibtex}
\newpage
\section{Checklist}
\begin{enumerate}

\item For all authors...
\begin{enumerate}
  \item Do the main claims made in the abstract and introduction accurately reflect the paper's contributions and scope?
    \answerYes{}
  \item Did you describe the limitations of your work?
    \answerYes{See Appendix~\ref{sec:app_limitation}.}
  \item Did you discuss any potential negative societal impacts of your work?
    \answerYes{See Appendix~\ref{sec:app_social_impact}.}
  \item Have you read the ethics review guidelines and ensured that your paper conforms to them?
    \answerYes{}
\end{enumerate}

\item If you are including theoretical results...
\begin{enumerate}
  \item Did you state the full set of assumptions of all theoretical results?
    \answerNA{}
	\item Did you include complete proofs of all theoretical results?
    \answerNA{}
\end{enumerate}

\item If you ran experiments (e.g. for benchmarks)...
\begin{enumerate}
  \item Did you include the code, data, and instructions needed to reproduce the main experimental results (either in the supplemental material or as a URL)?
    \answerYes{See \url{https://github.com/Juanhui28/HeaRT}.}
  \item Did you specify all the training details (e.g., data splits, hyperparameters, how they were chosen)?
    \answerYes{Yes. See Section~\ref{sec:setting} and Appendix~\ref{sec:app_data_parameter}.}
	\item Did you report error bars (e.g., with respect to the random seed after running experiments multiple times)?
    \answerYes{}
	\item Did you include the total amount of compute and the type of resources used (e.g., type of GPUs, internal cluster, or cloud provider)?
    \answerYes{See Appendix~\ref{sec:param_settings}.}
\end{enumerate}

\item If you are using existing assets (e.g., code, data, models) or curating/releasing new assets...
\begin{enumerate}
  \item If your work uses existing assets, did you cite the creators?
    \answerYes{}
  \item Did you mention the license of the assets?
    \answerYes{See Appendix~\ref{sec:code}.}
  \item Did you include any new assets either in the supplemental material or as a URL?
    \answerYes{See \url{https://github.com/Juanhui28/HeaRT}.}
  \item Did you discuss whether and how consent was obtained from people whose data you're using/curating?
    \answerYes{}
  \item Did you discuss whether the data you are using/curating contains personally identifiable information or offensive content?
    % \answerYes{}
    \answerNA{}
\end{enumerate}

\item If you used crowdsourcing or conducted research with human subjects...
\begin{enumerate}
  \item Did you include the full text of instructions given to participants and screenshots, if applicable?
    % \answerYes{}
    \answerNA{}
  \item Did you describe any potential participant risks, with links to Institutional Review Board (IRB) approvals, if applicable?
    % \answerYes{}
    \answerNA{}
  \item Did you include the estimated hourly wage paid to participants and the total amount spent on participant compensation?
   % \answerYes{}
   \answerNA{}
\end{enumerate}

\end{enumerate}

\appendix

% \section{Appendix}
\section{Common Neighbor Distribution}
\label{sec:app_cn_small}

In Figure~\ref{fig:cn_ogb} we demonstrate the common neighbor (CN) distribution among positive and negative test samples for ogbl-collab, ogbl-ppa, and ogbl-citation2. These results demonstrate that a vast majority of negative samples have no CNs. Since CNs is a typically good heuristic, this makes it easy to identify most negative samples. 

\begin{figure*}[t]
    \begin{center}
     \centerline{
        {\subfigure[Cora]
        {\includegraphics[width=0.35\linewidth]{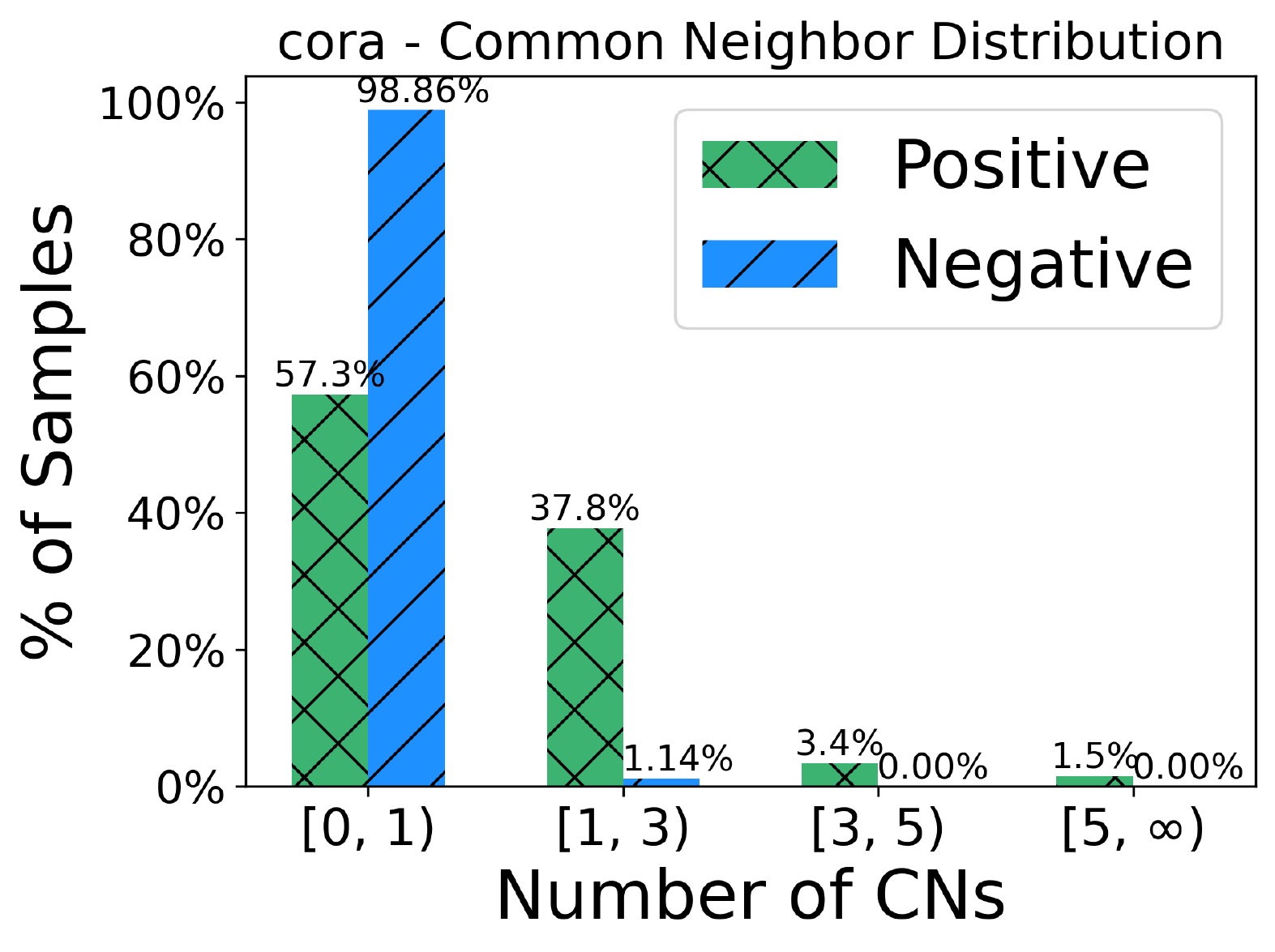} }
        
        }
        
        {\subfigure[Citeseer]
        {\includegraphics[width=0.35\linewidth]{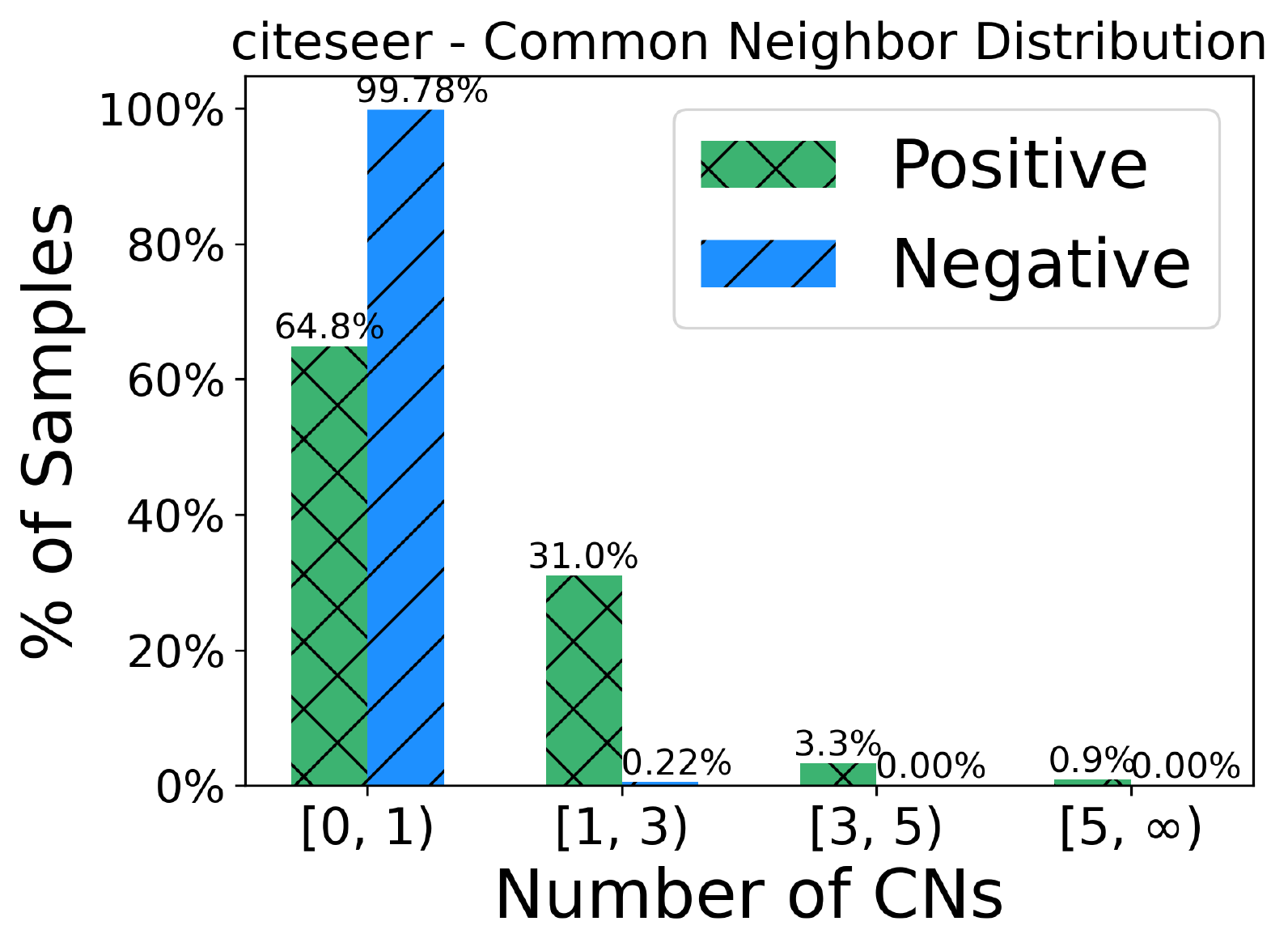} }}
        }
         \centerline{
        {\subfigure[Pubmed]
        {\includegraphics[width=0.35\linewidth]{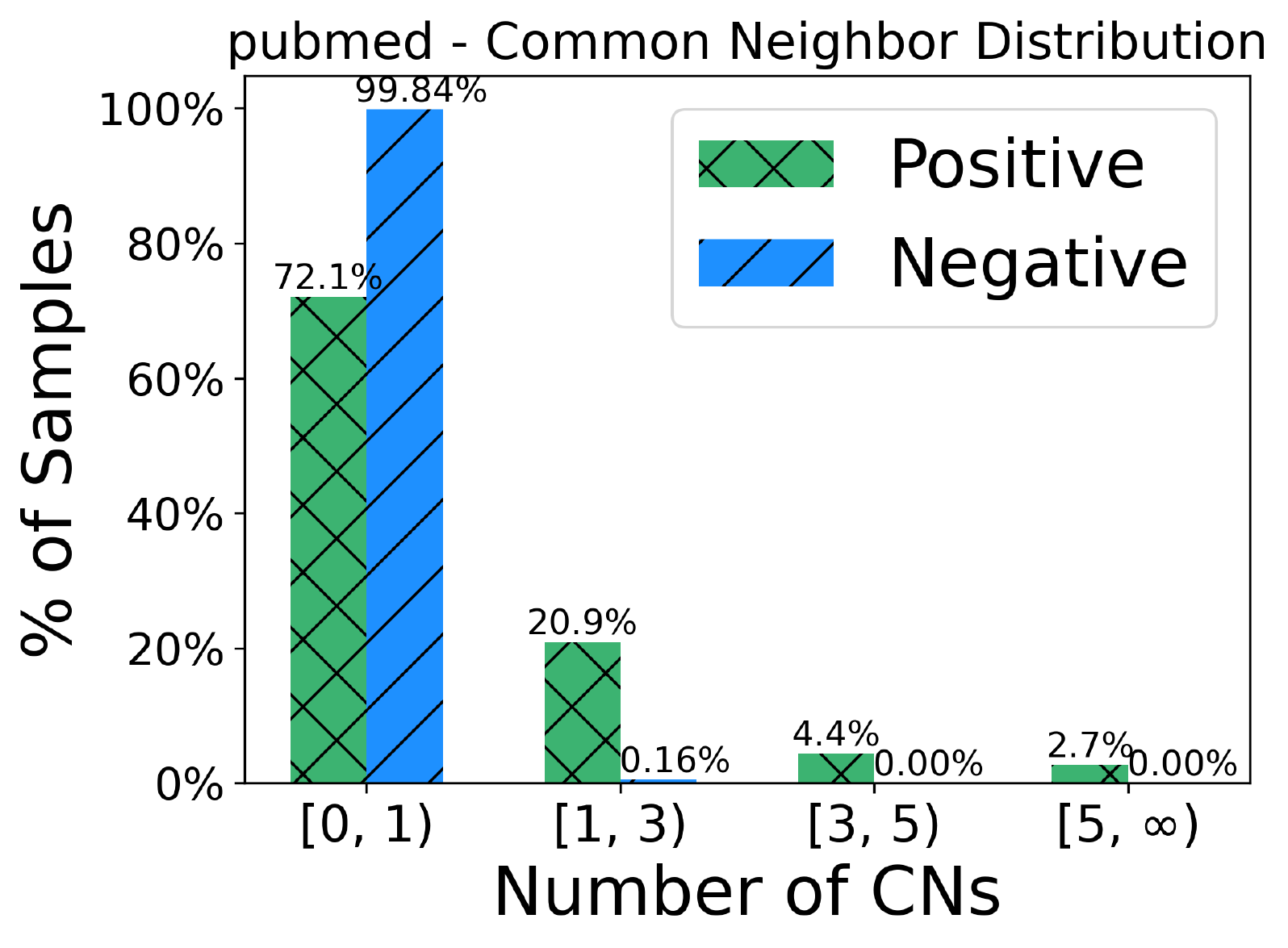} }}

         {\subfigure[ogbl-ddi]
        {\includegraphics[width=0.35\linewidth]{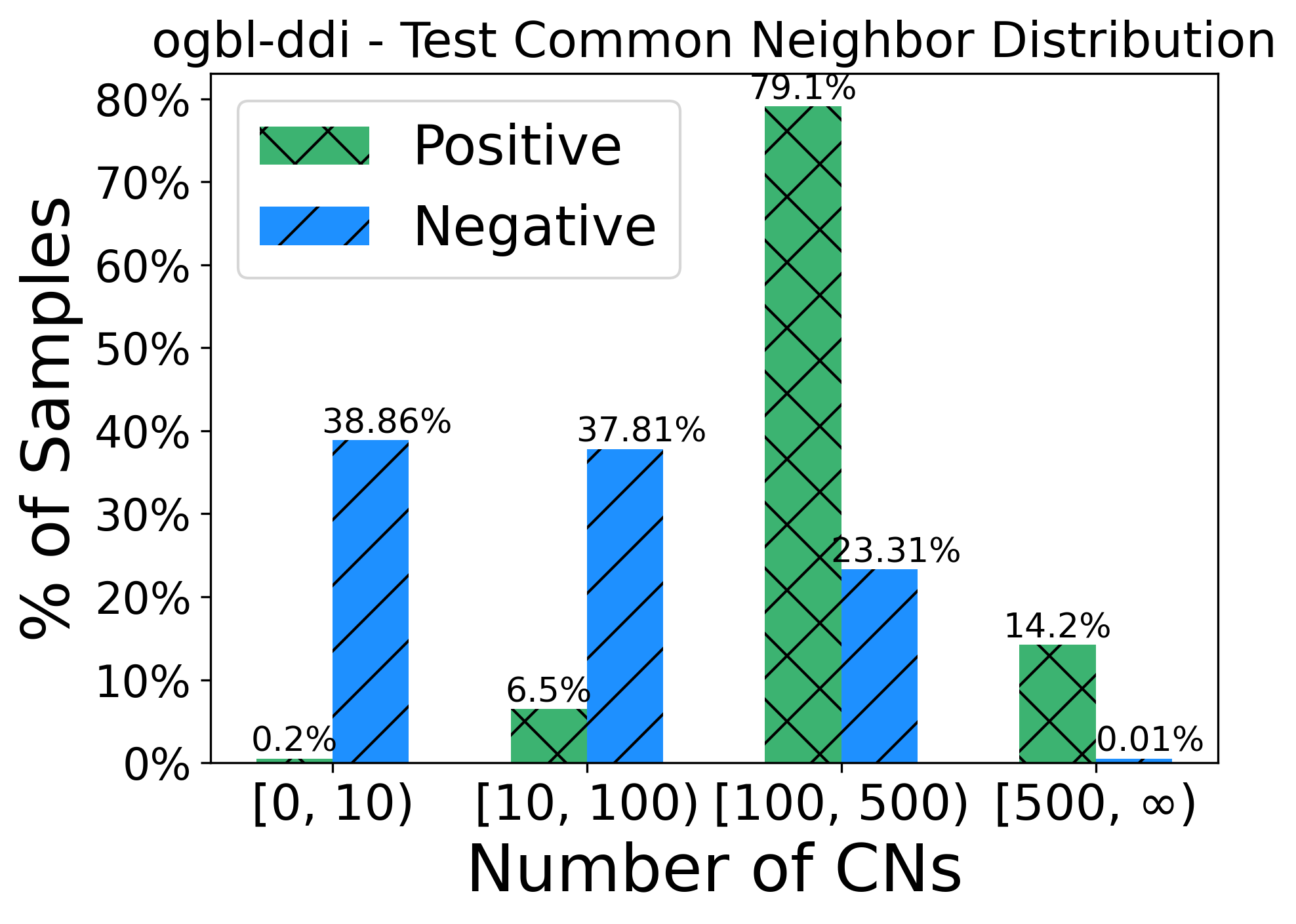} }}
    }
    
    \caption{Common neighbor distribution for the  positive and negative test samples for the Cora,  Citeseer,  Pubmed, and ogbl-ddi under the existing evaluation setting. 
    }

\label{fig:app_cn_small}
\end{center}

\end{figure*}

We further present the CN distribution of Cora, Citeseer, Pubmed, and ogbl-ddi in \figurename~\ref{fig:app_cn_small}. The CN distribution of Cora,  Citeseer, and Pubmed are consistent with our previous observations on the OGB datasets in Figure~\ref{fig:cn_ogb}. We note that ogbl-ddi exhibits a different distribution with other datasets. As compared to the other datasets, most of the negative samples in ogbl-ddi have common neighbors. This is likely because ogbl-ddi is considerably denser than the other graphs. As shown in Table~\ref{table:app_data}, the average node degree in ogbl-ddi is $625.68$, significantly larger than the second largest dataset ogbl-ppa with 105.25. Thus, despite the random sampling of negative samples, the high degree of node connectivity within the ogbl-ddi graph predisposes a significant likelihood for the occurrence of common neighbors.

\begin{figure*}[t]
    \begin{center}
     \centerline{
        {\subfigure[Cora]
        {\includegraphics[width=0.33\linewidth]{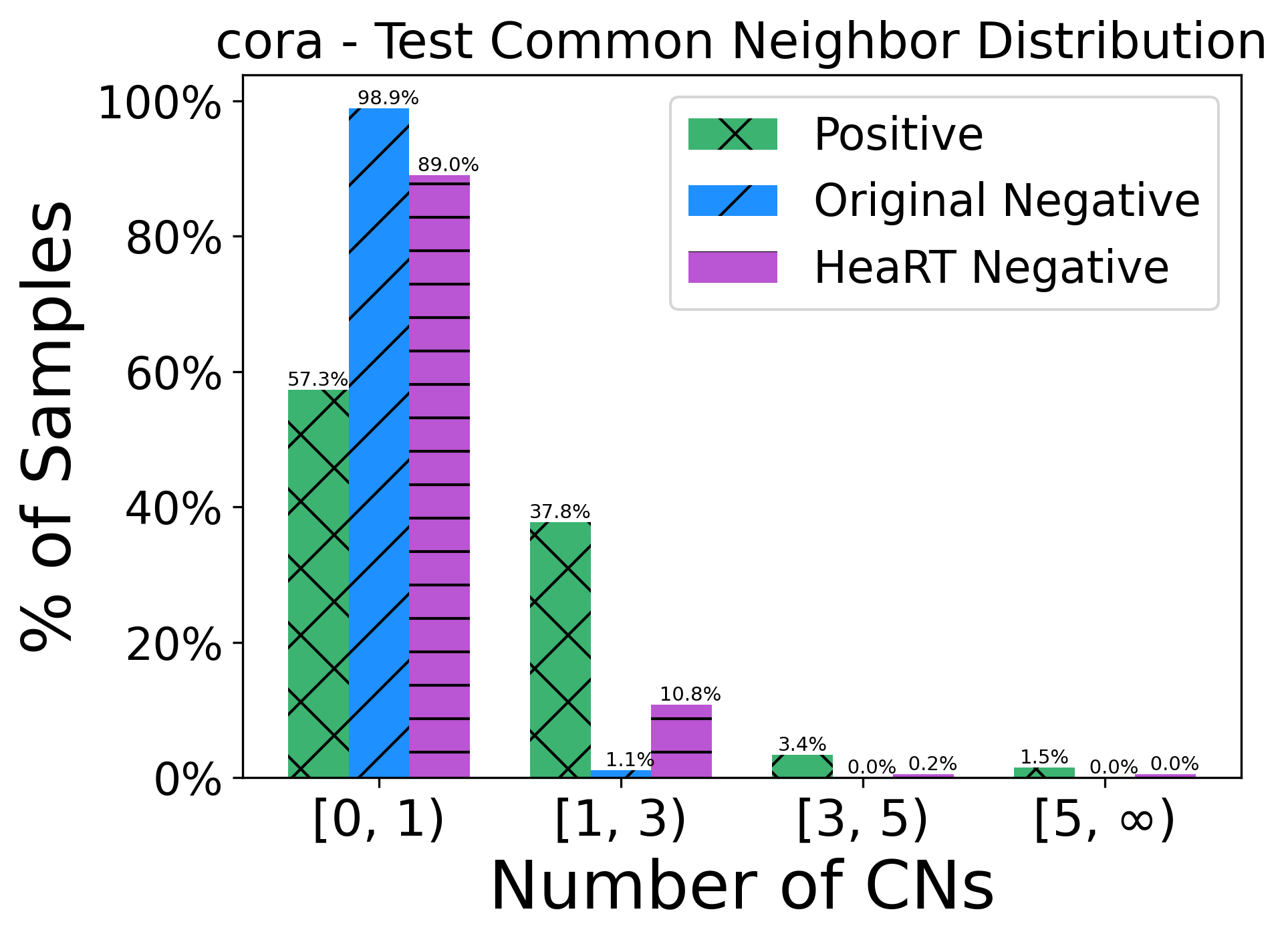} }}
        
        {\subfigure[Citeseer]
        {\includegraphics[width=0.345\linewidth]{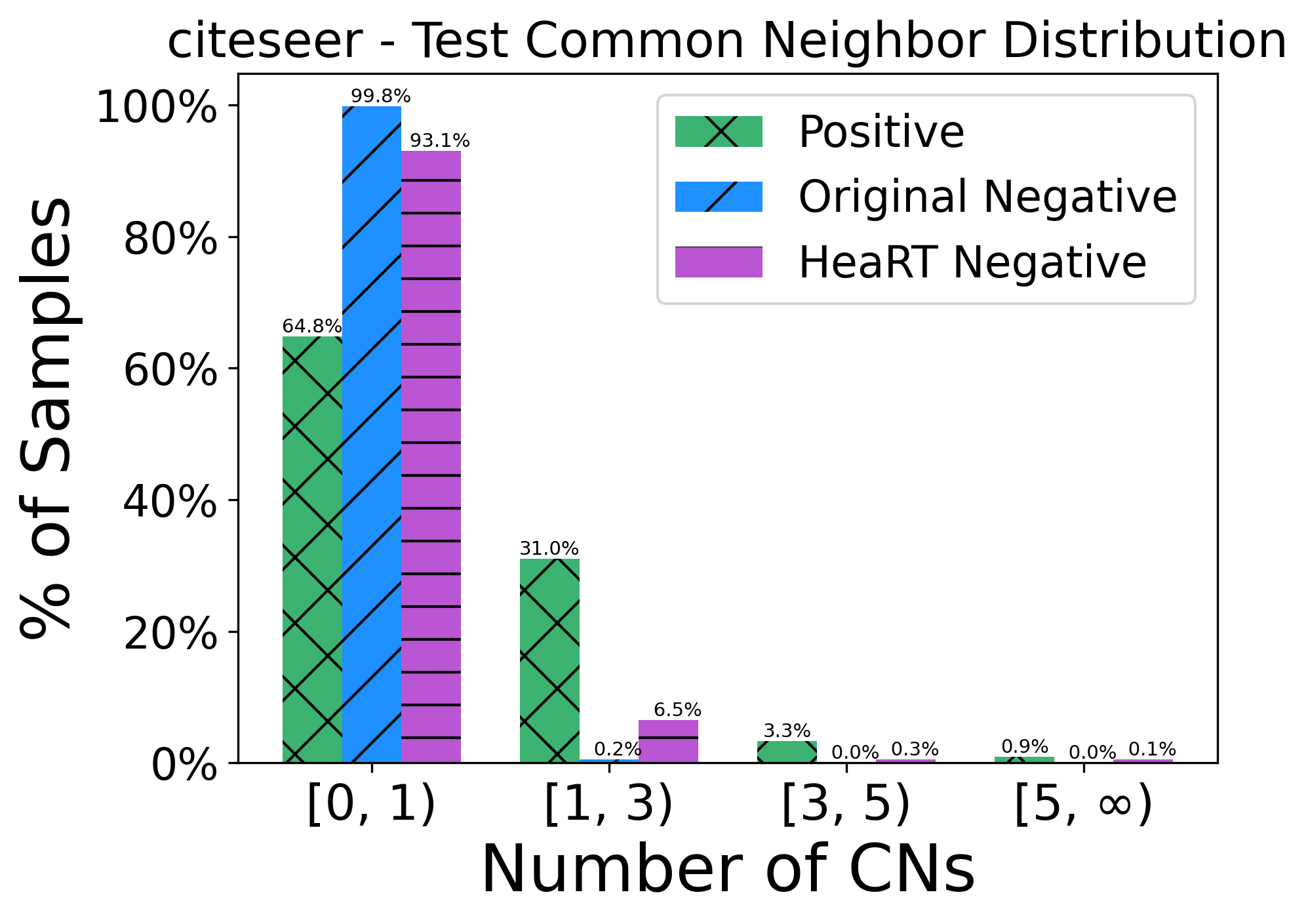} }}
        
        {\subfigure[Pubmed]
        {\includegraphics[width=0.345\linewidth]{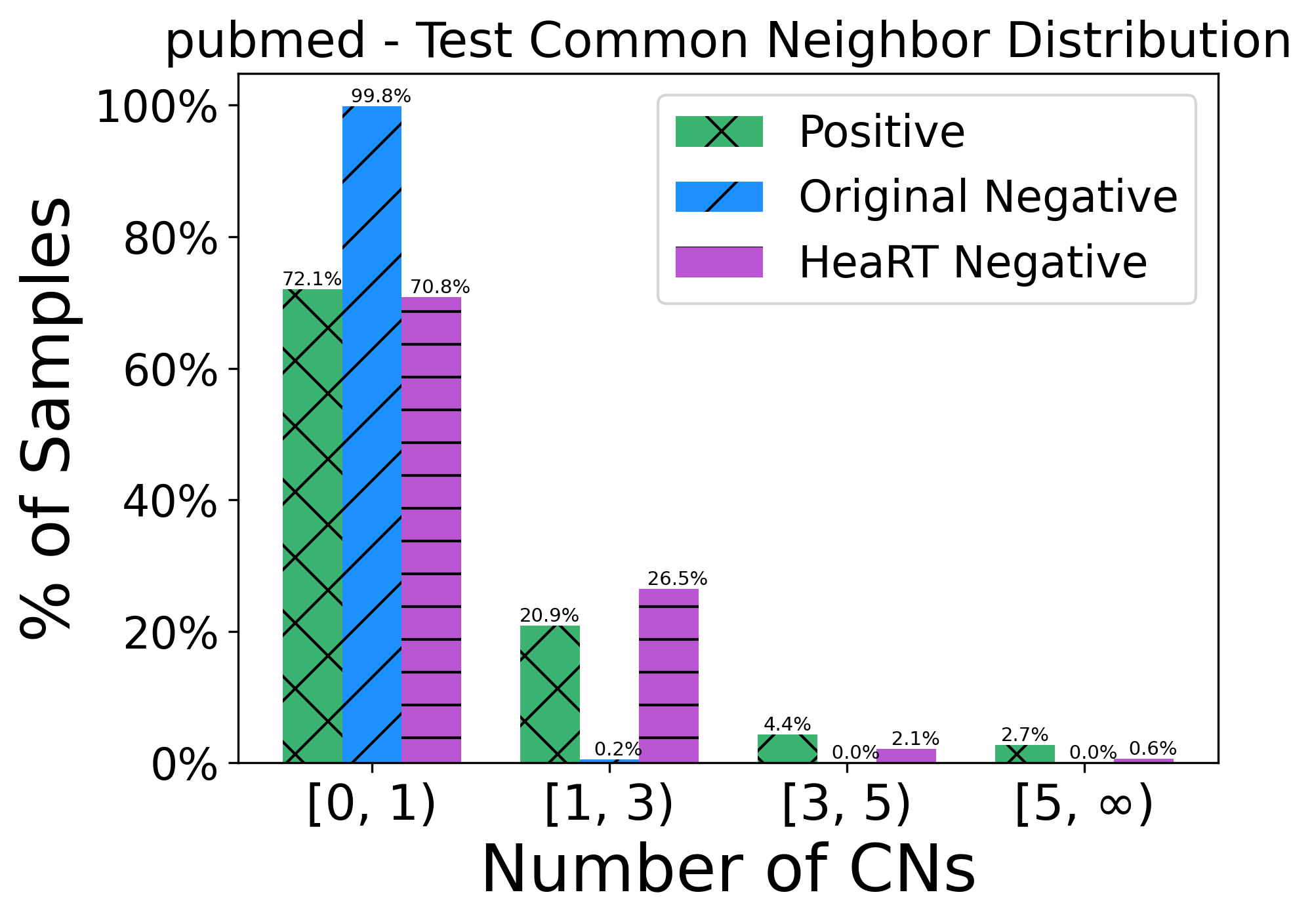} }}
    }
    
    \caption{{Common neighbor distribution for the positive negative samples under both evaluation settings for Cora, Citeseer, Pubmed. }}

\label{fig:cn_dist_heart1}
\end{center}
% \vspace{-.4in}
\end{figure*}

\begin{figure*}[t]
    \begin{center}
     \centerline{
        {\subfigure[ogbl-collab]
        {\includegraphics[width=0.345\linewidth]{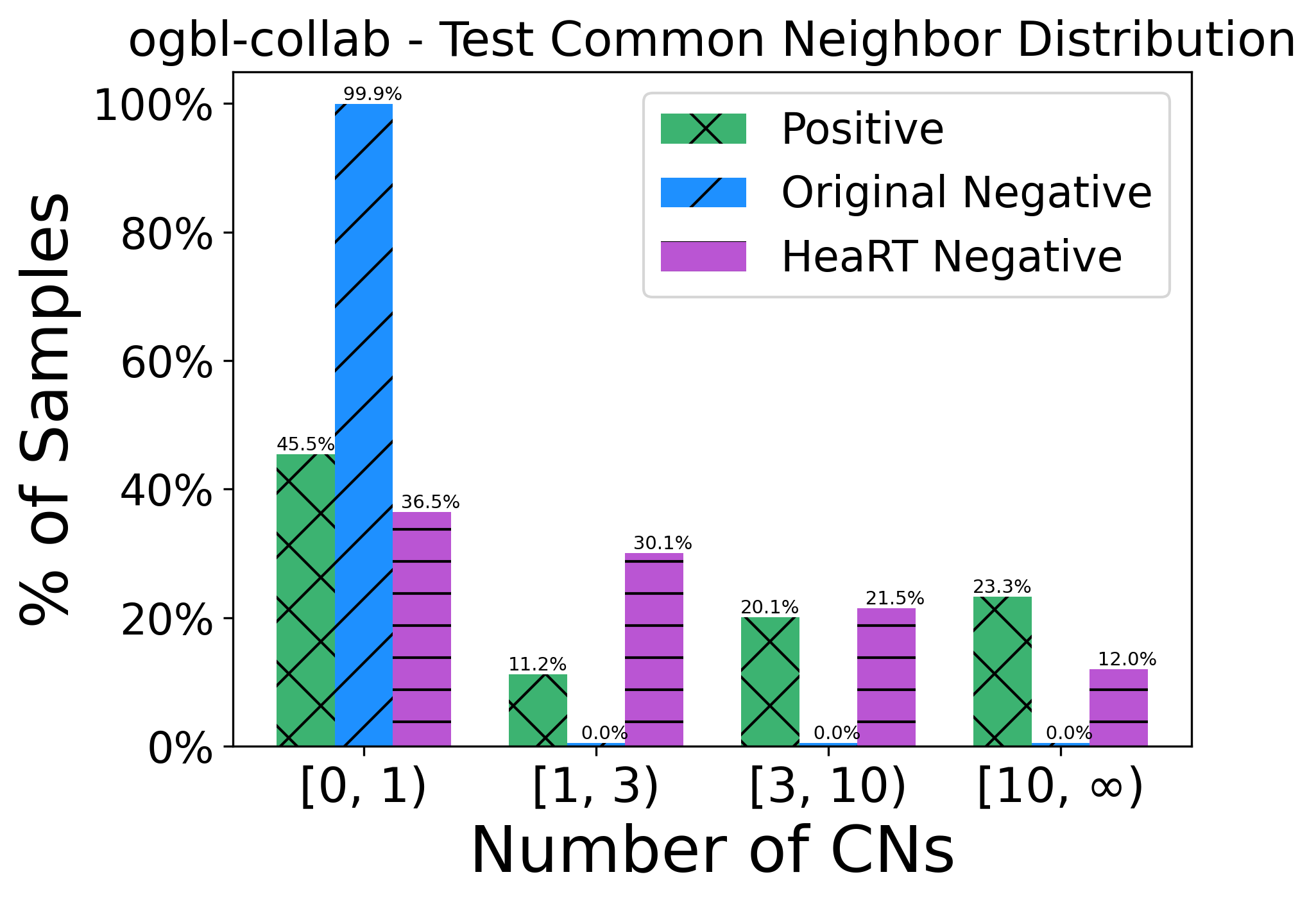} }
        
        }
        
        {\subfigure[ogbl-ddi]
        {\includegraphics[width=0.333\linewidth]{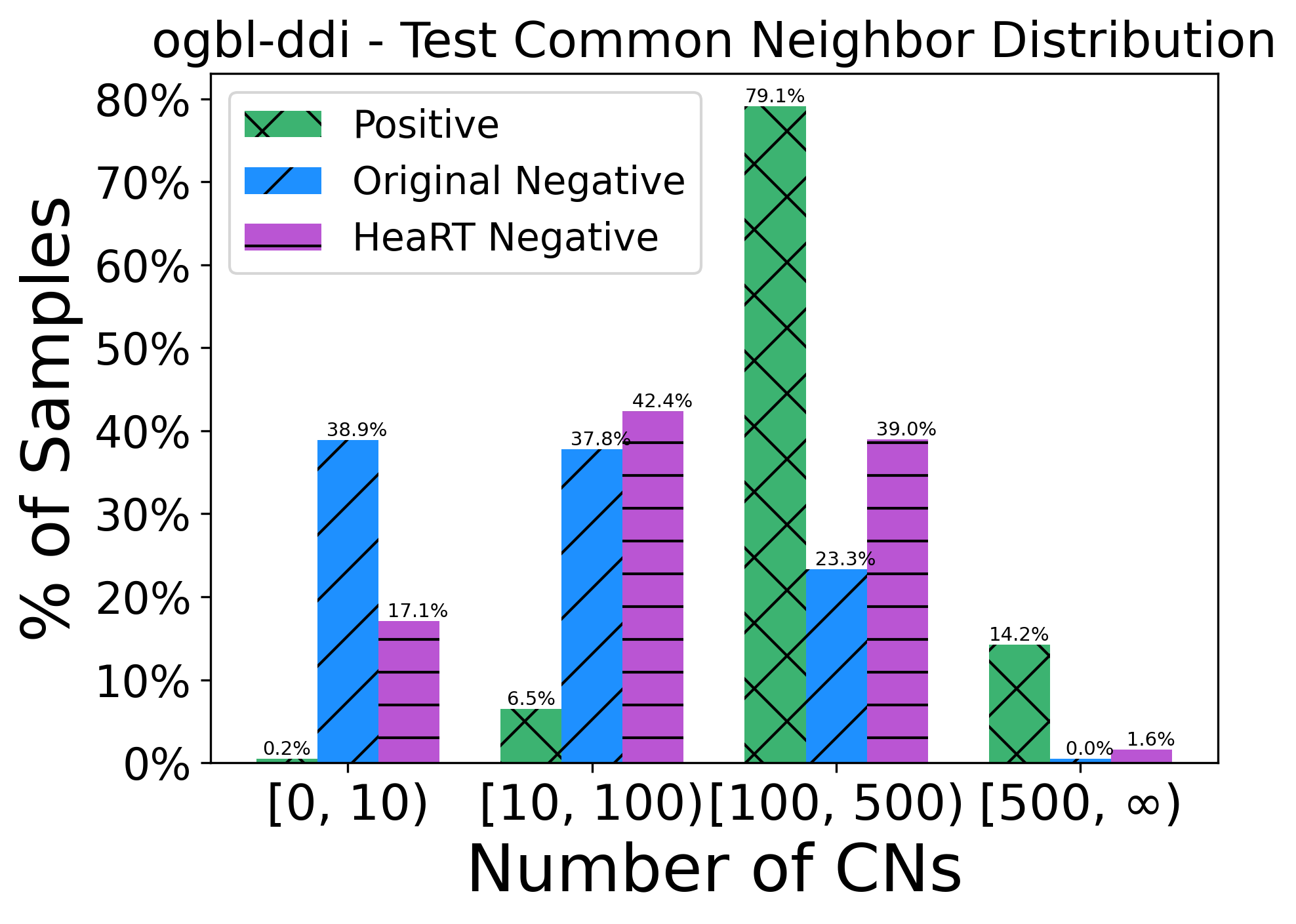} }}
        }
         \centerline{
        {\subfigure[ogbl-ppa]
        {\includegraphics[width=0.345\linewidth]{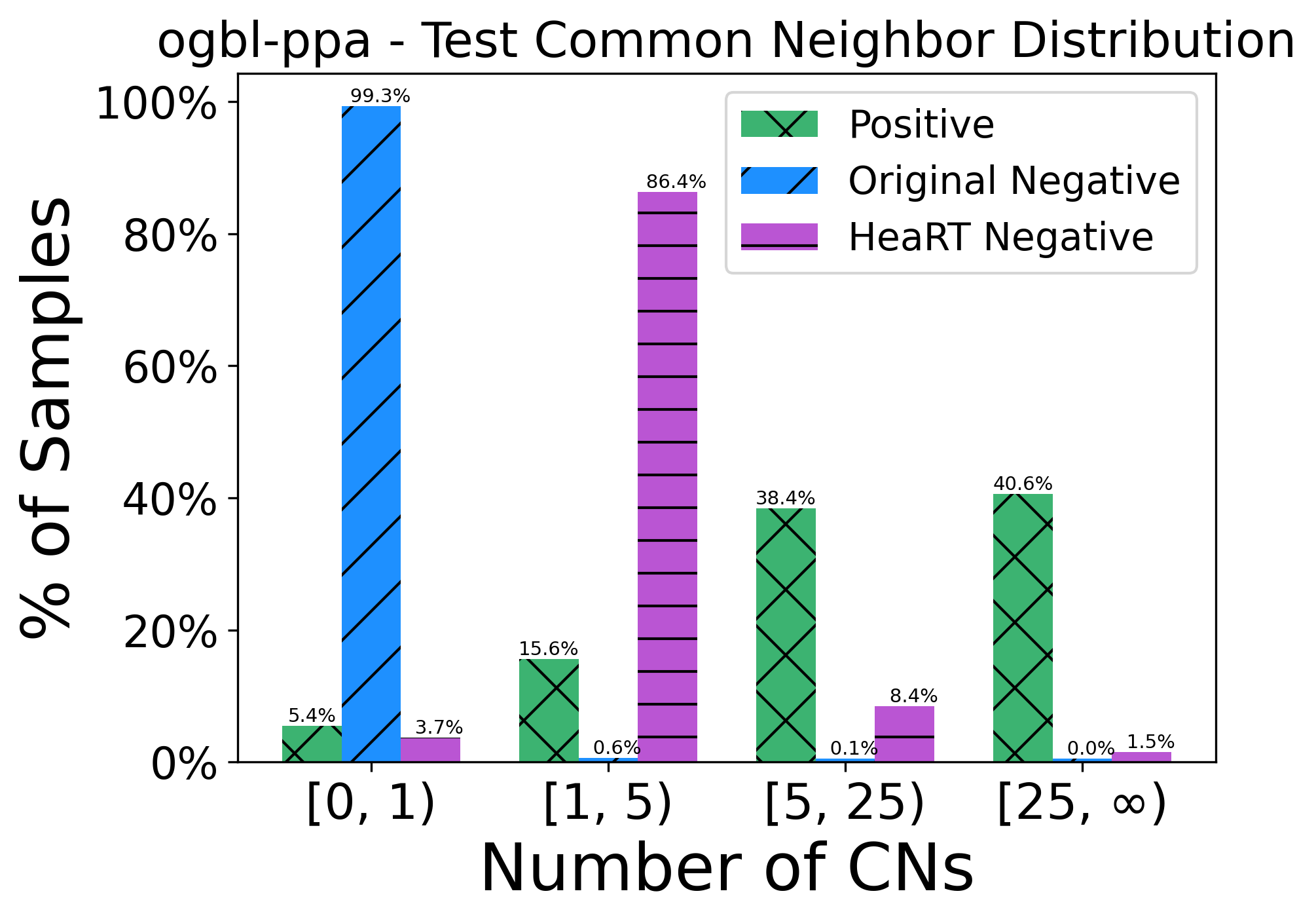} }}

         {\subfigure[ogbl-citation]
        {\includegraphics[width=0.36\linewidth]{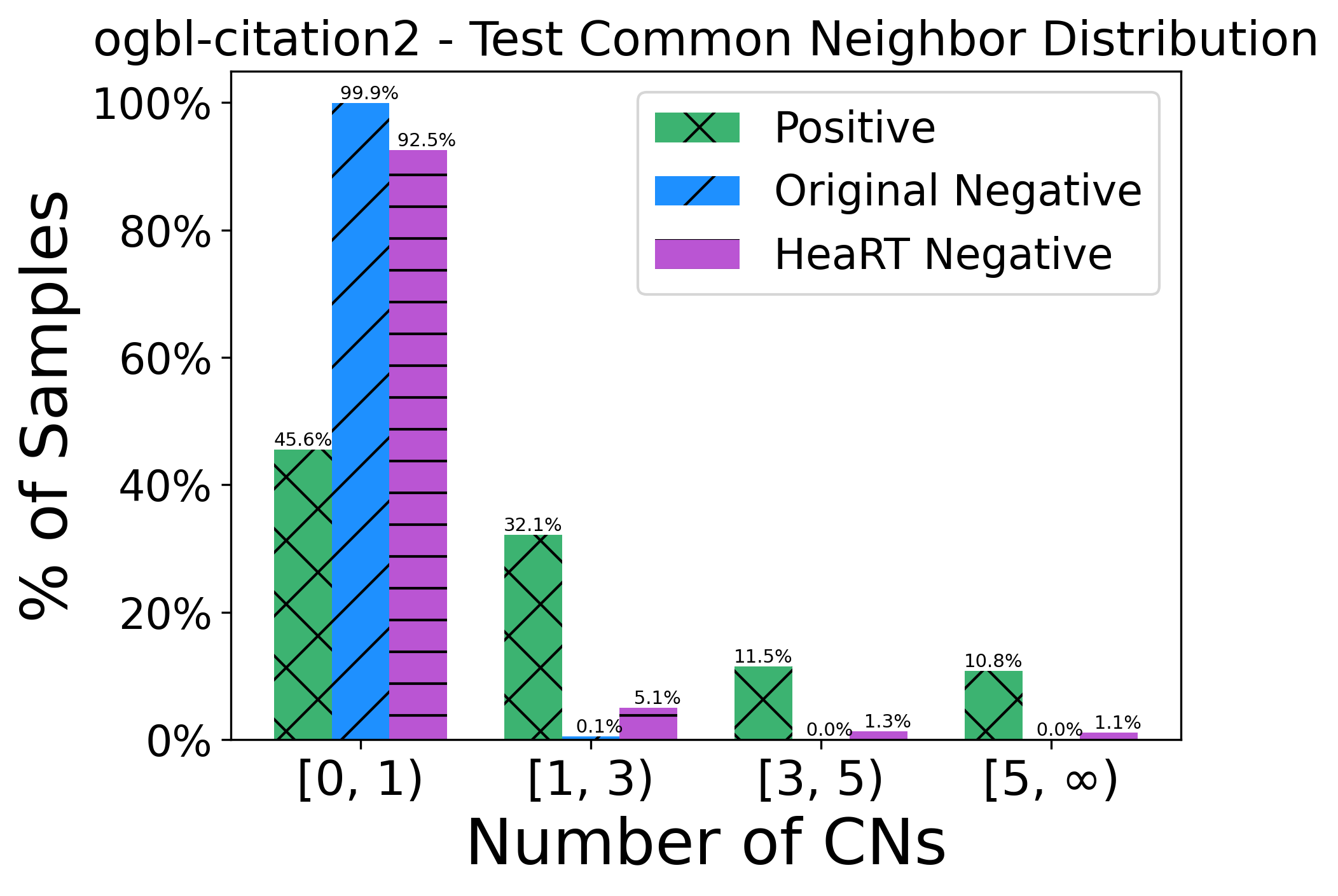} }}
    }
    
    \caption{{Common neighbor distribution for the positive negative samples under both evaluation settings for the OGB datasets.}
    }

\label{fig:cn_dist_heart2}
\end{center}

\end{figure*}

We also present the CN distributions under the HeaRT setting. The plots for Cora, Citeseer, Pubmed are shown in Figure~\ref{fig:cn_dist_heart1}. The plots for the OGB datasets are shown in Figure~\ref{fig:cn_dist_heart2}. We observe that the CN distribution of HeaRT is more aligned with the positive samples. This allows for a fairer evaluation setting by not favoring models that use CN information.

\section{Additional Definitions}

\subsection{Evaluation Metrics} \label{sec:eval_metrics_defs} 

{
In this section we define the various evaluation metrics used. Given a single positive sample and $M$ negative samples, we first score each sample and then rank the positive sample among the negatives. The rank is then given by $\text{rank}_i$. I.e., a rank of 1 indicates that the positive sample has a higher score than all negatives. The hope is that the positive sample ranks above most or all negative samples. Various metrics make use of this rank. We use $N$ to denote the number of positive samples.
}

{
{\bf Hits@K}. It measures whether the true positive is within the top K predictions or not: $\text{Hits@K} = \frac{1}{N}\sum_{i=1}^{N}\mathbf{1}(\text{rank}_{i} \leq \text{K})$.  $\text{rank}_i$ is the rank of the $i$-th sample. The indicator function {\bf 1} is 1 if $\text{rank}_{i} \leq \text{K}$, and 0 otherwise.
}

{
{\bf Mean Reciprocal Rank (MRR)}. It is the mean of the reciprocal rank over all positive samples: $\text{MRR}=\frac{1}{N}\sum_{i=1}^{N}\frac{1}{\text{rank}_i}$, where $\text{rank}_i$ is the rank of the $i$-th sample.
}

{
{\bf AUC}. It measures the likelihood that a
 positive sample is ranked higher than a random negative sample: $\text{AUC} = \frac{\sum_{i\in \mathcal{D}^0} \sum_{j\in \mathcal{D}^1} \mathbf{1}(\text{rank}_i < \text{rank}_j)}{|\mathcal{D}^0|\cdot |\mathcal{D}^1|}$, where $\mathcal{D}^0$ is the set of positive samples, $\mathcal{D}^1$ is the set of negative samples, and $\text{rank}_i$ is the rank of the $i$-th sample. The indicator function {\bf 1} is 1 if $\text{rank}_i < \text{rank}_j$, and 0 otherwise.
}
\subsection{Negative Sampling}
{
Since only positive links are observed, there is a need to generate negative links (i.e., edges that don't exist in $\mathcal{G}$) to both train and evaluate different models. We detail how these samples are generated in both training and evaluation.
}

{
{\bf Training Negative Samples}. During training, the negative samples are randomly selected, with all nodes being equally likely to be selected. Let $\mathcal{V}$ and $\mathcal{E}$ be the set of nodes and edges in $\mathcal{G}$. Furthermore, we define $v \in \text{Rand}(\mathcal{V})$ as returning a random node in $\mathcal{V}$. A single negative sample $(a^{-}, b^{-})$ is given by:
\begin{equation} \label{eq:rand_sample}
    (a^{-}, b^{-}) = \left(\text{Rand}(\mathcal{V}), \text{Rand}(\mathcal{V}) \right).
\end{equation}
Typically one negative sample is generated per positive sample.
}

{
{\bf Evaluation Negative Samples}. For the existing setting, a fixed set of randomly selected samples are used as negatives during evaluation. Furthermore, the same set of negative samples are used for each positive sample. This is equivalent to Eq.~\eqref{eq:rand_sample}. The only exception is the ogbl-citation2~\cite{hu2020open} dataset. For ogbl-citation2, each positive sample is only evaluated against its own set 1000 negative samples. For a positive sample, its negative samples are restricted to contain one of its two nodes (i.e., a corruption). The other node is randomly selected from $\mathcal{V}$. This is equivalent to selecting a set of random samples from the set $S(a, b)$ as defined in Eq.~\eqref{eq:candidate_set}.
}

\section{Datasets and Experimental Settings}
\label{sec:app_data_parameter}
\subsection{Datasets}

\begin{table}[h]
\centering

 \caption{Statistics of datasets. The split ratio is for train/validation/test. 
 }
  \begin{adjustbox}{width =1 \textwidth}
\begin{tabular}{cccccccc}
\toprule
 & Cora & Citeseer & Pubmed & ogbl-collab & ogbl-ddi & ogbl-ppa & ogbl-citation2 \\
 \midrule
\#Nodes & \multicolumn{1}{r}{2,708} & \multicolumn{1}{r}{3,327} & \multicolumn{1}{r}{18,717} & \multicolumn{1}{r}{235,868} & \multicolumn{1}{r}{4,267} & \multicolumn{1}{r}{576,289} & \multicolumn{1}{r}{2,927,963} \\
\#Edges & \multicolumn{1}{r}{5,278} & \multicolumn{1}{r}{4,676} & \multicolumn{1}{r}{44,327} & \multicolumn{1}{r}{1,285,465} & \multicolumn{1}{r}{1,334,889} & \multicolumn{1}{r}{30,326,273} & \multicolumn{1}{r}{30,561,187} \\
Mean Degree & \multicolumn{1}{r}{3.9} & \multicolumn{1}{r}{2.81} & \multicolumn{1}{r}{4.74} & \multicolumn{1}{r}{10.90} & \multicolumn{1}{r}{625.68} & \multicolumn{1}{r}{105.25} & \multicolumn{1}{r}{20.88} \\
Split Ratio& \multicolumn{1}{r}{85/5/10}&\multicolumn{1}{r}{85/5/10} &\multicolumn{1}{r}{85/5/10} &\multicolumn{1}{r}{92/4/4} & \multicolumn{1}{r}{80/10/10} & \multicolumn{1}{r}{70/20/10}&\multicolumn{1}{r}{98/1/1}\\
\bottomrule
\end{tabular}
\label{table:app_data}
\end{adjustbox}
\end{table}

The statistics of datasets are shown in Table~\ref{table:app_data}. 
 Generally,  Cora, Citeseer, and Pubmed are smaller graphs, with the OGB datasets having more nodes and edges.
We adopt the single fixed train/validation/test split with percentages 85/5/10\% for Cora, Citeseer, and Pubmed. For OGB datasets, we use the fixed splits provided by the OGB benchmark~\cite{hu2020open}.

\subsection{Experimental Settings} \label{sec:param_settings}

{\bf Training Settings}. We use the binary cross entropy loss to train each model. The loss is optimized using the Adam optimizer~\cite{kingma2014adam}.  
During training we randomly sample one negative sample per positive sample. 
Each model is trained for a maximum of 9999 epochs, with the process set to terminate when there are no improvements observed in the validation performance over $n$ checkpoints.
The choice of $n$ is influenced by both the specific dataset and the complexity of the model.  For smaller datasets, such as Cora, Citeseer, and Pubmed, we set $n=50$ uniformly across  models (except for NBFNet and SEAL where $n=20$ due to their computational inefficiency). When training on larger OGB datasets, we use a stratified approach: $n=100$ for the simpler methods (i.e., the embedding and GNN-based models) and $n=20$ for the more advanced methods. This is due to the increased complexity and runtime of more advanced methods. An exception is for ogbl-citation2, the largest dataset. To accommodate for its size, we limit the maximum number of epochs to the recommended value from each model's source code. Furthermore, we set  $n=20$ for all models.

In order to accommodate the computational requirements for our extensive experiments, we harness a variety of high-capacity GPU resources. This includes: Tesla V100 32Gb, NVIDIA RTX A6000 48Gb, NVIDIA RTX A5000 24Gb, and Quadro RTX 8000 48Gb. 

\begin{table}[h]
\centering

 \caption{Hyperparameter Search Ranges}
  \begin{adjustbox}{width =1 \textwidth}
\begin{tabular}{l|cccccc}
\toprule
 Dataset & Learning Rate &Dropout& Weight Decay & \# Model Layers & \# Prediction Layers  & Embedding Dim \\
 \midrule
Cora &(0.01, 0.001) &(0.1, 0.3, 0.5) &(1e-4, 1e-7, 0)&(1, 2, 3)&(1, 2, 3)&(128, 256)\\
Citeseer &(0.01, 0.001) &(0.1, 0.3, 0.5) &(1e-4, 1e-7, 0)&(1, 2, 3)&(1, 2, 3)&(128, 256)\\
Pubmed &(0.01, 0.001) &(0.1, 0.3, 0.5) &(1e-4, 1e-7, 0)&(1, 2, 3)&(1, 2, 3)&(128, 256)\\
ogbl-collab &(0.01, 0.001) &(0, 0.3, 0.5)&0 & 3& 3 & 256\\
ogbl-ddi &(0.01, 0.001) &(0, 0.3, 0.5)&0 & 3& 3 & 256\\
ogbl-ppa &(0.01, 0.001) &(0, 0.3, 0.5)&0 & 3& 3 & 256\\
ogbl-citation2 &(0.01, 0.001) &(0, 0.3, 0.5)&0 & 3& 3 & 128\\
\bottomrule
\end{tabular}
\label{table:app_parameter_setting}
\end{adjustbox}
\end{table}

{\bf Hyperparameter Settings}. We present the hyparameter searching range in Table~\ref{table:app_parameter_setting}.  For the smaller graphs, Cora, Citeseer, and Pubmed, we have a larger search space. However, it's not feasible to tune over such large space for OGB datasets. 
By following the most commonly used settings among published hyperparameters, we fix the weight decay, number of model and prediction layers, and the embedding dimension. Furthermore, due to GPU memory constraints, the embedding size is reduced to be 128 for the largest dataset ogbl-citation2. 

We note that several exceptions exist to these ranges when they result in significant performance degradations. In such instances, adjustments are guided by the optimal hyperparameters published in the respective source codes. This includes:
\begin{itemize} [leftmargin=0.3in]
    \item {PEG}~\cite{wang2022equivariant}: Adhering to the optimal hyperparameters presented in the source code,\footnote{https://github.com/Graph-COM/PEG/} when training on ogbl-ddi we set the number of model layers to 2 and the maximum number of epochs to 400.
    
    \item {NCN/NCNC}~\cite{wang2023neural}: When training on ogbl-ddi, 
    we adhere to the suggested optimal hyperparameters used in the source code.\footnote{https://github.com/GraphPKU/NeuralCommonNeighbor/} Specifically, we set the number of model layers to be 1,  and  we don't apply the pretraining for NCNC  to facilitate a fair comparison.

    \item {NBFNet}~\cite{zhu2021neural}: Due to the expensive nature of NBFNet, we further fix the weight decay to 0 when training on Cora, Citeseer, and Pubmed. Furthermore, we follow the suggested hyperparameters~\footnote{https://github.com/DeepGraphLearning/NBFNet/} and set the embedding dimension to be 32 and the number of model layers to be 6.

    \item {SEAL}~\cite{zhang2018link}: Due to the computational inefficiency of SEAL, when training on Cora, Citeseer and Pubmed  we further fix the weight decay to 0. Furthermore, we adhere to the published hyperparameters~\footnote{https://github.com/facebookresearch/SEAL\_OGB/} and fix the number of model layers to be 3 and the embedding dimension to be 256.

    \item {BUDDY}~\cite{chamberlain2022graph}: When training on ogbl-ppa, we incorporate the RA and normalized degree as input features while excluding the raw node features. This is based on the optimal hyperparameters published by the authors.\footnote{https://github.com/melifluos/subgraph-sketching/}
\end{itemize}

\section{Reported vs. Our Results on ogbl-ddi}
\label{sec:app_reportVSour_ddi}
In Section~\ref{sec:benchmark} (see observation 2), we remarked that there is divergence between the reported results and our results on ogbl-ddi for some methods. A comprehensive comparison of this discrepancy is shown in Table~\ref{table:app_reportVSour_ddi}.
The reported results for Node2Vec, MF, GCN, and SAGE are taken from~\cite{hu2020open}. The results for the other methods are from their original paper:  SEAL~\cite{zhang2021labeling},  BUDDY~\cite{chamberlain2022graph}, Neo-GNN~\cite{yun2021neo},	NCN~\cite{wang2023neural}, NCNC~\cite{wang2023neural}, and PEG~\cite{wang2022equivariant}.

\begin{table}[h]
\centering

 \caption{Comparison results between ours and reported results on ogbl-ddi (Hits@20).}
  \begin{adjustbox}{width =1 \textwidth}
\begin{tabular}{ccccccccccc}
\toprule
 &Node2Vec	&MF& GCN	&SAGE&	SEAL&	BUDDY&	Neo-GNN&	NCN	&NCNC&	PEG \\
 \midrule
Reported & 23.26 ± 2.09	&13.68 ± 4.75	&37.07 ± 5.07&{\bf 53.90 ± 4.74}	&{\bf 30.56 ± 3.86} &	{\bf 78.51 ± 1.36}	&{\bf 63.57 ± 3.52}&	{\bf 82.32 ± 6.10}	&{\bf 84.11 ± 3.67}	&{\bf 43.80 ± 0.32}\\
Ours&{\bf 34.69 ± 2.90}	&{\bf 23.50 ± 5.35}	&{\bf 49.90 ± 7.23}	&49.84 ± 15.56&	25.25 ± 3.90 &29.60 ± 4.75	&20.95 ± 6.03	&76.52 ± 10.47&	70.23 ± 12.11	&30.28 ± 4.92 \\
\bottomrule
\end{tabular}
\label{table:app_reportVSour_ddi}
\end{adjustbox}
\end{table}

\section{Additional Investigation on ogbl-ddi} 
\label{sec:app_ddi_investigate}

In this section, we present the additional investigation on the ogbl-ddi dataset. In Section~\ref{sec:app_ddi_old} we examine under the existing evaluation setting, there exists a poor relationship between the validation and test performance on ogbl-ddi for many methods. We then demonstrate in Section~\ref{sec:app_ddi_new} that under HeaRT this problem is lessened.  

\subsection{Existing Evaluation Setting} \label{sec:app_ddi_old}

\begin{figure*}[t]
    \begin{center}
     \centerline{
        {\subfigure[GCN]
        {\includegraphics[width=0.25\linewidth]{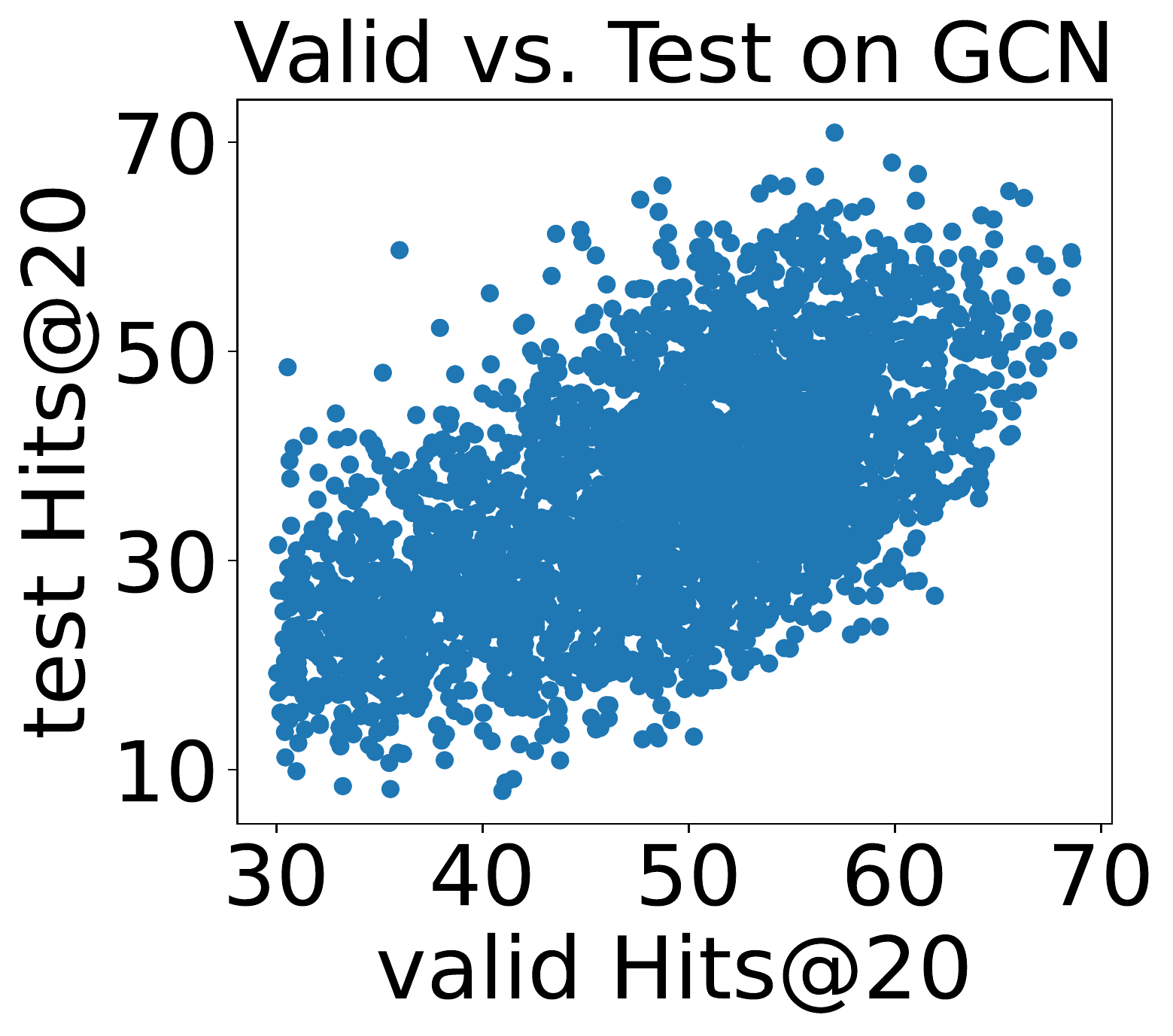} }}
        
        {\subfigure[BUDDY]
        {\includegraphics[width=0.25\linewidth]{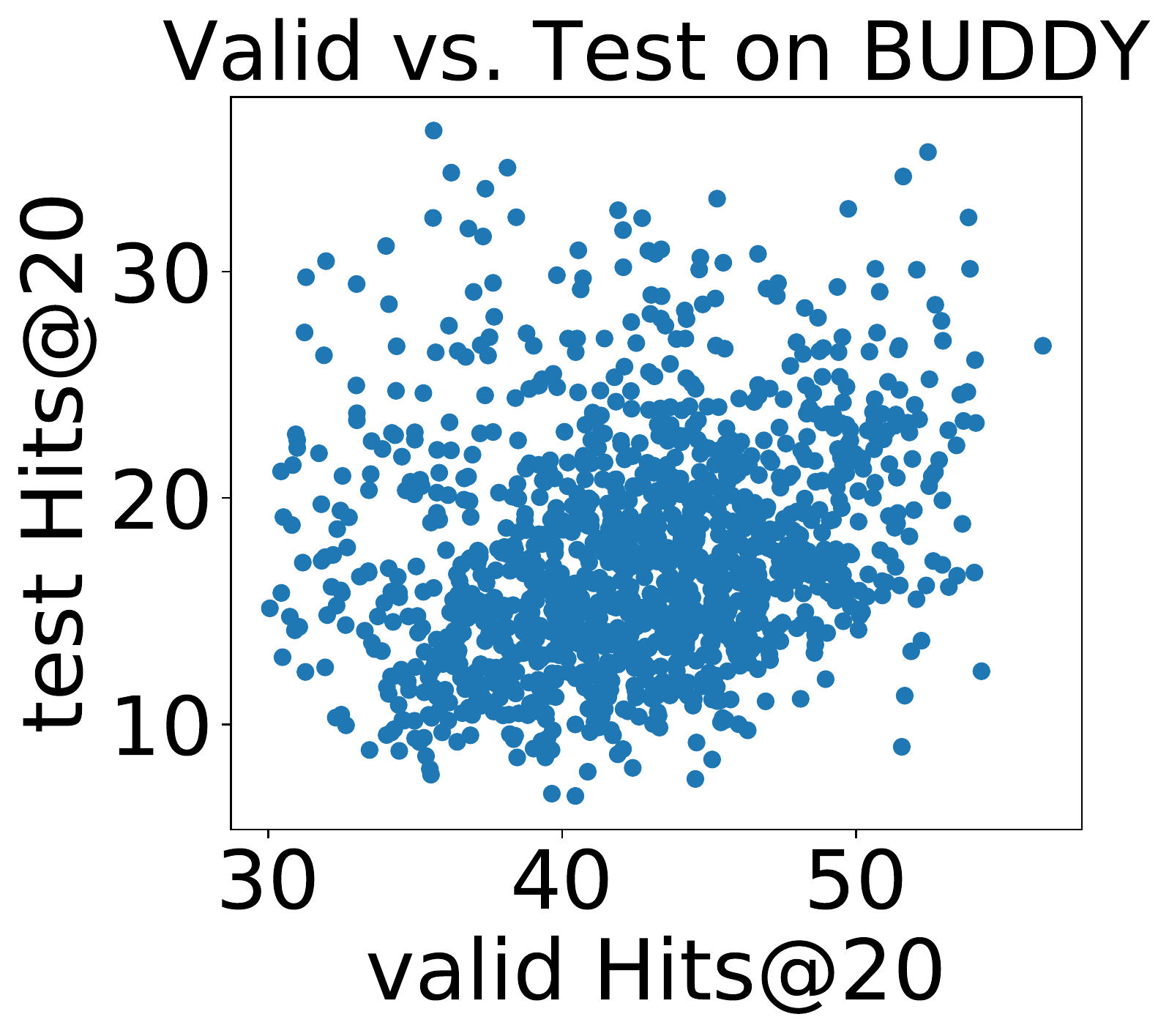} }}
        
        {\subfigure[NCN]
        {\includegraphics[width=0.25\linewidth]{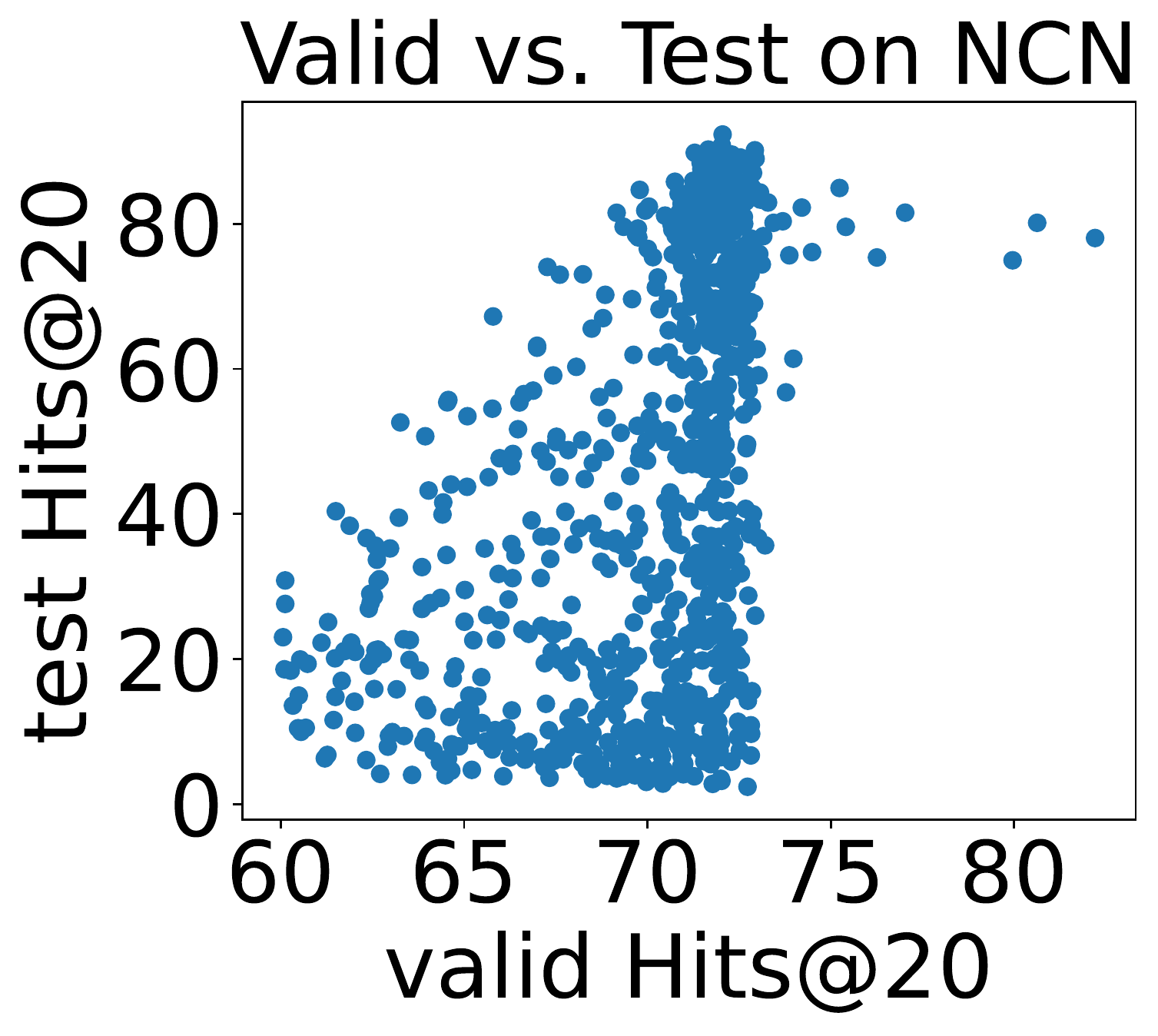} }}

        {\subfigure[Neo-GNN]
        {\includegraphics[width=0.25\linewidth]{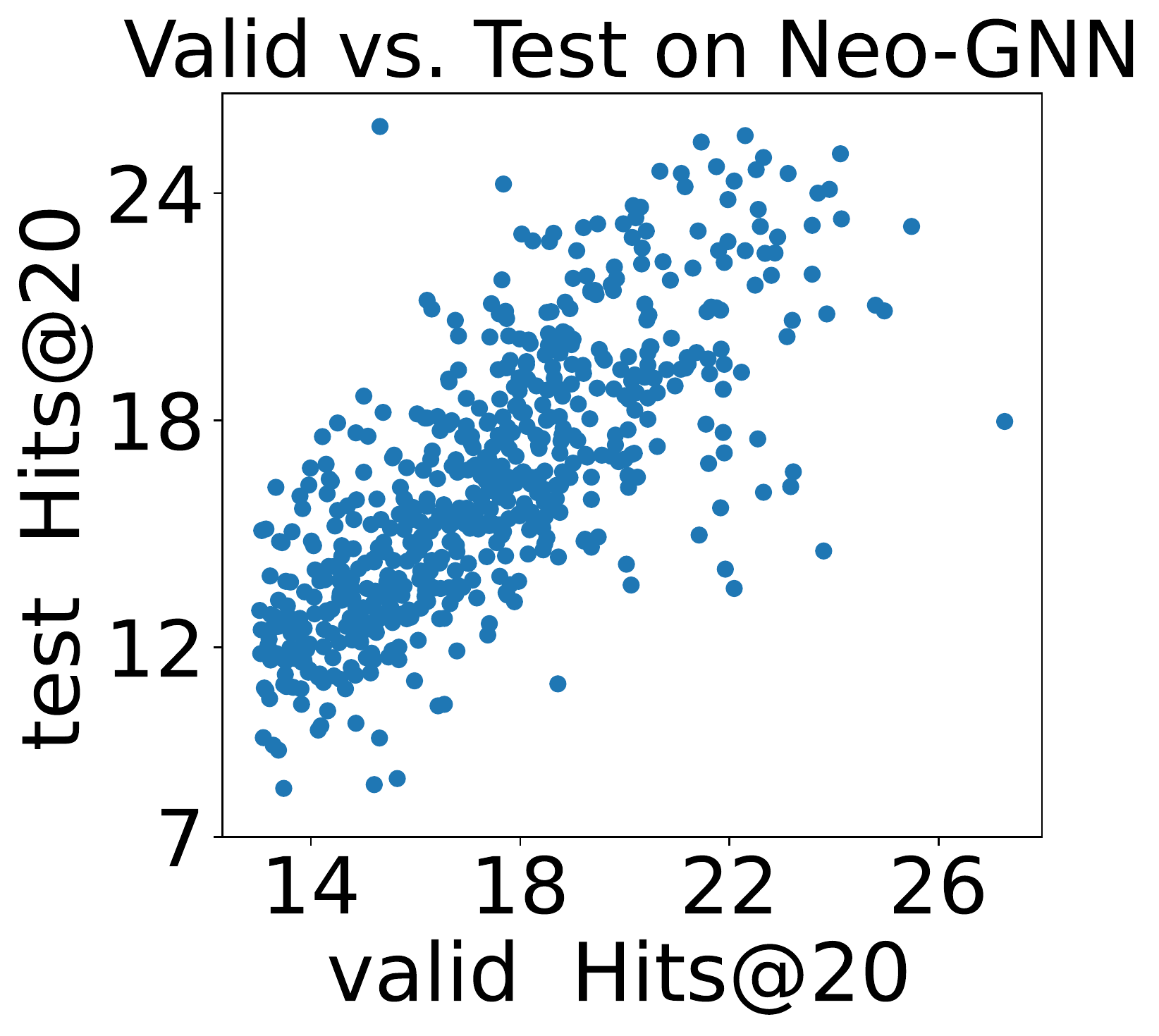} }}
    }
    
    \caption{Validation vs. test performance for GCN, BUDDY, NCN and Neo-GNN on ogbl-ddi under the existing evaluation setting. 
    }

\label{fig:ddi_old}
\end{center}
% \vspace{-0.35in}
\end{figure*}

Upon inspection, we found that there is a poor relationship between the validation and test performance on ogbl-ddi. Since we choose the best hyperparameters based on the validation set, this makes it difficult to properly tune any model on ogbl-ddi. To demonstrate this point, we record the validation and test performance at multiple checkpoints during the training process. The experiments are conducted over 10 seeds. To ensure that our results are not caused by our hyperparameter settings, we use the reported hyperparameters for each model. Lastly, we plot the results for GCN, BUDDY, NCN, and Neo-GNN in \figurename~\ref{fig:ddi_old}. It is clear from the results that there exists a poor relationship between the validation and test performance. For example, for NCN, a validation performance of 70 can imply a test performance of 3 to 80. Further investigation is needed to uncover the cause of this misalignment.

\begin{figure*}[t]
    \begin{center}
     \centerline{
        {\subfigure[GCN]
        {\includegraphics[width=0.25\linewidth]{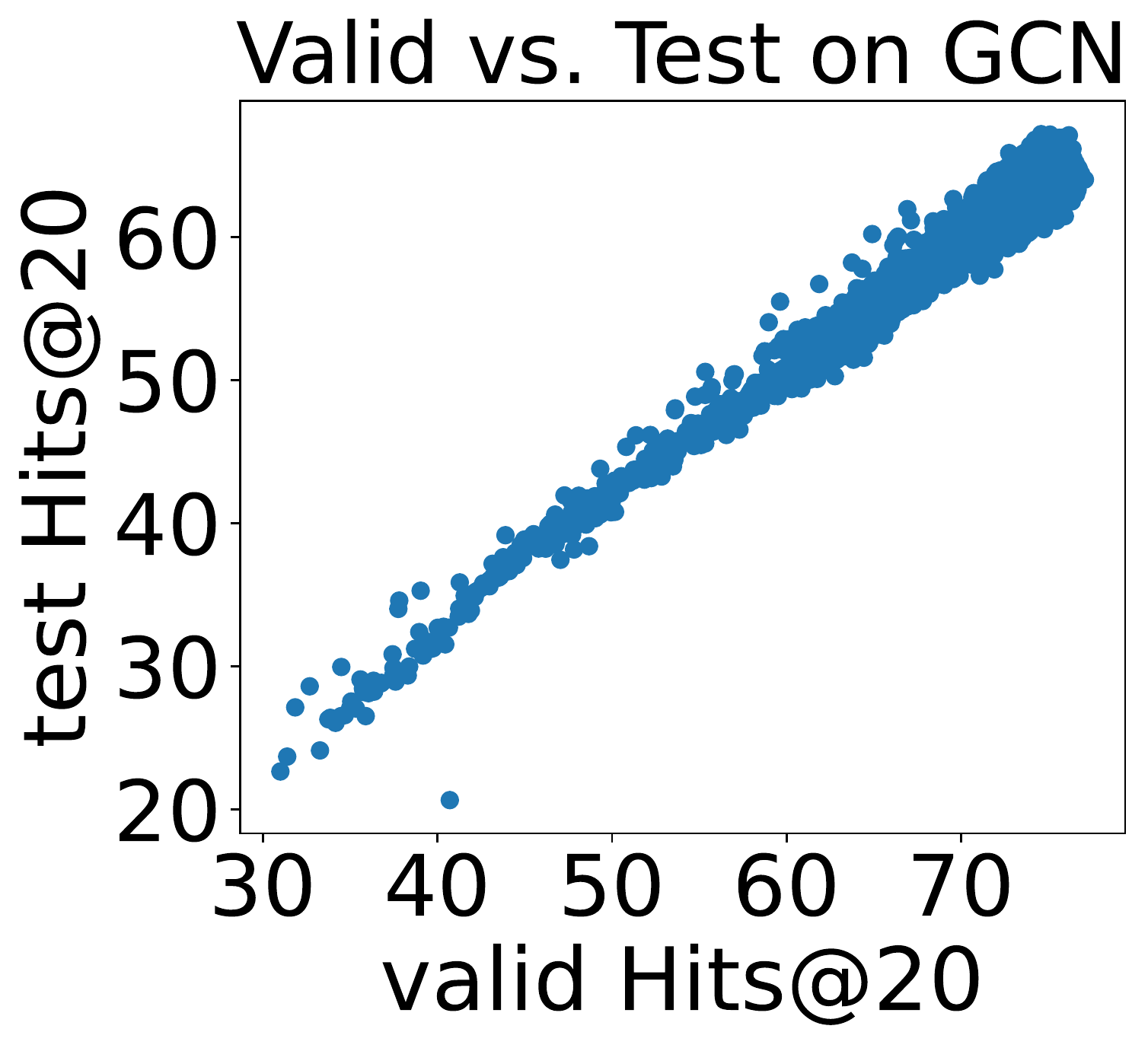} }}
        
        {\subfigure[BUDDY]
        {\includegraphics[width=0.25\linewidth]{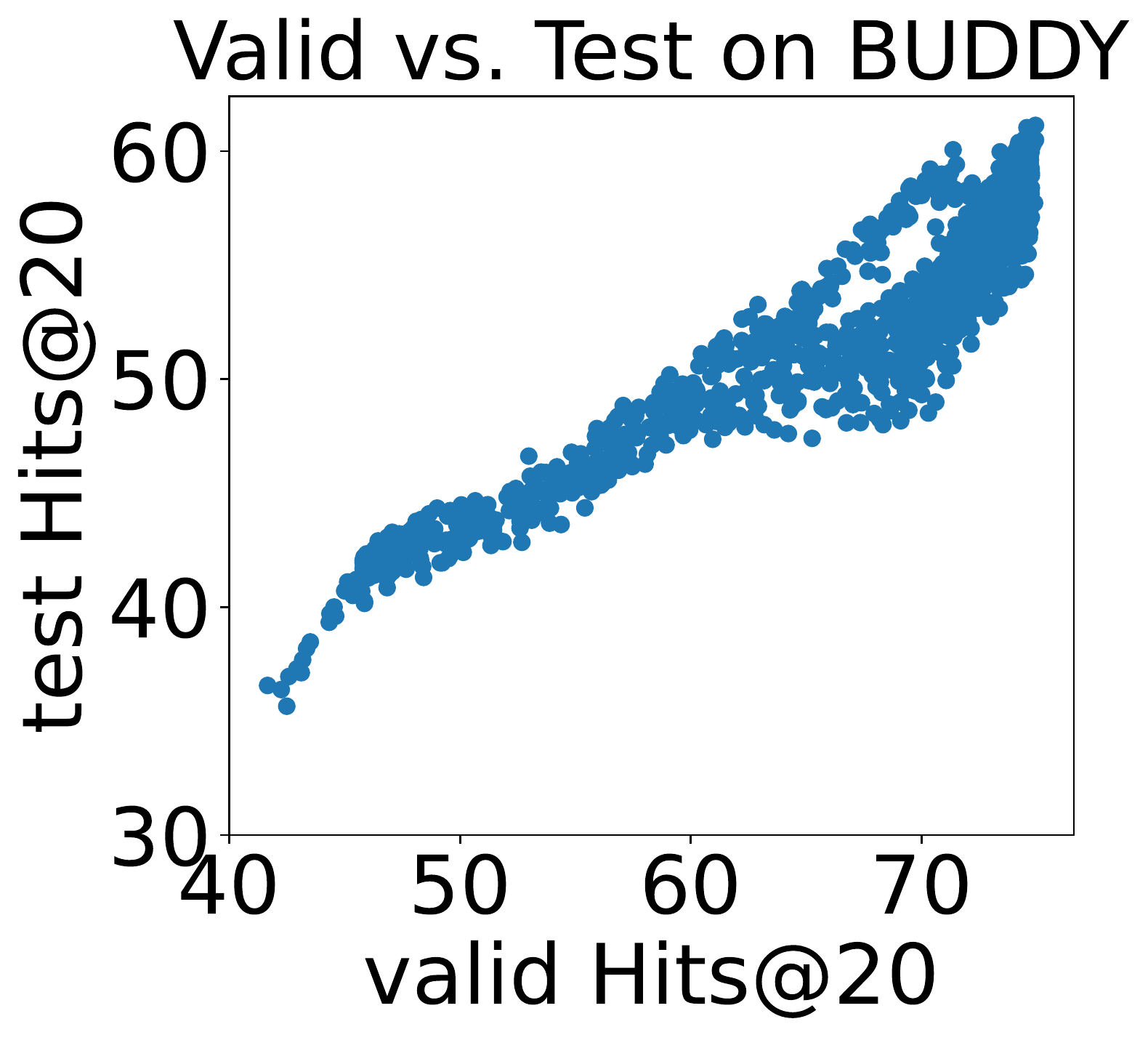} }}
        
        {\subfigure[NCN]
        {\includegraphics[width=0.25\linewidth]{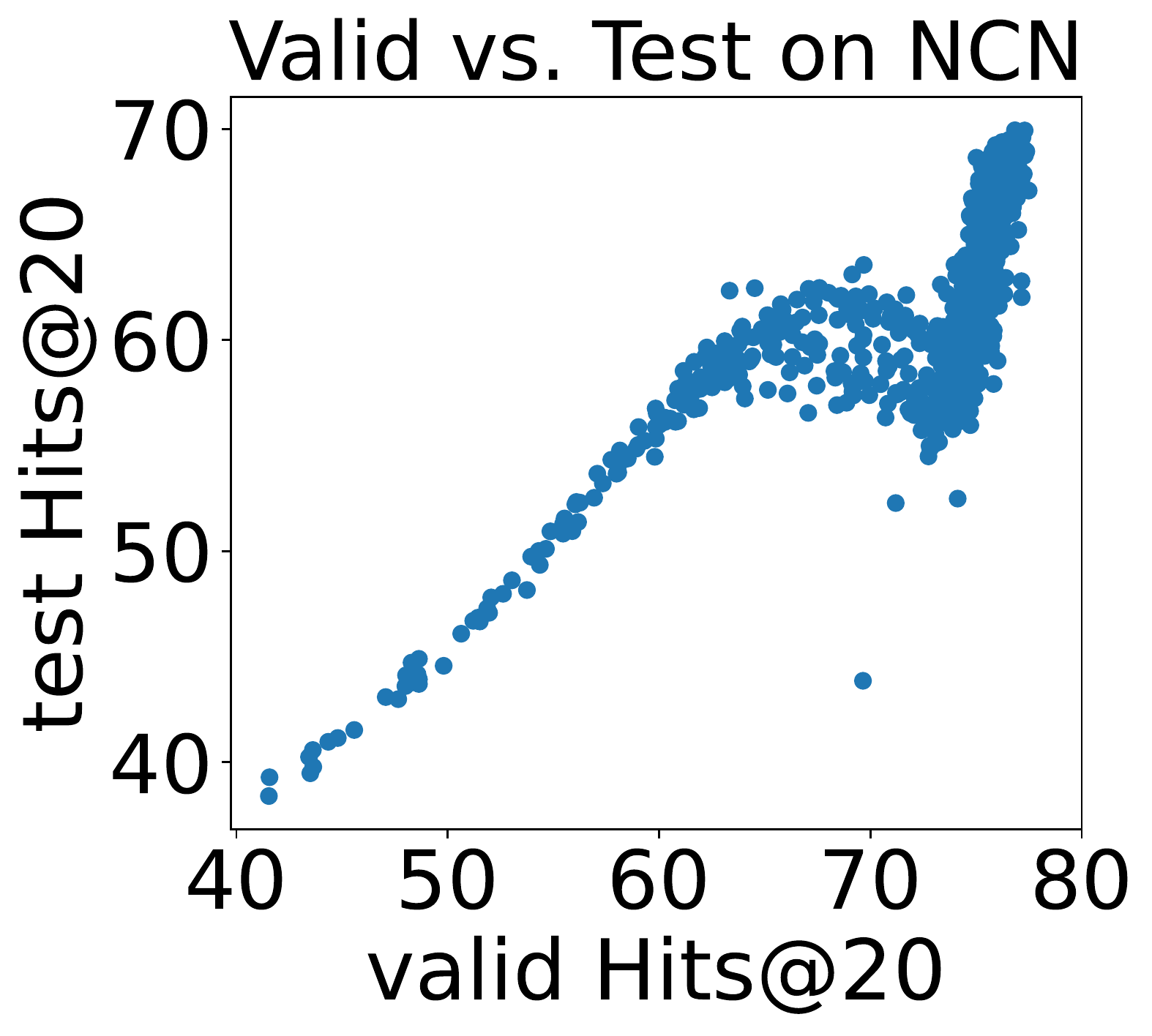} }}

        {\subfigure[Neo-GNN]
        {\includegraphics[width=0.25\linewidth]{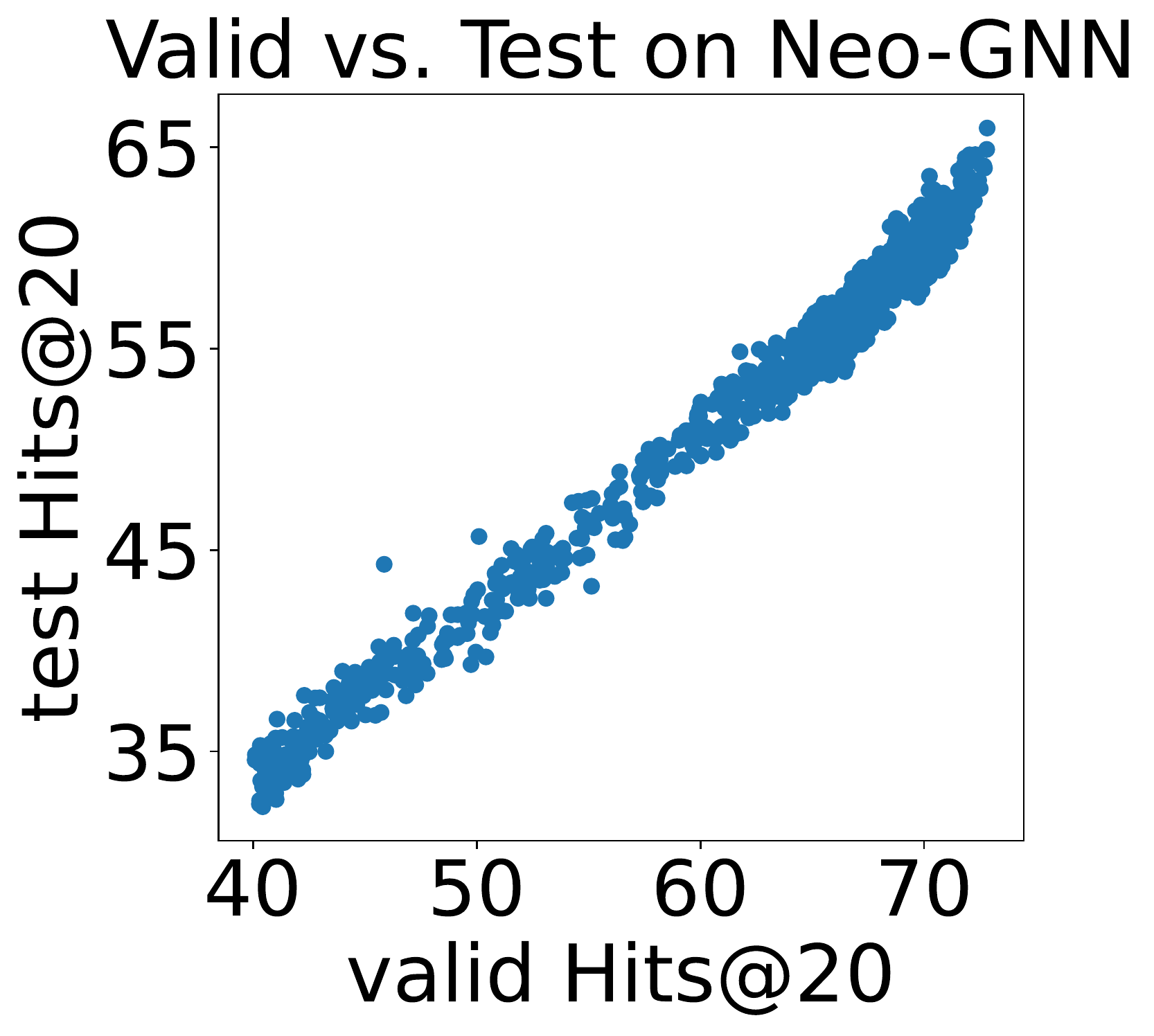} }}
    }
    
    \caption{Validation vs. test  performance when utilizing HeaRT for GCN, BUDDY, NCN and Neo-GNN on ogbl-ddi. We find they have a much stronger relationship than under the existing setting (see \figurename~\ref{fig:ddi_old}). 
    }

\label{fig:ddi_new}
\end{center}
% \vspace{-0.4in}
\end{figure*}

\subsection{New Evaluation Setting} \label{sec:app_ddi_new}

Under our new setting, we find that the validation and test performance have a much better relationship. In Section~\ref{sec:app_ddi_old} we observed that there exists a poor relationship between the validation and test performance on ogbl-ddi under the existing evaluation setting. This meddles with our ability to choose the best hyperparameters for each model, as good validation performance is not indicative of good test performance. However, this does not seem to be the case under the new evaluation setting. In Figure~\ref{fig:ddi_new} we plot the relationship between the validation and test performance by checkpoint for various models. Compared to the same plots under the existing setting (Figure~\ref{fig:ddi_old}), the new results display a much better relationship. 

While it's unclear what is the cause of the poor relationship between the test and validation performance under the existing setting, we conjecture that tailoring the negatives to each positive sample allows for a more natural comparison between a positive sample and its negatives. This may help produce more stable evaluation metrics, thereby strengthening the alignment between the validation and test performance.

\section{Additional Results Under the Existing Setting}
\label{sec:app_ben_more}
We present additional results of  Cora, Citeseer, Pubmed and OGB datasets in Tables~\ref{table:app_cora_ben_more}-\ref{table:app_ogb_ben_more} under the existing setting. We also omit the MRR for ogbl-collab, ogbl-ddi, and ogbl-ppa. This is because the large number of negative samples make it very inefficient to calculate.  
 
\begin{table}[]
\centering
\footnotesize
 \caption{Additional results on Cora(\%) under the existing evaluation setting. Highlighted are the results ranked \colorfirst{first}, \colorsecond{second}, and  \colorthird{third}.}
 
 \begin{adjustbox}{width =0.7 \textwidth}
\begin{tabular}{cc|cccc}
\toprule
 &Models  &Hits@1 & Hits@3   & Hits@10  & Hits@100  \\ 
 \midrule
 \multirow{5}{*}{Heuristic} & CN & \multicolumn{1}{c}{13.47} & \multicolumn{1}{c}{13.47} & \multicolumn{1}{c}{42.69} & \multicolumn{1}{c}{42.69} \\
 & AA & \multicolumn{1}{c}{22.2} & \multicolumn{1}{c}{39.47} & \multicolumn{1}{c}{42.69} & \multicolumn{1}{c}{42.69} \\
 & RA & \multicolumn{1}{c}{20.11} & \multicolumn{1}{c}{39.47} & \multicolumn{1}{c}{42.69} & \multicolumn{1}{c}{42.69} \\
 & Shortest Path & \multicolumn{1}{c}{0} & \multicolumn{1}{c}{0} & \multicolumn{1}{c}{42.69} & \multicolumn{1}{c}{71.35} \\
 & Katz & \multicolumn{1}{c}{19.17} & \multicolumn{1}{c}{28.46} & \multicolumn{1}{c}{51.61} & \multicolumn{1}{c}{74.57} \\
 \midrule
 \multirow{3}{*}{Embedding} & Node2Vec & \colorthird{22.3 ± 11.76} & \colorthird{41.63 ± 10.5} & 62.34 ± 2.35 & 84.88 ± 0.96 \\
 & MF & 7.76 ± 5.61 & 13.26 ± 4.52 & 29.16 ± 6.68 & 66.39 ± 5.03 \\
 & MLP & 18.79 ± 11.40 & 35.35 ± 10.71 & 53.59 ± 3.57 & 85.52 ± 1.44 \\
 \midrule
 \multirow{4}{*}{GNN} & GCN & 16.13 ± 11.18 & 32.54 ± 10.83 & 66.11 ± 4.03 & 91.29 ± 1.25 \\
 & GAT & 18.02 ± 8.96 & \colorsecond{42.28 ± 6.37} & 63.82 ± 2.72 & 90.70 ± 1.03 \\
 & SAGE &\colorsecond{ 29.01 ± 6.42} & \colorfirst{44.51 ± 6.57} & 63.66 ± 4.98 & 91.00 ± 1.52 \\
 & GAE & 17.57 ± 4.37 & 24.82 ± 4.91 & \colorthird{70.29 ± 2.75} & \colorthird{92.75 ± 0.95} \\
 \midrule
  \multirow{7}{*}{GNN+heuristic}& SEAL & 12.35 ± 8.57 & 38.63 ± 4.96 & 55.5 ± 3.28 & 84.76 ± 1.6 \\
   & BUDDY & 12.62 ± 6.69 & 29.64 ± 5.71 & 59.47 ± 5.49 & 91.42 ± 1.26 \\
 & Neo-GNN & 4.53 ± 1.96 & 33.36 ± 9.9 & 64.1 ± 4.31 & 87.76 ± 1.37 \\
  & NCN & 19.34 ± 9.02 & 38.39 ± 7.01 & \colorsecond{74.38 ± 3.15} & \colorsecond{95.56 ± 0.79} \\
 & NCNC & 9.79 ± 4.56 & 34.31 ± 8.87 & \colorfirst{75.07 ± 1.95} & \colorfirst{95.62 ± 0.84} \\
 & NBFNet & \colorfirst{29.94 ± 5.78} & 38.29 ± 3.03 & 62.79 ± 2.53 & 88.63 ± 0.46 \\
 & PEG & 5.88 ± 1.65 & 30.53 ± 6.42 & 62.49 ± 4.05 & 91.42 ± 0.8 \\
 \bottomrule
\end{tabular}
 \label{table:app_cora_ben_more}
 \end{adjustbox}
\end{table}

\begin{table}[]
\centering
\footnotesize
 \caption{ Additional results on Citeseer(\%) under the existing evaluation setting. Highlighted are the results ranked \colorfirst{first}, \colorsecond{second}, and  \colorthird{third}.}
 \begin{adjustbox}{width =0.7 \textwidth}
\begin{tabular}{cc|cccc}
\toprule
% &\multicolumn{5}{c}{Citeseer} \\
 & Models & \multicolumn{1}{c}{Hits@1} & \multicolumn{1}{c}{Hits@3} & \multicolumn{1}{c}{Hits@10} & \multicolumn{1}{c}{Hits@100} \\
 \midrule
 \multirow{5}{*}{Heuristic}& CN & \multicolumn{1}{c}{13.85} & \multicolumn{1}{c}{35.16} & \multicolumn{1}{c}{35.16} & \multicolumn{1}{c}{35.16} \\
 & AA & \multicolumn{1}{c}{21.98} & \multicolumn{1}{c}{35.16} & \multicolumn{1}{c}{35.16} & \multicolumn{1}{c}{35.16} \\
 & RA & \multicolumn{1}{c}{18.46} & \multicolumn{1}{c}{35.16} & \multicolumn{1}{c}{35.16} & \multicolumn{1}{c}{35.16} \\
 & Shortest Path & \multicolumn{1}{c}{0} & \multicolumn{1}{c}{53.41} & \multicolumn{1}{c}{56.92} & \multicolumn{1}{c}{62.64} \\
 & Katz & \multicolumn{1}{c}{24.18} & \multicolumn{1}{c}{54.95} & \multicolumn{1}{c}{57.36} & \multicolumn{1}{c}{62.64} \\
  \midrule
  \multirow{3}{*}{Embedding}& Node2Vec & 30.24 ± 16.37 & 54.15 ± 6.96 & 68.79 ± 3.05 & 89.89 ± 1.48 \\
 & MF & 19.25 ± 6.71 & 29.03 ± 4.82 & 38.99 ± 3.26 & 59.47 ± 2.69 \\
 & MLP & 30.22 ± 10.78 & 56.42 ± 7.90 & 69.74 ± 2.19 & 91.25 ± 1.90 \\
 \midrule
  \multirow{4}{*}{GNN}& GCN & 37.47 ± 11.30 & 62.77 ± 6.61 & 74.15 ± 1.70 & 91.74 ± 1.24 \\
 & GAT & 34.00 ± 11.14 & 62.72 ± 4.60 & 74.99 ± 1.78 & 91.69 ± 2.11 \\
 & SAGE & 27.08 ± 10.27 & 65.52 ± 4.29 & 78.06 ± 2.26 &\colorsecond{ 96.50 ± 0.53} \\
 & GAE & \colorfirst{54.06 ± 5.8} & 65.3 ± 2.54 & \colorsecond{81.72 ± 2.62} & 95.17 ± 0.5 \\
  \midrule
  \multirow{7}{*}{GNN+heuristic} & SEAL & 31.25 ± 8.11 & 46.04 ± 5.69 & 60.02 ± 2.34 & 85.6 ± 2.71 \\
   & BUDDY & \colorthird{49.01 ± 15.07} & \colorthird{67.01 ± 6.22} & \colorthird{80.04 ± 2.27} & 95.4 ± 0.63 \\
 & Neo-GNN & 41.01 ± 12.47 & 59.87 ± 6.33 & 69.25 ± 1.9 & 89.1 ± 0.97 \\
  & NCN & 35.52 ± 13.96 & 66.83 ± 4.06 & 79.12 ± 1.73 & \colorthird{96.17 ± 1.06} \\
 & NCNC & \colorsecond{53.21 ± 7.79} & \colorsecond{69.65 ± 3.19} & \colorfirst{82.64 ± 1.4} & \colorfirst{97.54 ± 0.59} \\
 & NBFNet & 17.25 ± 5.47 & 51.87 ± 2.09 & 68.97 ± 0.77 & 86.68 ± 0.42 \\

 & PEG & 39.19 ± 8.31 & \colorfirst{70.15 ± 4.3} & 77.06 ± 3.53 & 94.82 ± 0.81 \\

\bottomrule
\end{tabular}
 \label{table:app_citeseer_ben_more}
 \end{adjustbox}
\end{table}

\begin{table}[]
\centering
\footnotesize
 \caption{ Additional results on Pubmed(\%) under the existing evaluation setting. Highlighted are the results ranked \colorfirst{first}, \colorsecond{second}, and  \colorthird{third}.}
 \begin{adjustbox}{width =0.7 \textwidth}
\begin{tabular}{cc|cccc}
\toprule
% &\multicolumn{5}{c}{Citeseer} \\
 & Models & \multicolumn{1}{c}{Hits@1} & \multicolumn{1}{c}{Hits@3} & \multicolumn{1}{c}{Hits@10} & \multicolumn{1}{c}{Hits@100} \\
 \midrule
 \multirow{5}{*}{Heuristic}& CN & \multicolumn{1}{c}{7.06} & \multicolumn{1}{c}{12.95} & \multicolumn{1}{c}{27.93} & \multicolumn{1}{c}{27.93} \\
 & AA & \multicolumn{1}{c}{12.95} & \multicolumn{1}{c}{16} & \multicolumn{1}{c}{27.93} & \multicolumn{1}{c}{27.93} \\
 & RA & \multicolumn{1}{c}{11.67} & \multicolumn{1}{c}{15.21} & \multicolumn{1}{c}{27.93} & \multicolumn{1}{c}{27.93} \\
 & Shortest Path & \multicolumn{1}{c}{0} & \multicolumn{1}{c}{0} & \multicolumn{1}{c}{27.93} & \multicolumn{1}{c}{60.36} \\
 & Katz & \multicolumn{1}{c}{12.88} & \multicolumn{1}{c}{25.38} & \multicolumn{1}{c}{42.17} & \multicolumn{1}{c}{61.8} \\
 \midrule
   \multirow{3}{*}{Embedding}& Node2Vec & \colorthird{29.76 ± 4.05} & 34.08 ± 2.43 & 44.29 ± 2.62 & 63.07 ± 0.34 \\
 & MF & 12.58 ± 6.08 & 22.51 ± 5.6 & 32.05 ± 2.44 & 53.75 ± 2.06 \\
 & MLP & 7.83 ± 6.40 & 17.23 ± 2.79 & 34.01 ± 4.94 & 84.19 ± 1.33 \\
 \midrule
   \multirow{4}{*}{GNN}& GCN & 5.72 ± 4.28 & 19.82 ± 7.59 & 56.06 ± 4.83 & 87.41 ± 0.65 \\
 & GAT & 6.45 ± 10.37 & 23.02 ± 10.49 & 46.77 ± 4.03 & 80.95 ± 0.72 \\
 & SAGE & 11.26 ± 6.86 & 27.23 ± 7.48 & 48.18 ± 4.60 & \colorthird{90.02 ± 0.70} \\
 & GAE & 1.99 ± 0.12 & 31.75 ± 1.13 & 45.48 ± 1.07 & 84.3 ± 0.31 \\
  \midrule
   \multirow{7}{*}{GNN+heuristic}& SEAL & \colorsecond{30.93 ± 8.35} & \colorsecond{40.58 ± 6.79} & 48.45 ± 2.67 & 76.06 ± 4.12 \\
    & BUDDY & 15.31 ± 6.13 & 29.79 ± 6.76 & 46.62 ± 4.58 & 83.21 ± 0.59 \\
 & Neo-GNN & 19.95 ± 5.86 & 34.85 ± 4.43 & \colorthird{56.25 ± 3.42} & 86.12 ± 1.18 \\
  & NCN & 26.38 ± 6.54 & \colorthird{36.82 ± 6.56} & \colorfirst{62.15 ± 2.69} & \colorsecond{90.43 ± 0.64} \\
 & NCNC & 9.14 ± 5.76 & 33.01 ± 6.28 & \colorsecond{61.89 ± 3.54} & \colorfirst{91.93 ± 0.6} \\
 & NBFNet & \colorfirst{40.47 ± 2.91} & \colorfirst{44.7 ± 2.58} & 54.51 ± 0.84 & 79.18 ± 0.71 \\

 & PEG & 8.52 ± 3.73 & 24.46 ± 6.94 & 45.11 ± 4.02 & 76.45 ± 3.83 \\

 \bottomrule
\end{tabular}
 \label{table:app_pubmed_ben_more}
 \end{adjustbox}
\end{table}

\begin{table}[]
\centering
% \footnotesize
 \caption{Additional results on OGB datasets(\%) under the existing evaluation setting. Highlighted are the results ranked \colorfirst{first}, \colorsecond{second}, and  \colorthird{third}.}
 \begin{adjustbox}{width =1 \textwidth}
\begin{tabular}{c|ccccccccc}
\toprule
 & \multicolumn{2}{c}{ogbl-collab} &  \multicolumn{2}{c}{ ogbl-ddi} & \multicolumn{2}{c}{ogbl-ppa}   & \multicolumn{3}{c}{ogbl-citation2}     \\
 & Hits@20 & Hits@100 & Hits@50 & Hits@100 & Hits@20 & Hits@50 & Hits@20 & Hits@50 & Hits@100 \\
 \midrule
CN & 49.98 & 65.6 & 26.51 & 34.52 & 13.26 & 19.67 & 77.99&	77.99	&77.99 \\
AA & {55.79} & 65.6 & 27.07 & 36.35 & 14.96 & 21.83 & 77.99	&77.99	&77.99 \\
RA & 55.01 & 65.6 & 19.14 & 31.17 & 25.64 & \colorthird{38.81} & 77.99	&77.99	&77.99 \\
Shortest Path & 46.49 & 66.82 & 0 & 0 & 0 & 0 & >24h & >24h & >24h \\
Katz & \colorfirst{58.11} & \colorsecond{71.04} & 26.51	&34.52 & 13.26	&19.67 & 78	&78	&78 \\
 \midrule
Node2Vec & 40.68 ± 1.75 & 55.58 ± 0.77 & 59.19 ± 3.61 & 73.49 ± 3.18 & 11.22 ± 1.91 & 19.22 ± 1.69 & 82.8 ± 0.13 & 92.33 ± 0.1 & 96.44 ± 0.03 \\
MF & 39.99 ± 1.25 & 43.22 ± 1.94 & 45.51 ± 11.13 & 61.72 ± 6.56 & 9.33 ± 2.83 & 21.08 ± 3.92 & 70.8 ± 12.0 & 74.48 ± 10.42 & 75.5 ± 10.13 \\
MLP & 27.66 ± 1.61 & 42.13 ± 1.09 & N/A & N/A & 0.16 ± 0.0 & 0.26 ± 0.03 & 74.16 ± 0.1 & 86.59 ± 0.08 & 93.14 ± 0.06 \\
 \midrule
GCN & 44.92 ± 3.72 & 62.67 ± 2.14 & 74.54 ± 4.74 & 85.03 ± 3.41 & 11.17 ± 2.93 & 21.04 ± 3.11 & \colorfirst{98.01 ± 0.04 }&\colorfirst{99.03 ± 0.02} & \colorsecond{99.48 ± 0.02} \\
GAT & 43.59 ± 4.17 & 62.24 ± 2.29 & 55.46 ± 10.16 & 69.74 ± 10.01 & OOM & OOM & OOM & OOM & OOM \\
SAGE & 50.77 ± 2.33 & 65.36 ± 1.05 & \colorsecond{93.48 ± 1.36} & \colorsecond{97.37 ± 0.55} &  19.37 ± 2.65&	31.3 ± 2.36& 97.48 ± 0.03 & 98.75 ± 0.03 & 99.3 ± 0.02 \\
GAE & OOM & OOM & 12.39 ± 8.74 & 14.03 ± 9.22 & OOM & OOM & OOM & OOM & OOM \\
 \midrule
SEAL & 54.19 ± 1.57 & {69.94 ± 0.72} & 43.34 ± 3.23 & 52.2 ± 1.78 & 21.81 ± 4.3 & 36.88 ± 4.06 & 94.61 ± 0.11 & 95.0 ± 0.12 & 95.37 ± 0.14 \\
BUDDY & \colorsecond{57.78 ± 0.59} & 67.87 ± 0.87 & 53.36 ± 2.57 & 71.04 ± 2.56 & \colorthird{26.33 ± 2.63} & 38.18 ± 1.32 & \colorthird{97.79 ± 0.07} & \colorthird{98.86 ± 0.04} & 99.38 ± 0.03 \\
Neo-GNN & \colorthird{57.05 ± 1.56} & \colorfirst{71.76 ± 0.55} & 33.88 ± 10.1 & 46.55 ± 13.29 & 26.16 ± 1.24 & 37.95 ± 1.45 & 97.05 ± 0.07 & 98.75 ± 0.03 & \colorthird{99.41 ± 0.02} \\
NCN & 50.27 ± 2.72 & 67.58 ± 0.09 & \colorfirst{95.51 ± 0.87} &\colorfirst{ 97.54 ± 0.7}& \colorfirst{40.29 ± 2.22} & \colorfirst{53.35 ± 1.77} & \colorsecond{97.97 ± 0.03} &\colorsecond{ 99.02 ± 0.02} & \colorfirst{99.5 ± 0.01} \\
NCNC & 54.91 ± 2.84 & \colorthird{70.91 ± 0.25} & \colorthird{92.34 ± 2.42} &\colorthird{96.35 ± 0.52} & \colorsecond{40.1 ± 1.06} & \colorsecond{52.09 ± 1.99} & 97.22 ± 0.78 & 98.2 ± 0.71 & 98.77 ± 0.6 \\
NBFNet & OOM & OOM & >24h & >24h & OOM & OOM & OOM & OOM & OOM \\
PEG & 33.57 ± 7.40 & 55.14 ± 2.10 & 47.93 ± 3.18 & 59.95 ± 2.52 & OOM & OOM & OOM & OOM & OOM \\
\bottomrule
\end{tabular}
 \label{table:app_ogb_ben_more}
\end{adjustbox}
\end{table}

\section{Additional Details on HeaRT} \label{sec:appendix_neg_sample}

As described in Section~\ref{sec:new_eval_procedure}, given a positive sample $(a, b)$, we seeks to generate $K$ negative samples to evaluate against. The negative samples are drawn from the set of possible corruptions of $(a, b)$, i.e, $S(a, b)$ (see Eq.~\eqref{eq:candidate_set}). Multiple heuristics are used to determine which $K$ negative samples to use. Furthermore, the negative samples are split evenly between both nodes. That is, we generate $K/2$ negative samples that contain either node $a$ and $b$, respectively. This process is illustrated in Figure~\ref{fig:gen_hard_negatives}.

The rest of this section is structured as follows. In Section~\ref{sec:diff_sample} we  describe how we use multiple heuristics for estimating the difficulty of negative samples. Then in Section~\ref{sec:combine_ranks} we describe how we combine the ranks given by different heuristic methods.

\subsection{Determining Hard Negative Samples} \label{sec:diff_sample} 

% We are first tasked with {\it how to choose the negative samples}. As discussed and shown in Section~\ref{sec:new_eval_procedure}, we want to select the negative samples from $S(a, b)$ such that they are non-trivial to classify. This requires the use of some criteria that can be used a proxy for difficulty. To achieve this, we employ the use of multiple heuristics to rank the different negative samples. We use heuristics as they correlate well with link prediction performance (see Table~\ref{table:ogb}) and are typically efficient to calculate. Furthermore, this is common in real-world applications~\cite{gupta2013wtf, eksombatchai2018pixie} where candidates are drawn according to a simpler approach and then ranked.

We are first tasked with {\it how to choose the negative samples}. As discussed and shown in Section~\ref{sec:new_eval_procedure}, we want to select the negative samples from $S(a, b)$ such that they are non-trivial to classify. {Hence, as inspired by the candidate generation process in real-world recommender systems~\cite{gupta2013wtf, eksombatchai2018pixie}, we aim to select a set of 'hard' negative samples that are more relevant to the source node. The candidate generation process is typically based on some primitive and simple link prediction heuristics. These heuristics can be also treated as link prediction methods (see Tables~\ref{table:small} and \ref{table:ogb}).}

% We use multiple heuristics that capture a variety of different information. To capture {\it local} information we use resource allocation (RA)~\cite{zhou2009predicting}, a CN-based approach. To capture {\it global} information we use the personalized pagerank score (PPR)~\cite{pagerank}. Lastly, we further include the cosine feature similarity for the Cora, Citeseer, and Pubmed datasets. This is due to the strong performance of a MLP on those datasets. 

{
We use multiple heuristics that capture a variety of different information. Most link prediction heuristics can be categorized into two main categories: local heuristics and global heuristics~\cite{lu2011link}. Local heuristics attempt to capture the local neighborhood information that exists near the node pair while global heuristics attempt to use the whole graph structure. To capture the {\it local} information we use resource allocation (RA)~\cite{zhou2009predicting}, a CN-based approach. Existing results show that RA can achieve strong performance on most datasets (see Tables~\ref{table:small} and~\ref{table:ogb}). To measure the {\it global} information we use the personalized pagerank score (PPR)~\cite{pagerank}. Random walk based methods are commonly used for candidate generation~\cite{gupta2013wtf, eksombatchai2018pixie}. Lastly, we further include the cosine feature similarity for the Cora, Citeseer, and Pubmed datasets. This is due to the strong performance of a MLP on those datasets. By combining these heuristics, we are able to generate a diverse set of negative samples for each positive sample.
}

For each heuristic we then rank all the possible negative samples. We first denote the score of a heuristic $i$ for a pair of nodes $a$ and $b$ as $h_i(a, b)$. Let's say we want to rank all negative samples that contain a node $a$, i.e., $(a, *)$.  The rank across all nodes is given by:
\begin{equation}
    R_i = \underset{v \in \bar{V}}{\text{ArgSort}} \; h_i(a, v),
\end{equation}
where $R_i$ denotes the ranking for heuristic $i$ and and $\bar{V}$ is a subset of the set of nodes in the graph $V$. 
We apply a filtering process to exclude all positive training samples, self-loops, and the sample itself under consideration from being selected as negative samples. Additionally, when choosing negative samples for the test samples, we disallow validation samples to be chosen as well. As such, we only consider a subset of nodes $\bar{V} \in V$. This is analogous to the filtered setting used in KGC~\cite{bordes2013translating}. However, we adopt a distinct filtering strategy for the ogbl-collab dataset, which is a dynamic collaboration graph. Specifically, {\bf positive training samples are not excluded in the generation of negative samples for validation and test}. Similarly, positive validation samples are not omitted when creating negative test samples. This approach is justified for ogbl-collab which is a dynamic graph, as prior collaboration between authors does not necessarily indicates future collaborations. Further details and discussions are provided in  Appendix~\ref{sec:collab_observation}.

We now have all possible negative samples ranked according to multiple heuristics. However, it is unclear how to choose the negative samples from multiple ranked lists. In the next subsection we detail how we combine the ranks according to each heuristic. This will give us a final ranking, of which we can choose the top $K/2$ as the negative samples for that node.

\subsection{Combining Heuristic Ranks} \label{sec:combine_ranks}  

In this subsection we tackle the problem of combing the negative sample ranks given by multiple heuristics. More concretely, say we use $m$ heuristics and rank all the  samples according to each. We want to arrive at a combined ranking $R_{\text{total}}$ that is composed of each rank,   
\begin{equation}
    R_{\text{total}} = \phi(R_1, R_2, \cdots, R_m).
\end{equation}
We model $\phi$ via Borda's method~\cite{borda1781m}.  
Let $R_i(a, v)$ be the rank of the node pair $(a, v)$ for heuristic $i$. The combined rank $R_{\text{total}}(a, v)$ across $m$ ranked lists is given by:
\begin{equation} \label{eq:combine_ranks}
    R_{\text{total}}(a, v) = g\left(R_1(a, v), R_2(a, v), \cdots, R_m(a, v)\right),
\end{equation}
where $g$ is an aggregation function. We set $g = \text{min}(\cdot)$. This is done as it allows us to capture a more distinct set of samples by selecting the ``best" for each heuristic. This is especially true when there is strong disagreement between the different heuristics. A final ranking is then done on $R_{\text{total}}$ to select the top nodes,
\begin{equation}
    R_f = \underset{v \in \bar{V}}{\text{ArgSort}} \; R_{\text{total}},
\end{equation}
where $R_f$ is the final ranking. The highest $K/2$ nodes are then selected from $R_f$. Lastly, we note that for some nodes there doesn't exist sufficient scores to rank $K/2$ total nodes. In this case the remaining nodes are chosen randomly.
The full generation process for a node $a$ is detailed in Algorithm~\ref{alg:new_eval}.

\begin{algorithm}[t]
\small
% \captionsetup{font=small, labelfont=small}
\caption{Generating Negative Samples of Form $(a, *)$} 
\begin{algorithmic}[1]

\Require
    \Statex $a = $ Node to generate samples for 
    \Statex $\bar{V} =$ Possible nodes to use for negative samples
    \Statex $\mathcal{H} = \{h_1, h_2, \cdots, h_m \}$ \Comment{Set of $m$ heuristics}
\newline
\For{$i \in \lvert \mathcal{H} \rvert$}
    \State $R_i = \underset{v \in \bar{V}}{\text{ArgSort}} \; h_i(a, v)$ \Comment{Sort by each heuristic individually}
\EndFor
\For{$v \in \bar{V}$}
    \State $R_{\text{total}}(a, v) = \text{min} \left(R_1(a, v), R_2(a, v), \cdots, R_m(a, v) \right)$ \Comment{Combine the rankings}
\EndFor
\newline 
\State $R_f = \underset{v \in \bar{V}}{\text{ArgSort}} \; R_{\text{total}}(a, v)$ \Comment{Sort by combined ranking}
\newline 
\State \Return $R_f[:K/2]$ \Comment{Return the top K/2 ranked nodes}
\end{algorithmic}
\label{alg:new_eval}
\end{algorithm}

\section{Additional Results Under HeaRT}
\label{sec:app_new_more}
We present additional results of Cora, Citeseer, Pubmed, and OGB datasets under HeaRT in Table~\ref{table:app_small_new} and Table~\ref{table:app_ogb_new}. These results include other hit metrics not found in the main tables.

\begin{table}[]
\centering
% \footnotesize
 \caption{ Additional results on Cora, Citeseer, and Pubmed(\%) under  HeaRT. Highlighted are the results ranked \colorfirst{first}, \colorsecond{second}, and  \colorthird{third}.}
 \begin{adjustbox}{width =1 \textwidth}
\begin{tabular}{c|ccccccccc}
\toprule
   &\multicolumn{3}{c}{Cora} & \multicolumn{3}{c}{Citeseer}  &  \multicolumn{3}{c}{Pubmed}    \\
 
  Models & Hits1 & Hits3 & Hits100 & Hits1 & Hits3 & Hits100 & Hits1 & Hits3 & Hits100 \\
   \midrule
  CN & 3.98 & 10.25 & 38.71 & 2.2 & 9.45 & 33.63 & 0.47 & 1.49 & 19.29 \\
  AA & 5.31 & 12.71 & 38.9 & 3.96 & 12.09 & 33.85 & 0.74 & 1.87 & 20.37 \\
  RA & 5.31 & 12.52 & 38.52 & 4.18 & 11.21 & 34.07 & 0.72 & 1.78 & 20.04 \\
  Shortest Path & 0.57 & 2.85 & 55.6 & 0.22 & 3.52 & 53.85 & 0 & 0.02 & 21.57 \\
  Katz & 4.64 & 11.95 & 59.96 & 3.74 & 11.87 & 55.82 & 0.74 & 2.12 & 32.78 \\
 \midrule
  Node2Vec & 5.69 ± 0.81 & 15.1 ± 0.99 & 77.21 ± 2.34 & 9.63 ± 0.82 & 23.5 ± 1.37 & 84.46 ± 1.86 & 0.75 ± 0.14 & 2.4 ± 0.32 & 52.27 ± 0.65 \\
  MF & 1.46 ± 0.8 & 5.46 ± 1.67 & 59.68 ± 3.41 & 2.68 ± 0.92 & 7.25 ± 0.98 & 53.25 ± 2.91 & 1.13 ± 0.24 & 3.25 ± 0.44 & 50.56 ± 1.11 \\
  MLP & 5.48 ± 0.99 & 14.15 ± 1.56 & 77.0 ± 1.02 & 10.44 ± 0.82 & 26.46 ± 1.24 & 86.83 ± 1.36 & 1.28 ± 0.22 & 4.33 ± 0.28 & 76.34 ± 0.79 \\
 \midrule
  GCN & \colorsecond{7.59 ± 0.61} & \colorsecond{17.46 ± 0.82} & \colorsecond{85.47 ± 0.52} & 9.27 ± 0.99 & 23.19 ± 0.98 & 89.1 ± 2.13 & 2.09 ± 0.31 & 5.58 ± 0.27 & 73.59 ± 0.53 \\
  GAT & 5.03 ± 0.81 & 13.66 ± 0.67 & 80.87 ± 1.32 & 8.02 ± 1.21 & 20.09 ± 0.82 & 86.83 ± 1.09 & 1.14 ± 0.16 & 3.06 ± 0.36 & 67.06 ± 0.69 \\
  SAGE & 5.48 ± 0.97 &\colorthird{ 15.43 ± 1.07} & 81.61 ± 0.96 & 8.37 ± 1.62 & 23.74 ± 1.62 & \colorthird{92.33 ± 0.68} & \colorfirst{3.03 ± 0.46} &\colorfirst{ 8.19 ± 1.0} & \colorfirst{79.47 ± 0.53 }\\
  GAE & \colorfirst{9.72 ± 0.73} & \colorfirst{19.24 ± 0.76} & 79.66 ± 0.95 & \colorsecond{13.81 ± 0.82} & \colorsecond{27.71 ± 1.34} & 85.49 ± 1.37 & 1.48 ± 0.23 & 4.05 ± 0.39 & 59.79 ± 0.67 \\
 \midrule
  SEAL & 3.89 ± 2.04 & 10.82 ± 4.04 & 61.9 ± 13.97 & 5.08 ± 1.31 & 13.68 ± 1.32 & 68.94 ± 2.3 & 1.47 ± 0.32 & 4.71 ± 0.68 & 65.81 ± 2.43 \\
  BUDDY & 5.88 ± 0.60 & 13.76 ± 1.03 & 82.46 ± 1.79 & 10.09 ± 0.50 & 26.11 ± 1.26 & \colorsecond{92.66 ± 0.92} & \colorthird{2.24 ± 0.17} & 5.93 ± 0.21 & 72.01 ± 0.46 \\
  Neo-GNN & 5.71 ± 0.41 & 13.89 ± 0.82 & 80.28 ± 1.08 & 6.81 ± 0.73 & 17.8 ± 1.19 & 85.51 ± 1.01 & 1.90 ± 0.24 & \colorthird{6.07 ± 0.47} & \colorthird{76.57 ± 0.58} \\
  NCN & 4.85 ± 0.81 & 14.46 ± 0.98 & \colorthird{84.14 ± 1.24} & \colorfirst{16.77 ± 2.05} & \colorfirst{30.51 ± 0.97} & 90.42 ± 0.98 & 1.13 ± 0.18 & 3.95 ± 0.24 & 71.46 ± 0.97 \\
  NCNC & 4.78 ± 0.71 & 14.72 ± 1.24 & \colorfirst{85.62 ± 0.83} & \colorthird{11.14 ± 0.82} & \colorthird{27.21 ± 0.96} &\colorfirst{ 92.73 ± 1.16} &\colorsecond{ 2.73 ± 0.49} & \colorsecond{7.05 ± 0.72} & \colorsecond{79.22 ± 0.96} \\
  NBFNet & 5.31 ± 1.16 & 14.95 ± 0.72 & 76.24 ± 0.68 & 5.95 ± 1.06 & 14.53 ± 1.19 & 72.66 ± 0.95 & >24h &>24h  & >24h \\
  PEG & \colorthird{6.98 ± 0.57 }& 14.93 ± 0.61 & 82.52 ± 1.28 & 9.93 ± 0.6 & 21.91 ± 0.59 & 90.15 ± 1.43 & 0.88 ± 0.18 & 2.61 ± 0.39 & 64.95 ± 1.81 \\
 \bottomrule
\end{tabular}
\end{adjustbox}
\label{table:app_small_new}
\end{table}

\begin{table}[]
\centering
% \footnotesize
 \caption{ Additional results on OGB datasets(\%) under  HeaRT. Highlighted are the results ranked \colorfirst{first}, \colorsecond{second}, and  \colorthird{third}.}
 \begin{adjustbox}{width =1 \textwidth}
\begin{tabular}{c|cccccccc}
\toprule
 & \multicolumn{2}{c}{ogbl-collab}  & \multicolumn{2}{c}{ogbl-ddi}   & \multicolumn{2}{c}{ogbl-ppa}   & \multicolumn{2}{c}{ogbl-citation2}   \\
 Models& Hits50 & Hits100 & Hits50 & Hits100 & Hits50 & Hits100 & Hits50 & Hits100 \\
 \midrule
CN & {30.52} & {42.80} & 70.12 & 86.53 & 80.53 & 86.51 & 57.56	&68.04 \\
AA & {33.74} & {45.20} & 71.08 & 87.36 & 81.93 & 87.55 & 58.87	&69.39 \\
RA & \colorthird{36.68} & {46.42}  & 76.39 & 90.96 & 81.65 & 86.84 & 58.88	&68.83 \\
Shortest Path & {33.77} & {45.85} & 0 & 0 & 1.34 & 1.4 & \textgreater{}24h & \textgreater{}24h \\
Katz & {\colorfirst{39.18}} & {48.80}& 70.12&	86.53 &80.53	&86.51 & 54.97 & 67.56 \\
\midrule
Node2Vec & 28.56 ± 0.17 & 41.84 ± 0.25& \colorthird{98.38 ± 0.7} & 99.91 ± 0.01 & 69.94 ± 0.06 & 81.88 ± 0.06 & 61.22 ± 0.16 & 77.11 ± 0.13 \\
MF & 30.83 ± 0.22 & 43.23 ± 0.34 & 95.52 ± 0.72 & 99.54 ± 0.08 & 83.29 ± 3.35 & 89.75 ± 1.9 & 29.64 ± 7.3 & 65.87 ± 8.37 \\
MLP &28.88 ± 0.32 & 46.83 ± 0.33 & N/A & N/A & 5.36 ± 0.0 & 22.01 ± 0.01 & 61.29 ± 0.07 & 76.94 ± 0.1 \\
\midrule
GCN & {35.29 ± 0.49} & {\colorfirst{50.83 ± 0.21}}  & 97.65 ± 0.68 & 99.85 ± 0.06 & 81.48 ± 0.48 & 89.62 ± 0.23 & 70.77 ± 0.34 & \colorthird{85.43 ± 0.18} \\
GAT & 32.92 ± 1.41 & 46.71 ± 0.84 & 98.15 ± 0.24 & 99.93 ± 0.02 & OOM & OOM & OOM & OOM \\
SAGE & 33.48 ± 1.40 & 48.33 ± 0.49 &\colorfirst{ 99.17 ± 0.11} & \colorfirst{99.98 ± 0.01 }& 81.84 ± 0.24 & 89.46 ± 0.13 & \colorsecond{71.91 ± 0.1} & \colorsecond{85.86 ± 0.09}\\
GAE & OOM & OOM & 28.29 ± 13.65 & 48.34 ± 15.0 & OOM & OOM & OOM & OOM \\
\midrule
SEAL &  33.57 ± 0.84 & 43.06 ± 1.09 & 82.42 ± 3.37 & 92.63 ± 2.05 & \colorthird{87.34 ± 0.49} & \colorthird{92.45 ± 0.26} & 65.11 ± 2.33 & 77.64 ± 2.43 \\
BUDDY &\colorsecond{39.04 ± 0.11} & \colorsecond{50.49 ± 0.09} & 97.81 ± 0.31 & \colorthird{99.93 ± 0.01} & 82.5 ± 0.51 & 88.36 ± 0.32 & 67.47 ± 0.32 & 81.94 ± 0.26 \\
Neo-GNN & 36.11 ± 2.36 & \colorthird{49.25 ± 0.81} & 83.45 ± 11.03 & 94.7 ± 4.82 & 81.21 ± 1.39 & 88.31 ± 0.19 & 62.14 ± 0.51 & 79.13 ± 0.42 \\
NCN & 34.53 ± 0.98 & 45.69 ± 0.42 & \colorsecond{98.43 ± 0.22} & \colorsecond{99.96 ± 0.01} & \colorsecond{89.37 ± 0.28} & \colorsecond{93.11 ± 0.27} & \colorthird{71.56 ± 0.03} & 84.01 ± 0.05 \\
NCNC & 34.96 ± 3.80 & 46.93 ± 2.04 & \textgreater{}24h & \textgreater{}24h & \colorfirst{91.0 ± 0.24} & \colorfirst{94.72 ± 0.18} & \colorfirst{72.85 ± 0.9} & \colorfirst{86.35 ± 0.51} \\
NBFNet & OOM & OOM & \textgreater{}24h & \textgreater{}24h & OOM & OOM & OOM & OOM \\
PEG & 30.12 ± 0.63 & 45.40 ± 0.66 & 84.21 ± 9.2 & 95.76 ± 3.48 & OOM & OOM & OOM & OOM \\
\bottomrule
\end{tabular}
\end{adjustbox}
\label{table:app_ogb_new}
\end{table}

\section{Additional Investigation on ogbl-collab} \label{sec:collab_observation}

% Please add the following required packages to your document preamble:
% \usepackage[table,xcdraw]{xcolor}
% Beamer presentation requires \usepackage{colortbl} instead of \usepackage[table,xcdraw]{xcolor}

\begin{table}[]
\centering
% \footnotesize
 \caption{ Results on ogbl-collab under HeaRT when {\bf excluding the positive train/validation samples} of being negative samples during testing. Highlighted are the results ranked \colorfirst{first}, \colorsecond{second}, and  \colorthird{third}.}
 \label{table:collab_new_negative}
 \begin{adjustbox}{width =0.6 \textwidth}
\begin{tabular}{c|cccc}
\toprule

 & \textbf{MRR} & \textbf{Hits20} & \textbf{Hits50} & \textbf{Hits100} \\
 \midrule
CN   &12.60  & 27.51  & 38.39 & 47.4 \\
AA   & {16.40} &{32.65} & 42.61 & 50.25 \\
RA   & {28.14} & {41.16}  & 46.9 & 51.78  \\
Shortest Path & \colorsecond{46.71} & \colorsecond{46.56} &\colorthird{46.97} & 48.11 \\
 Katz &{\colorfirst{47.15}} &{\colorfirst{48.66}} & \colorfirst{51.07} & 54.28 \\
\midrule
Node2Vec & 12.10 ± 0.20 & 25.85 ± 0.21 & 35.49 ± 0.22 & 46.12 ± 0.34 \\
MF & 26.86 ± 1.74 & 38.44 ± 0.07  & 43.62 ± 0.08 & 51.75 ± 0.14 \\
MLP & 12.61 ± 0.66  & 23.05 ± 0.89  & 35.32 ± 0.74 & 51.09 ± 0.37 \\
\midrule
GCN & 18.28 ± 0.84  & 32.90 ± 0.66 &43.17 ± 0.36 & \colorsecond{54.93 ± 0.14} \\
GAT  & 10.97 ± 1.16 & 29.58 ± 2.42  & 42.07 ± 1.51 & 53.45 ± 0.64\\
SAGE & 20.89 ± 1.06 & 33.83 ± 0.93 & 43.02 ± 0.63 & \colorthird{54.38 ± 0.27}\\
GAE & OOM & OOM & OOM & OOM \\
\midrule
SEAL  & 22.53 ± 3.51  & 36.48 ± 2.55  & 43.5 ± 1.75 & 49.25 ± 1.39 \\
BUDDY & \colorthird{32.42 ± 1.88} & \colorthird{45.62 ± 0.52}& \colorsecond{ 50.57 ± 0.18} & \colorfirst{55.63 ± 0.68} \\
 Neo-GNN   & 21.90 ± 0.65  & 38.40 ± 0.29  & 46.93 ± 0.17 & 53.81 ± 0.19 \\
NCN & 17.51 ± 2.50  & 37.07 ± 2.97 & 45.89 ± 1.11 & 52.36 ± 0.33 \\
NCNC & 19.02 ± 5.32  & 35.67 ± 6.78  & 44.76 ± 4.64 & 52.41 ± 2.09 \\
NBFNet & OOM & OOM & OOM & OOM \\
PEG   & 15.68 ± 1.10  & 29.74 ± 0.95  &38.71 ± 0.17 & 49.34 ± 0.70  \\
 \bottomrule
\end{tabular}
\end{adjustbox}
\end{table}

As introduced in Section~\ref{sec:diff_sample}, we adopt a different strategy to generate the hard negative samples for ogbl-collab which is a dynamic collaboration graph. In this dataset, nodes represent authors and edges represent a collaboration between two authors. Each edge further includes an attribute that specifies the year of collaboration. Specifically, each edge takes the form of (Author 1, Year, Author 2). The task is to predict collaborations in 2019 (test) based on those until 2017 (training) and 2018 (validation).

Contrary to other datasets, we \textbf{do not exclude} the positive training samples when generating the negative samples for validation and test. We note that we also do not exclude positive validation edges when generating the negatives for test. In simpler terms, when creating negative samples for testing, both positive samples from training and validation are considered. 
This means that negative samples during testing could present in the training and validation positive samples.
This approach is reasonable and well-aligned with the real-world scenario in the context of collaboration graphs. Specifically, authors who collaborated in the past might not do so in the future. For instance, just because the positive sample (Author 2, 2017, Author 3) exists, it does not imply that (Author 2, 2019, Author 3) is also true. However, this is implied to be true if we exclude positive train/validation samples from appearing as negative samples during testing.

We validate this approach by contrasting it with the strategy that \textbf{excludes train/validation data when generating hard negatives.}
The  results are presented in Table~\ref{table:collab_new_negative}.  
We observed that under this setting, both the Shortest Path and Katz perform considerably well on ogbl-collab. Specifically, the MRR gap between the second-ranked method (Shortest Path) and the third (BUDDY) is 14.29. We found that it is due to the ogbl-collab being a dynamic graph. Of the positive samples in the test set, around $46\%$ also appear as positive samples in the training set. In particular, an edge (Author 1, 2017, Author 2) in the training data may also ``appear'' in the test data in the form of (Author 1, 2019, Author 2). This is because two authors who collaborated in the past often tend to collaborate again in the future. As such, when evaluating the test sample (Author 1, 2019, Author 2), there exists path of length 1 between the two authors in the graph. Furthermore, this phenomenon is common among positive samples but not observed among negatives. This is because we exclude the positive training samples when generating the negative samples for evaluation. 
{\bf As a result of this exclusion, the presence of a direct link (i.e., a shortest path of length 1) between two authors suggest a positive sample, while its absence often corresponds to a negative sample.} As such, it provides an easy ``shortcut'' to distinguish positive and negative samples during testing. This explains why methods like Shortest Path and Katz can achieve good performance on ogbl-collab when excluding positive train/validation samples of being negative samples during testing.

On the contrary, when allowing positive train/validation samples to also be negative samples for HeaRT, the results on ogbl-collab, shown in Tables~\ref{table:ogb_newsetting} and~\ref{table:app_ogb_new}, indicate that Shortest Path does not maintain its superior performance as observed in Table~\ref{table:collab_new_negative}.  
Additionally, the overall results under HeaRT are inferior to the ones in Table~\ref{table:ogb_newsetting}. For instance, while all the MRR values in Table~\ref{table:ogb_newsetting} exceed 10, the highest MRR in Table~\ref{table:collab_new_negative} is approximately 6. This indicates that excluding those positive samples from being negative samples disproportionately helps the Shortest Path.

\section{Dataset Licenses} \label{sec:code}
The license for each dataset can be found in Table~\ref{table:app_license}.

\begin{table}[!h]
\centering

 \caption{Dataset Licenses.}
  \begin{adjustbox}{width =0.3 \textwidth}
\begin{tabular}{c|c}
\toprule
 Datasets	&License	 \\
  \midrule
 Cora & NLM License \\
 Citeseer & NLM License \\
 Pubmed& NLM License \\
 ogbl-collab & MIT License \\
  ogbl-ddi & MIT License \\
   ogbl-ppa & MIT License \\
    ogbl-citation2 & MIT License \\
\bottomrule
\end{tabular}
\label{table:app_license}
\end{adjustbox}
\end{table}

\section{Limitation}
\label{sec:app_limitation}

One potential limitation of HeaRT lies in the generation of customized negative samples for each positive sample. This design may result in an increased number of negative samples compared to the existing setting. Although this provides a more realistic evaluation, it could have an impact on the efficiency of the evaluation process, especially in scenarios where a significant number of positive samples exist. Nonetheless, this limitation does not detract from the potential benefits of HeaRT in providing a more realistic and meaningful link prediction evaluation setting.  Furthermore, as each evaluation node pair is independent, it offers scope for parallelization, mitigating any potential efficiency concerns to a large extent.
Future work can investigate ways to optimize this process.

\section{Social Impact}
\label{sec:app_social_impact}
Our method HeaRT harbors significant potential for positive societal impact. By aligning the evaluation setting more closely with real-world scenarios, it enhances  the applicability of link prediction research. This not only contributes to the refinement of existing prediction methods but also stimulates the development of more effective link prediction methods. As link prediction has far-reaching implications across numerous domains, from social network analysis to recommendation systems and beyond, improving its performance and accuracy is of paramount societal importance.  We also carefully consider the broader impact from various perspectives such as fairness, security, and harm to people. No apparent risk is related to our work.

\end{document}